\RequirePackage{fix-cm}
\documentclass[smallextended, natbib]{svjour3}      
\smartqed  

\usepackage{graphicx}
\usepackage[justification=centering, labelfont=bf, labelsep=period]{caption}
\usepackage{enumitem}

\usepackage[dvipsnames]{xcolor}
\usepackage{color, colortbl}
\definecolor{carminered}{rgb}{1.0, 0.0, 0.22}
\definecolor{whitesmoke}{rgb}{0.9, 0.9, 0.9}
\definecolor{my_blue}{RGB}{175,225,255}

\usepackage{tabu}
\usepackage{multirow}
\usepackage{arydshln}
\newcolumntype{C}[1]{>{\centering\arraybackslash}m{#1}}

\usepackage{subfig}
\usepackage{adjustbox}
\usepackage{stackengine}

\usepackage{algpseudocode, algorithm, algorithmicx}

\usepackage{mathtools}
\usepackage{amsmath}

\usepackage{amssymb}
\numberwithin{equation}{section}

\usepackage{url}
\usepackage{hyperref}
\hypersetup{
	colorlinks=true,
	linkcolor=blue,
	citecolor=blue,
	urlcolor=blue,
}

\usepackage{marvosym}
\usepackage{hyperref}

\usepackage{soul}
\usepackage{comment}

\usepackage[T1]{fontenc}

\journalname{Noname}

\begin{document}

\title{
	Reservoir of Diverse Adaptive Learners and \\ Stacking Fast Hoeffding Drift Detection Methods for \\ Evolving Data Streams
}


\titlerunning{Adaptive Learners Reservoir and FHDDMS}

\author{
	Ali Pesaranghader\textsuperscript{ \Letter, 1}  \and
	Herna Viktor \textsuperscript{1} 
	Eric Paquet \textsuperscript{1,2}
}

\authorrunning{A. Pesaranghader, H.L. Viktor, E. Paquet}

\institute{
Ali Pesaranghader\textsuperscript{ \Letter, 1} \and \email{apesaran@uottawa.ca} \\
Herna L Viktor\textsuperscript{1} \and \email{hviktor@uottawa.ca} \\
Eric Paquet\textsuperscript{1,2} \and \email{eric-paquet@nrc-cnrc.gc.ca} \\
\textsuperscript{1} School of Electrical Engineering and Computer Science,  University of Ottawa, Ottawa, ON K1N 6N5, Canada. 
\\
\textsuperscript{2} National Research Council (NRC) of Canada, 1200 Montreal Road, Ottawa, ON K1A 0R6, Canada.
}

\date{Received: date / Accepted: date}

\maketitle

\begin{abstract}
The last decade has seen a surge of interest in adaptive learning algorithms for data stream classification, with applications ranging from predicting ozone level peaks, learning stock market indicators, to detecting computer security violations. In addition, a number of methods have been developed to detect concept drifts in these streams. Consider a scenario where we have a number of classifiers with diverse learning styles and different drift detectors. Intuitively, the current `best' (classifier, detector) pair is application dependent and may change as a result of the stream evolution. Our research builds on this observation. We introduce the \textsc{Tornado} framework that implements a reservoir of diverse classifiers, together with a variety of drift detection algorithms. In our framework, all (classifier, detector) pairs proceed, in parallel, to construct models against the evolving data streams. At any point in time, we select the pair which currently yields the best performance. We further incorporate two novel stacking-based drift detection methods, namely the FHDDMS and FHDDMS\textsubscript{add} approaches. The experimental evaluation confirms that the current `best' (classifier, detector) pair is not only heavily dependent on the characteristics of the stream, but also that this selection evolves as the stream flows. Further, our FHDDMS variants detect concept drifts accurately in a timely fashion while outperforming the state-of-the-art.  
\end{abstract}

\section{Introduction}
\label{sec_introduction}

The last decade has seen a rapid increase in the amount of massive, rapidly evolving data streams. Today's decision makers require new solutions to comprehend these fast-evolving knowledge sources in near real-time. That is, they need near-instant models to aid them to detect traffic congestion, to analyze smart phone usage patterns, to track the trends in the online sales of merchandise, for mobile crowd sensing, or to trace the spread of ideas, opinions and movements in social networks. This fact has resulted in a surge of interest in adaptive learning algorithms for data stream classification, that are able to learn incrementally and to rapidly adapt to changes in the data (so-called concept drift) \citep{vzliobaite2016overview}. Such approaches take into consideration that the learning environment is non-stationary and, as a result, build models that evolve over time, as the data arrive. 

A number of incremental learners, such as the Hoeffding Tree (HT)\footnote{It is also known as Very Fast Decision Tree (VFDT) in the literature.} \citep{domingos2000mining}, Naive Bayes, Perceptron \citep{bifet2010fast, freund1999large} and K-Nearest Neighbors (K-NN) methods have been developed, in order to learn from data streams. Intuitively, the performance of an individual classifier may vary as a stream evolves. Also, the difference in learning styles may cause a specific classifier to excel against one stream, while failing to accurately model another. Further, no single concept drift detection technique outperforms others in all settings. Rather, the current `best' pair of classifier and drift detector also changes, as the stream evolves. 

Based on these observations, we introduce the \textsc{Tornado} framework. In our framework a reservoir of classifiers with diverse learning styles co-exists, together with a number of different drift detector algorithms. These classifiers learn incrementally, in parallel. The \textsc{Tornado} framework operates as follows. Each of the classifiers in the reservoir incrementally accepts the incoming instances, one at a time, and proceeds to build a model. Each classifier is combined with each one of the drift detectors. That is, classifier $C_1$ is combined with drift detectors $D_1, D_2, ..., D_m$, to form pairs ($C_1, D_1$), ($C_1, D_2$), ..., ($C_1, D_m$), classifier $C_2$ is combined with drift detectors $D_1, D_2, ..., D_m$, to form pairs ($C_2, D_1$), ($C_2, D_2$), ..., ($C_2, D_m$), and so on. Our CAR measure, as introduced in Section \ref{sec_car_measure}, is used in order to rank the current best performing (classifier, detector) pair. The CAR measure not only considers the classification error-rate, but also takes into consideration the memory usage, runtime as well as the drift detection delay, together with the number of false positives and false negatives. 

In addition, we also incorporate two new drift detection methods into the \textsc{Tornado} framework. They extend the FHDDM algorithm, as introduced in \citep{pesaranghader2016fast}, in two ways. Firstly, we introduce the FHDDMS algorithm that creates a so-called ``stack'' of sliding windows of different sizes. The windows monitor the streams using bitmaps and alarm for concept drift using threshold values. The intuition behind this approach is that, by utilizing the windows of various sizes to monitor the stream, concept drift is detected faster and more accurately. In addition, we present FHDDMS\textsubscript{add}, a variant of FHDDMS, that employs data summaries, instead of bitwise operations.

Our experimental evaluation against synthetic and real-world data streams confirms that the current best (classifier, drift detector) pair evolves as the characteristics of the stream changes. In addition, our FHDDMS methods  detect changes faster and more accurate, with shorter delays, fewer false positives and false negatives, when compared to the state-of-the-art. 

This paper is organized as follows. Section \ref{sec_related_work} discusses related work. In Section \ref{sec_car_measure}, we introduce our CAR measure. This is followed, in Section \ref{sec_tornado_framework}, by an overview of the \textsc{Tornado} framework. Section \ref{sec_fhddm_fhddms} presents the FHDDMS and FHDDMS\textsubscript{add} algorithms. In Section \ref{sec_experiments}, we detail our experimental setup and results. Section \ref{sec_discussion} provides a detailed discussion regarding our experiments. Section \ref{sec_con_fut} concludes the paper and highlights future work.

\section{Related Work}
\label{sec_related_work}

This section discusses related work on performance measures for data stream mining, by focusing on classification and adaptation measures in a streaming setting, which we use as a foundation for defining the CAR measure in Section 3. In addition, we review the state-of-the-art in terms of drift detection algorithms.

\subsection{Performance Measures for Adaptive Online Learning}

Researchers agree that the evaluation of data stream algorithms is a complex task. This fact is due to many challenges, including the presence of concept drift, limited processing time in real-world applications and the need for time-oriented evaluation, amongst others \citep{gama2004learning}.
The error-rate (or accuracy) is most often used as the defining measures of the classification performance for evaluating learning algorithms in most streaming studies \citep{hulten2001mining, gama2004learning, gama2006decision, bifet2007learning, huang2015drift,baena2006early}. The error-rate is calculated incrementally using either the prequential or hold-out evaluation procedures \citep{bifet2009data}.
The interplay between the error-rate and other factors, such as memory usage and runtime considerations, has received limited attention. \cite{bifet2009new} considered the memory, time and accuracy measures separately, in order to compare the performances of ensembles of classifiers. \cite{bifet2010fast} further introduced the RAM-Hour measure, where every RAM-Hour equals to 1 GB of RAM occupied for one hour, to compare the performances of three versions of perceptron-based Hoeffding Trees. \cite{pesaranghader2016framework} introduced the EMR measure which combines error-rate, memory usage and runtime for evaluating and ranking learning algorithms. \newline

\cite{vzliobaite2015towards} introduced the return on investment (ROI) measure to determine whether the {\it adaptation} of a learning algorithm is beneficial. They concluded that adaptation should only take place if the expected gain in performance, measured by accuracy, exceeds the cost of other resources (e.g. memory and time) required for adaptation. In their work, the ROI measure was used to indicate whether an adaptation to a concept drift is beneficial, over time. \cite{olorunnimbe2015intelligent} extended the above-mentioned ROI measure, in order to dynamically adapt the number of base learners in online bagging ensembles.
\cite{pesaranghader2016fast} proposed an approach to count true positive (TP), false positive (FP), and false negative (FN) of drift detection, in order to evaluate the performances of concept drift detectors. They introduced the acceptable delay length notion as a threshold that determines how far a detected drift could be from the real location of drift to be considered as a true positive. \newline

\par As explained above, the performance measures of classification and adaptation have been often used, separately, to evaluate adaptive learning algorithms against evolving data streams. To date, no single measure that considers classification, adaptation, and resource consumption together, has been developed. Such a measure would allow one to assess the ``big picture'', in terms of the costs and benefits of a specific learning and adaptation strategy. In Section \ref{sec_car_measure}, we introduce the CAR measure in order to address this deficiency. 

\subsection{Drift Detection Methods}
\label{subsec_ddms}

\cite{gama2014survey} categorized concept drift detectors into three general groups, as follows:

\begin{enumerate}

\item \textit{Sequential Analysis based Methods} sequentially evaluate prediction results as they become available, and alarm for drifts when a pre-defined threshold is met. The Cumulative Sum (CUSUM) and its variant Page-Hinkley (PH) \citep{page1954continuous}, as well as Geometric Moving Average (GMA) \citep{roberts2000control} are members of this group. 

\item \textit{Statistical based Approaches} probe the statistical parameters such as mean and standard deviation of prediction results to detect drifts in a stream. The Drift Detection Method (DDM) \citep{gama2004learning}, Early Drift Detection Method (EDDM) \citep{baena2006early} and Exponentially Weighted Moving Average (EWMA) \citep{ross2012exponentially} are members of this group. 

\item \textit{Windows based Methods} usually use a fixed reference window summarizing the past information and a sliding window summarizing the most recent information. A significant difference between the distributions of these two windows suggests the occurrence of a drift. Statistical tests or mathematical inequalities, with the null-hypothesis indicating that the distributions are equal, are thus employed. Kifer's \citep{kifer2004detecting}, Nishida's \citep{nishida2007detecting}, Bach's \citep{bach2008paired}, the Adaptive Windowing (ADWIN) \citep{bifet2007learning}, SeqDrift detectors \citep{sakthithasan2013one, pears2014detecting}, Drift Detection Methods based on Hoeffding's Bound (HDDM\textsubscript{A-test} and HDDM\textsubscript{W-test}) \citep{frias2015online}, and Adaptive Cumulative Windows Model (ACWM) \citep{sebastiaofading} are members of this family.

\end{enumerate} 

CUSUM and its variant Page-Hinkley (PH) are some of the pioneer methods in the community. DDM, EDDM, and ADWIN have frequently been considered as benchmarks in the literature \citep{huang2015drift, frias2015online, baena2006early, nishida2007detecting, bifet2007learning, pesaranghader2016fast}. SeqDrift2 and HDDMs are recently proposed methods, and have shown comparable results to the other benchmarks. We, therefore, consider all these methods for our experimental evaluation, and we briefly describe them as follows.

\begin{description}
	\item[CUSUM: Cumulative Sum --] CUSUM, by \cite{page1954continuous}, is a sequential analysis technique that alarms for a change when the mean of the input data significantly deviates from zero. The input of CUSUM can be any filter residual; for instance, the prediction error from a Kalman filter \citep{gama2014survey}. The CUSUM test is in the form of $g_t = max(0, g_{t-1} + (x_t - \delta))$, and it alarms for a concept drift when $g_t > \lambda$. In this test, $x_t$ is the currently observed value, $\delta$ specifies the magnitude of changes that are allowed, while $g_0 = 0$ and $\lambda$ is a user-defined threshold. The accuracy of CUSUM depends on the values of parameters $\delta$ and $\lambda$. Lower values of $\delta$ result in faster detection, at the cost of an increased number of false alarms.
	
	\item[PH: Page-Hinkley --] PH, by \cite{page1954continuous}, is a variant of CUSUM typically used for change detection in signal processing applications \citep{gama2014survey}. The test variable $m_T$ is defined as a cumulative difference between the observed values and their mean until the current time $T$; and calculated by $m_T = \sum_{t=1}^{T}(x_t - \bar{x}_T - \delta)$, where $\bar{x} = \frac{1}{T}\sum_{t = 1}^{T}x_t$ and $\delta$ defines the allowed magnitude of changes. The PH method also updates the minimum $m_T$, denoted as $M_T$, using $M_T = min(m_t, t= 1 ... T)$. A significant difference between $m_T$ and $M_T$, i.e.\ $PH_T:m_T - M_T > \lambda$ where $\lambda$ is a user-defined threshold, implies a concept drift. A large value of $\lambda$ typically causes fewer false alarms, but it may increase false negative rate.

	\item[DDM: Drift Detection Method --] DDM, by \cite{gama2004learning}, monitors the error-rate of the classification model to detect drifts. On the basis of PAC learning model \citep{mitchell1997machine}, the method considers that the error-rate of a classifier decreases or stays constant as the number of instances increases. Otherwise, it suggests the occurrence of a drift. Consider $p_t$ as the error-rate of the classifier with a standard deviation of $s_t = \sqrt{(p_t(1 - p_t) / t)}$ at time $t$. As instances are processed, DDM updates two variables $p_{min}$ and $s_{min}$ when $p_t + s_t < p_{min} + s_{min}$. DDM warns for a drift when $p_t + s_t \ge p_{min} + 2 * s_{min}$, and it detects a drift when $p_t + s_t \ge p_{min} + 3 * s_{min}$. The $p_{min}$ and $s_{min}$ are reset when a drift is detected.
	\item[EDDM: Early Drift Detection Method --] EDDM, by \cite{baena2006early}, evaluates the distances between wrong predictions to detect concept drifts. The algorithm is based on the observation that a drift is more likely to occur when the distances between errors are smaller. EDDM calculates the average distance between two recent errors, i.e.\ $p'_t$, with its standard deviation $s'_t$ at time $t$. It updates two variables $p'_{max}$ and $s'_{max}$ when $p'_t +2*s'_t > p'_{max} + 2*s'_{max}$. The method warns for a drift when $(p'_t + 2 * s'_t) / (p'_{max} + 2*s'_{max}) < \alpha$, and indicates that a drift occurred when  $(p'_t + 2 * s'_t) / (p'_{max} + 2*s'_{max})  < \beta$. The authors set $\alpha$ and $\beta$ to 0.95 and 0.90, respectively. The $p'_{max}$ and $s'_{max}$ are reset only a drift is detected.
	\item [HDDMs --] HDDM\textsubscript{A-test} and HDDM\textsubscript{W-test} are proposed by \cite{frias2015online}. The former compares the moving averages to detect drifts. The latter uses the EMWA forgetting scheme \citep{ross2012exponentially} to weight the moving averages. Then, weighted moving averages are compared to detect concept drifts. For both cases, the Hoeffding's inequality \citep{hoeffding1963probability} is used to set an upper bound to the level of difference between averages. The authors noted that the first and the second methods are ideal for detecting abrupt and gradual drifts, respectively.
	\item[ADWIN: Adaptive Sliding Window --] ADWIN, by Bifet \cite{bifet2007learning}, slides a window $w$ as the predictions become available, in order to detect drifts. The method examines two sub-windows of sufficient width, i.e.\ $w_0$ with size $n_0$ and $w_1$ with size $n_1$, of $w$, where $w_0 \cdot w_1 = w$. A significant difference between the means of two sub-windows indicates a concept drift, i.e.\ $|\hat{\mu}_{w_0} - \hat{\mu}_{w_1}| \ge \varepsilon$ where $\varepsilon = \sqrt{\frac{1}{2m}\ln{\frac{4}{\delta'}}}$, $m$ is the harmonic mean of $n_0$ and $n_1$, $\delta' = \delta/n$. Here $\delta$ is the confidence level while $n$ is the size of window $w$. After a drift is detected, elements are removed from the tail of the window until no significant difference is seen.
	\item[SeqDrift2 --] SeqDrift2, by \cite{pears2014detecting}, uses the reservoir sampling method \citep{vitter1985random}, as an adaptive sampling strategy, for random sampling from input data. SeqDrift2 stores entries into two repositories called \textit{left} and \textit{right}. As entries are processed over time, the left repository contains a combination of older and new entries by applying the reservoir sampling strategy. The right repository collects the new arriving entries. SeqDrift2 subsequently finds an upper bound for the difference in between the means of the two repositories, i.e.\ $\hat{\mu}_l$ for the left repository and $\hat{\mu}_r$ for the right repository, using the Bernstein inequality \citep{bernstein1946theory}. Finally, a significant difference between the two means suggests a concept drift. 
	
\end{description}

\noindent \textit{Discussion --}
CUSUM and Page-Hinkley (PH) detect concept drift by calculating the difference of observed values from the mean and alarm for a drift when this value is larger than a user-defined threshold. These algorithms are sensitive to the parameter values, resulting a tradeoff between false alarms and detecting true drifts.
Recall that DDM, EDDM and HDDM maintain sets of variables, in order to monitor a stream for concept drift. The ADWIN and SeqDrift2 methods, on the other the hand, maintain more than one subset of the stream, either using windowing or repositories.
DDM and EDDM have lower memory footprints as they only maintain a small number of variables \citep{gama2014survey}. These two approaches also require less execution runtime to update the values of the variables for drift detection. However, EDDM may frequently alarm for concept drift during the early stages of learning, since the distance between wrong predictions is small.  
HDDM employs the Hoeffding's inequality in order to detect concept drift.
ADWIN and SeqDrift2 generally require more memory for storing prediction results, as maintained within sliding windows or repositories. 
They are also computationally more expensive, due to the sub-window compression or reservoir sampling procedures. 
Recall that the SeqDrift2 algorithm of \cite{pears2014detecting} employs the Bernstein inequality in order to detect concept drift. SeqDrift2 uses the sample variance, and assumes that the sampled data follow a normal distribution. It follows that that this assumption may be too restrictive, in real-world domains. Further, the Bernstein's inequality is conservative and requires a variance parameter, in contrast to, for instance, the Hoeffding's inequality. These shortcomings may lead to longer detection delays and a potential loss of accuracy. 
In summary, our preliminary experimentation confirmed that the aforementioned methods may cause long detection delay, high false positives as well as high false negatives. In Section \ref{sec_fhddm_fhddms}, we will introduce our new Stacking Fast Hoeffding Drift Detection Method (FHDDMS), that extends our earlier introduced Fast  Hoeffding Drift Detection Method (FHDDM) technique \citep{pesaranghader2016fast}.
FHDDM slides a window over the stream, in order to detect concept drift. We maintain two variables, namely the mean of elements inside the window at the current time and the maximum mean observed so far.
FHDDM subsequently employs the Hoeffding's inequality to detect drifts. Our approach thus differs from HDDM, in that we use a sliding window and only maintain two variables. 


\section{The CAR Performance Measure}
\label{sec_car_measure}

\begin{figure}[h]
	\begin{center}
		\includegraphics[scale=0.35]{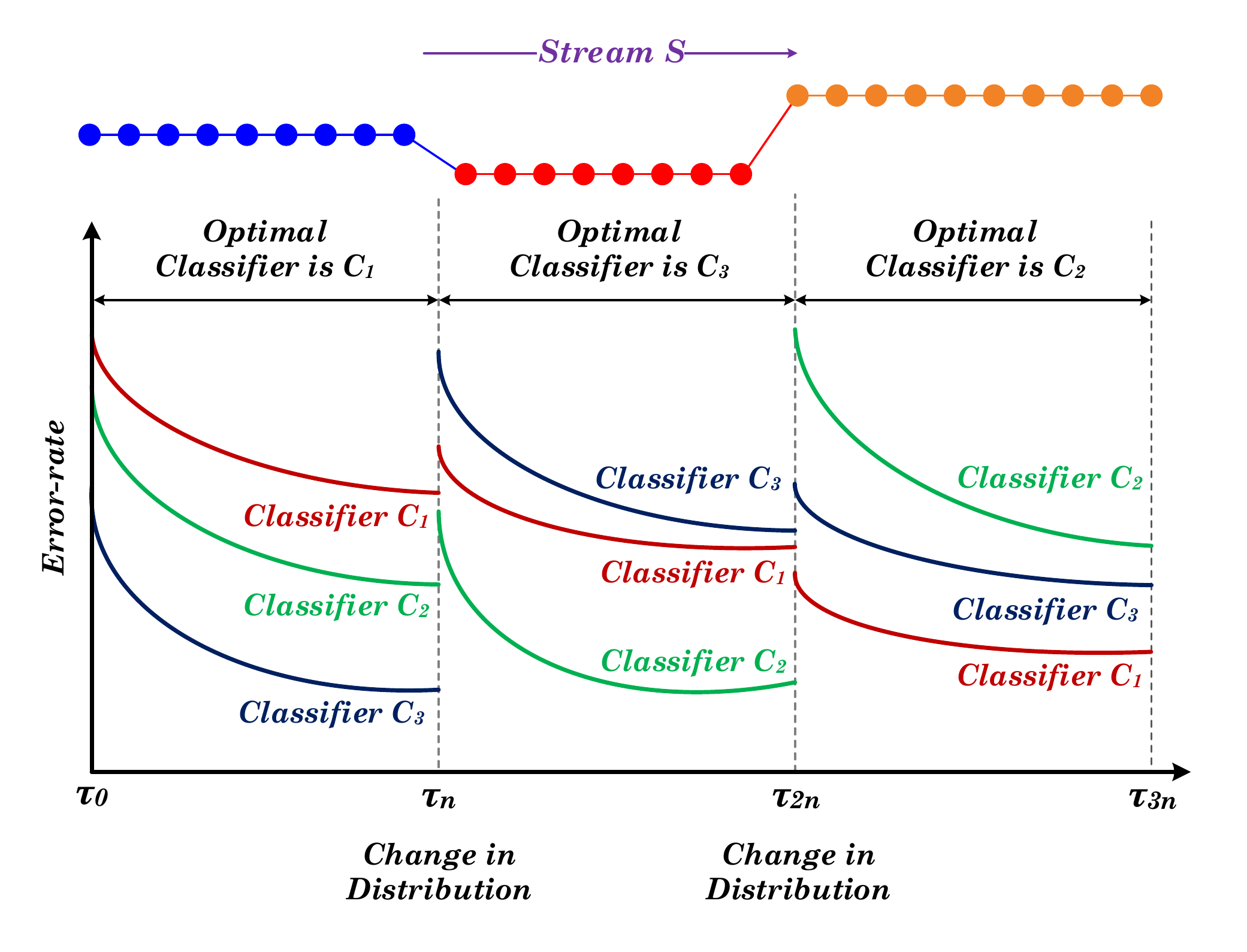}
		\caption{Illustration of Error-rate and Distributional Change Interplay}
		\label{fig_error_rates}
	\end{center}
\end{figure}

This section presents our CAR measure, which is employed in order to balance classification, adaptation and resource utilization requirements. The motivation for introducing this measure is as follows.
Intuitively, as illustrated in Fig.\ \ref{fig_error_rates}, a change in the data distribution, as caused by a concept drift, may result in an increase or decrease of the error-rates for different types of classifiers \citep{olorunnimbe2015intelligent}. We simulated a number of drift points and showed that, as a concept drift occurs, the classifier with the lowest error-rate changes. Consequently, it follows that a learning system where different types of classifiers co-exist and where the model, from the current ``best'' learner is provided to the users, may hold much value.

However, following an ``\textit{error-rate-only}'' approach is not beneficial in all settings. For instance, in an emergency response context, the response time, i.e. the time required to present a model to the users, may be the most important criterion. That is, users may be willing to sacrifice accuracy for speed and partial information. Further, consider the area of pocket (or mobile) data mining, which has much application in areas such as defense and environmental impact assessment \citep{gaber2014pocket}. Here, the memory resources may be limited, due to connection issues, and thus reducing the memory footprint is also of importance \citep{olorunnimbe2015intelligent}.

To address this challenge, the EMR measure was proposed for evaluating the overall performance of learning algorithms based on Error-rate, Memory usage, and Runtime \citep{pesaranghader2016framework}. In this approach, classifiers receive a score based on its current EMR measure value which is then used to rank them. The EMR measure does not fully reflect the performance of learning methods in an environment with concept drift.
For instance, consider medical applications, where an adaptive learning algorithm must handle concept drifts very rapidly. A scenario where an airplane is on autopilot is another application where adaptive algorithms, that frequently alarm falsely for concept drifts, are not considered suitable. In such contexts, the EMR measure does not effectively represent the overall performance of adaptive learning algorithms because the detection delay as well as the number of false alarms are both ignored. Therefore, we need to integrate into a single measure the performances of the classifiers, their adaptability as well as the resources allocated.

We introduce the CAR measure which not only considers the classification error-rates, memory usages, and runtimes but also the drift detection delays, false positives and false negatives. The CAR measure, as defined in Equation \eqref{equ_21}, consists of three components namely \textit{Classification}, \textit{Adaptation}, and \textit{Resource Consumption}. The classification part consists of the error-rate (E\textsubscript{C}) of classifier $C$, the adaptation part represents the detection delay (D\textsubscript{D}), false positive (FP\textsubscript{D}) and false negative (FN\textsubscript{D}) of drift detector $D$, while the resource consumption part is associated with the memory consumption (M\textsubscript{(C,D)}) and runtime (R\textsubscript{(C,D)}) of the (classifier, detector) pairs. Please note that, in Equation \eqref{equ_21}, the $\diamond$ and $\oplus$ symbols only represent the combination of three components.

\vspace{-1em}

\begin{align}
	\label{equ_21}
	\mbox{CAR}_{(C, D)} & := \mbox{Classification}_{C} \diamond \mbox{Adaptation}_{D} \diamond \mbox{Resourse Use}_{(C, D)} \nonumber \\
	 &  := {E}_{C} \diamond ({D}_{D} \oplus {FP}_{D} \oplus {FN}_{D}) \diamond (M_{(C, D)} \oplus R_{(C, D)}) 
\end{align}

The score associated with  the pair $(C, D)$ is obtained from Equation \eqref{equ_22}. The equation implies that a pair with a high CAR has a low score, i.e.

\begin{equation}
	\label{equ_22}
	\mbox{Score}_{(C, D)} := 1 - \mbox{CAR}_{(C, D)}
\end{equation}

In order to compute the CAR measure, a matrix containing all results of the classification, adaptation, and resource consumption of (classifier, detector) pairs is created each time an instance is processed.
There are $n$ classifiers and $m$ drift detectors, which means that \ $n \times m$ pairs, are considered concurrently. The measures associated with each pair for the $t^{th}$ instance are placed, row-by-row, into a matrix $M^t$, as shown in Equation \eqref{equ_23}. This matrix is defined as follows:

\begin{equation}
	\label{equ_23}
	M^t = \begin{bmatrix}
		E_{C_1}^t & D_{D_1}^t & FP_{D_1}^t & FN_{D_1}^t & (M_{C_1}^t + M_{D_1}^t) & (R_{C_1}^t + R_{D_1}^t)  \\
		E_{C_1}^t & D_{D_2}^t & FP_{D_2}^t & FN_{D_2}^t & (M_{C_1}^t + M_{D_2}^t) & (R_{C_1}^t + R_{D_2}^t)  \\
		E_{C_1}^t & D_{D_3}^t & FP_{D_3}^t & FN_{D_3}^t & (M_{C_1}^t + M_{D_3}^t) & (R_{C_1}^t + R_{D_3}^t)  \\
		\hdotsfor{6} \\
		\hdotsfor{6} \\
		E_{C_n}^t & D_{D_m}^t & FP_{D_m}^t & FN_{D_m}^t & (M_{C_n}^t + M_{D_m}^t) & (R_{C_n}^t + R_{D_m}^t)  \\
\end{bmatrix} \end{equation}

\noindent Subsequently, the elements of the matrix are normalized, column-by-column, using the `min-max' scaling approach. The resulting normalized matrix, $\overline{M^t}$, is defined in Equation \eqref{equ_24}:

\begin{equation}
	\label{equ_24}
	\overline{M^t} = \begin{bmatrix}
		\overline{E_{C_1}^t} & \overline{D_{D_1}^t} & \overline{FP_{D_1}^t} & \overline{FN_{D_1}^t} & (\overline{M_{C_1}^t + M_{D_1}^t}) & (\overline{R_{C_1}^t + R_{D_1}^t})  \\
		\overline{E_{C_1}^t} & \overline{D_{D_2}^t} & \overline{FP_{D_2}^t} & \overline{FN_{D_2}^t} & (\overline{M_{C_1}^t + M_{D_2}^t}) & (\overline{R_{C_1}^t + R_{D_2}^t})  \\
		\overline{E_{C_1}^t} & \overline{D_{D_3}^t} & \overline{FP_{D_3}^t} & \overline{FN_{D_3}^t} & (\overline{M_{C_1}^t + M_{D_3}^t}) & (\overline{R_{C_1}^t + R_{D_3}^t})  \\
		\hdotsfor{6} \\
		\hdotsfor{6} \\
		\overline{E_{C_n}^t} & \overline{D_{D_n}^t} & \overline{FP_{D_n}^t} & \overline{FN_{D_n}^t} & (\overline{M_{C_n}^t + M_{D_n}^t}) & (\overline{R_{C_n}^t + R_{D_n}^t})  \\
\end{bmatrix} \end{equation}

A weight is associated to each measure. The weights are combined into a weight vector $\overrightarrow{w}$ which is defined in Equation \eqref{equ_25}. The elements of the weight vector $\overrightarrow{w}$ are the weights associated with the classification error-rate, the detection delay, the false positive rate, the false negative rate, the memory usage, and the runtime. Each weight emphasizes the importance of a particular measure in the evaluation process.

\begin{equation}
	\label{equ_25}
	\overrightarrow{w} = \begin{bmatrix} w_e & w_d & w_{fp} & w_{fn} & w_{m} & w_{r} \end{bmatrix}^T
\end{equation}

The CAR measures and scores for the $n \times m$ pairs are evaluated with Equations \eqref{equ_26} and \eqref{equ_27}, respectively. Please note that $J_{n\times m, 1}$ is a vector that solely consists of unit entries.

\begin{equation}
	\label{equ_26}
	\mbox{CAR}_{n\times m, 1}^t = \frac{\overline{M^t} \cdot \overrightarrow{w}}{J_{1, 6} \cdot \overrightarrow{w}}
\end{equation}
\begin{equation}
	\label{equ_27}
	\mbox{Score}_{n\times m, 1}^t = J_{n\times m, 1} -  \mbox{CAR}_{n\times m, 1}^t
\end{equation}

The index of the classifier, that is recommended at time $t$, \textit{Index\textsubscript{opt}}, is defined by Equation \eqref{equ_28}\footnote{The $imax$ is a function that finds the index of the pair presenting the highest score.}:

\begin{equation}
	\label{equ_28}
	\mbox{\textit{Index\textsubscript{opt}}} = imax(Score_{n\times m, 1}^t)
\end{equation}

One should notice that the weights are application dependent. For instance, if the memory resources are limited, such as in the case of a pocket data mining scenario \citep{gaber2014pocket}, the value of \textit{w\textsubscript{m}} should be set to a higher value. On the other hand, if memory is abundant, but accuracy and speed of model construction are important, the \textit{w\textsubscript{m}} value may be decreased (or even set to zero). In medical applications, where reacting rapidly to concept drifts is critical, $w_d$ may be the dominant weight.


\section{\textsc{Tornado:} A Reservoir of Diverse Learning Strategies}
\label{sec_tornado_framework}

\begin{figure}[h]
	\begin{center}
		\includegraphics[scale=0.475]{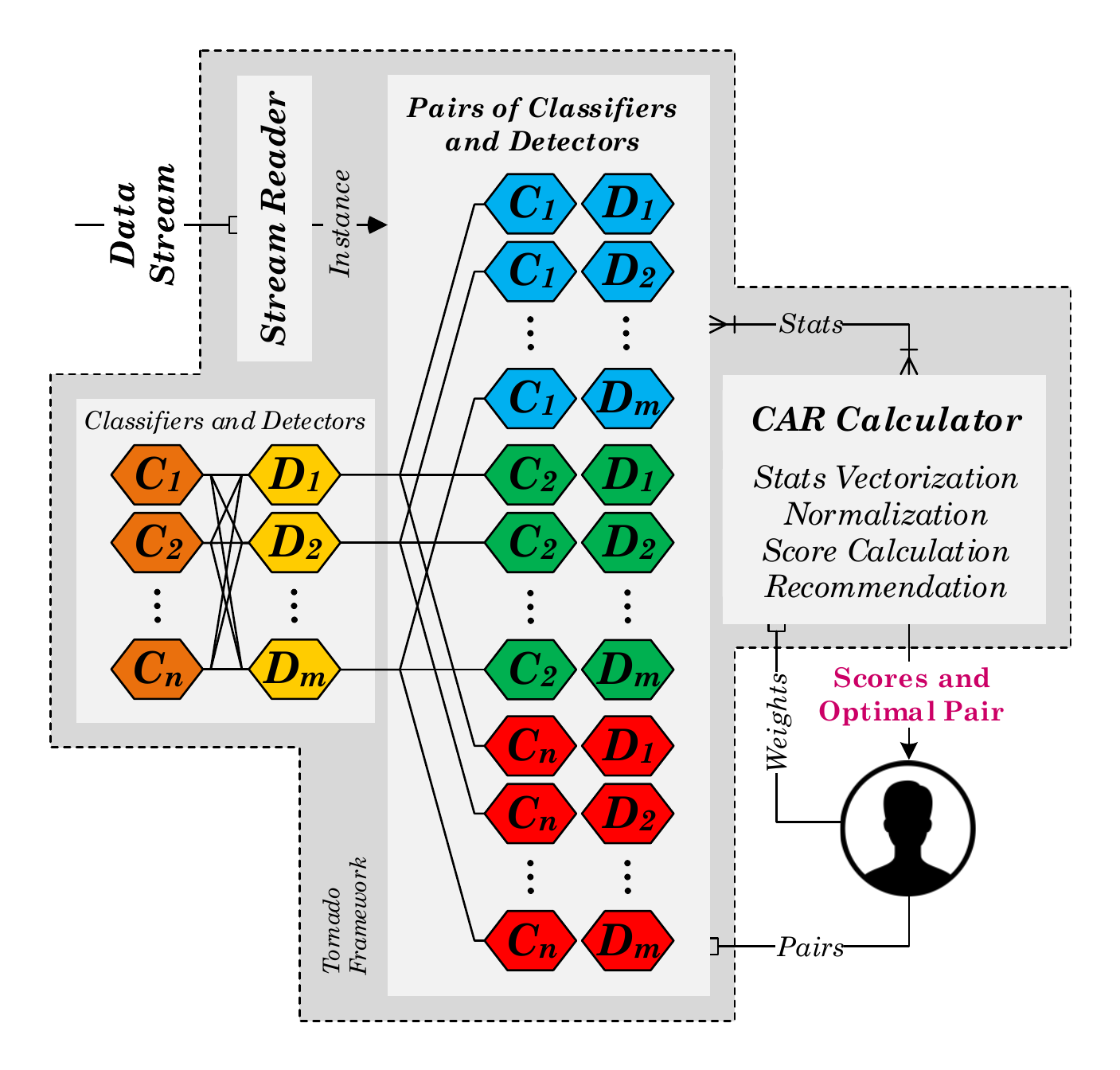}
		\caption{The \textsc{Tornado} Framework}
		\label{fig_tornado}
	\end{center}
\end{figure}

In this section, we introduce the \textsc{Tornado} framework which is outlined in Fig.\ \ref{fig_tornado}. We implemented our framework using the Python programming language. Recall that, in our framework, a number of distinct pairs of classifiers and drift detectors are executed in \textit{parallel}, against the same data stream. In this figure, $C_n$ and $D_m$ represent the $n^{th}$ classifier and the $m^{th}$ detector, respectively. 
A number of classifiers with different learning styles are implemented. Currently, the Naive Bayes (NB), Decision Stump (DS), Hoeffding Tree (HT) \citep{domingos2000mining}, Perceptron (PR) \citep{bifet2010fast}, 
and K-Nearest Neighbors (K-NN) learning algorithms are available. Furthermore, various concept drift detection methods, based on statistical or window-based approaches, are provided.  Specifically, we have implemented Cumulative Sum (CUSUM) and its variant Page-Hinkley (PH) \citep{page1954continuous}, Drift Detection Method (DDM) \citep{gama2004learning}, Early Drift Detection Method (EDDM) \citep{baena2006early}, Hoeffding's bound based Drift Detection Methods (HDDM\textsubscript{A-test} and HDDM\textsubscript{W-test}) \citep{frias2015online}, Adaptive Windowing (ADWIN) \citep{bifet2007learning}, SeqDrift2 \citep{pears2014detecting}, Fast Hoeffding Drift Detection Method (FHDDM) \citep{pesaranghader2016fast}, and our new Stacking Fast Hoeffding Drift Detection Methods (FHDDMS), which is introduced in Section \ref{sec_fhddm_fhddms}. 

As shown in Fig.\ \ref{fig_tornado}, \textsc{Stream Reader}, \textsc{Classifiers} and \textsc{Detectors}, \textsc{Pairs of Detectors and Detectors}, and \textsc{CAR Calculator} are the main components of the framework. The input is constituted of a Stream, pairs of classifiers and detectors, and a  weight vector. Our framework follows the \textit{prequential} approach where instances are first tested and then used for training \citep{gama2013evaluating, gama2014survey, pesaranghader2016fast}.

The data flow may be described as follows: The (classifier, detector) pairs are constructed as shown in Fig.\ \ref{fig_tornado}, prior to the learning process. The \textsc{Stream Reader} reads instances from the stream and sends them one-by-one to the (classifier, detector) pairs for model construction. Each learner builds an incremental model, prequentially. That is, each instance is first used for testing and then for training. Simultaneously, \textsc{Classifier}s send their statistics, e.g.\ error-rates or the current prediction results, to their corresponding \textsc{Drift Detector} in order to detect potential occurrences of concept drifts. Subsequently, the \textsc{CAR Calculator} determines the score of each (classifier, detector) pair by considering the classification error-rate, detection delay, detection false positive rate, detection false negative rate, total memory usage and runtime. Subsequently, the model with the highest score is presented to the user. This model may change as a result of incremental learning and concept drift. This process continues until either a predefined condition is met or all the instances in the stream are processed.

\begin{figure}[h]
	\begin{center}
		\includegraphics[scale=0.35]{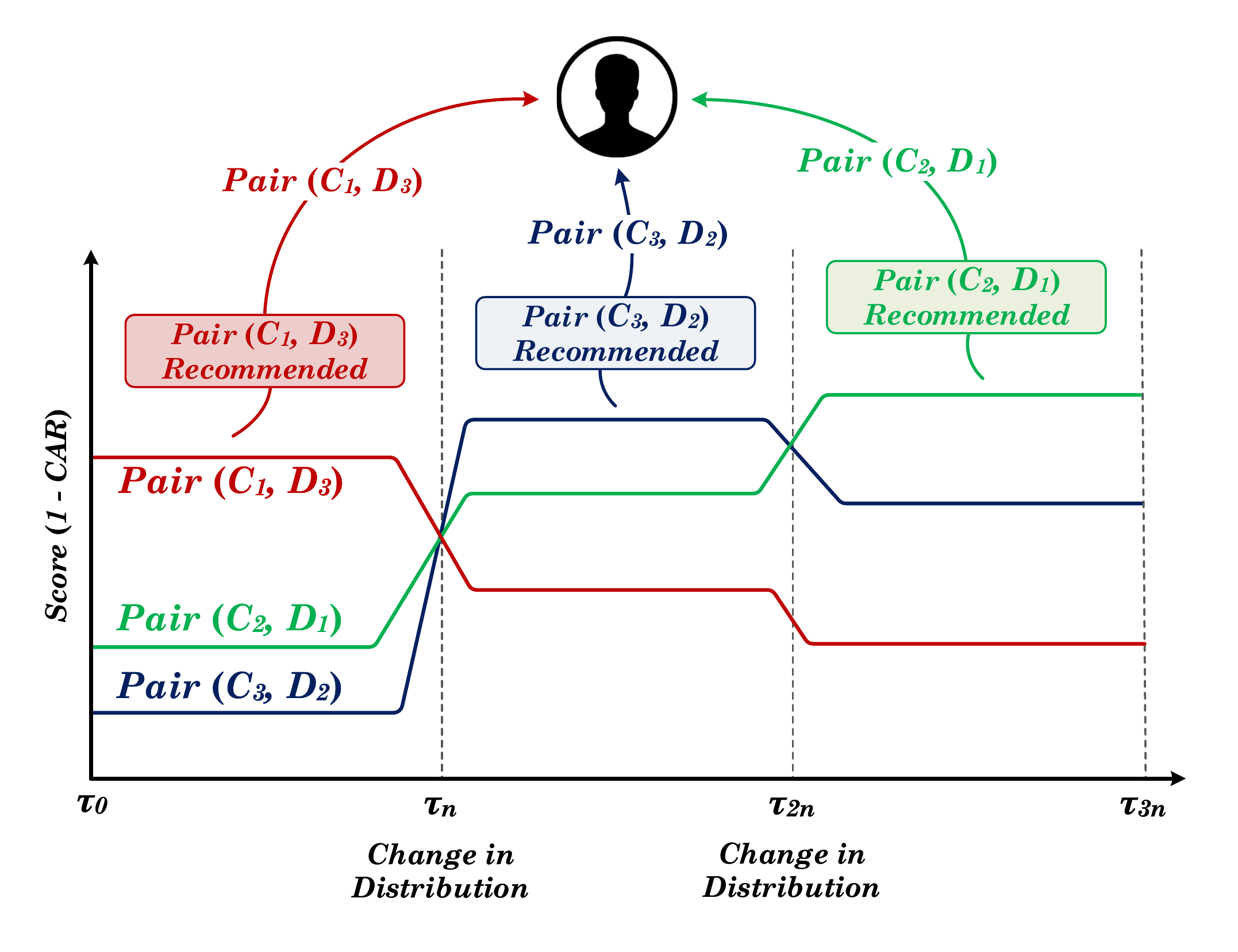}
		\caption{Recommendation of (Classifier, Detector) Pairs over Time}
		\label{fig_pair_recom}
	\end{center}
\end{figure}

Fig.\ \ref{fig_pair_recom} illustrates that, while the various pairs are executed concurrently, the one with the best score is recommended at each time interval. An interval is the time difference in between two consecutive concept drifts. 
As illustrated by the figure, during the interval $\tau_0$ to $\tau_n$, the pair $(C_1, D_3)$ has the best score. Suddenly, at time $\tau_n$, the data distribution is altered resulting in pair $(C_3, D_2)$ being recommended to the user. In the illustrative example, another drift occurs at $\tau_{2n}$ resulting in pair $(C_2, D_1)$ having the highest score.

The following observation is noteworthy. 
Recall that the continuous outputs of our \textsc{Tornado} framework are the current best performing (classifier, detector) pair, which may change over time.
Consequently, our work should thus not be confused with a hybrid ensemble of classifiers setting \citep{Hsu2017Theoretical, min2011activity, verikas2010hybrid, salgado2006hybrid}. 
Typically, a hybrid ensemble contains a number of diverse classifiers that form a committee, which aims at increasing the predictive accuracy by utilizing the diversity of the members of the ensemble. 
In contrast, within the \textsc{Tornado} framework, the individual learners proceed independently to construct their models. Recall that the rationale behind our design is that we aim to utilize diverse learning strategies that potentially address concept drifts more efficiently. However, future work may also involve incorporating ensembles, as one of our classifiers, into the \textsc{Tornado} framework.


\section{Stacking Fast Hoeffding Drift Detection Methods}
\label{sec_fhddm_fhddms}

In this section, we briefly review the Fast Hoeffding Drift Detection Method (FHDDM). We subsequently introduce our Stacking Fast Hoeffding Drift Detection Method (FHDDMS) and Additive FHDDMS (FHDDMS\textsubscript{add}) algorithm.

\subsection{Fast Hoeffding Drift Detection Method (FHDDM)}
\label{subsec_fhddm}

Recently, \cite{pesaranghader2016fast} introduced the Fast Hoeffding Drift Detection Method (FHDDM) which is based on a sliding window mechanism and the Hoeffding's inequality. The FHDDM algorithm slides a window of size $n$ over the classification results.  A 1 is inserted in the window if a particular prediction is correct, while a 0 is inserted otherwise.  As the instances are processed, the mean associated with a particular sliding window at time $t$, $\mu^t$, is evaluated while the maximum mean observed so far, $\mu^m$, is updated if the mean of the current sliding window is higher.

On the basis of the probably approximately correct (PAC) learning model \citep{mitchell1997machine}, the classification accuracy either increases or remains constant as the number of instances increases \citep{gama2004learning}. Should this not be the case, the probability of a concept drift increases. As a result, the value of $\mu^m$ either increases or remains constant as instances are processed. Therefore, a concept drift is more likely if the value of $\mu^m$ remains approximately constant while the value of $\mu^t$ decreases over time. As demonstrated by \cite{pesaranghader2016fast}, if the difference in between the maximum and the current mean is greater than a certain threshold  $\varepsilon_d$, it may be safely assumed that a concept drift has occurred.  The threshold is evaluated with the Hoeffding's inequality. \newline

\noindent \textbf{Theorem I: Hoeffding's Inequality --} Let $X_1, X_2, ..., X_n$ be $n$ independent random variables bounded by the interval $[0, 1]$, then with a probability of at most $\delta$, the difference in between the empirical mean of these variables $\overline{X} = \frac{1}{n}\sum_{i=1}^{n}X_i$ and their expected values $E[\overline{X}]$  is at least $\varepsilon_H$, i.e. $Pr(|\overline{X} - E[\overline{X}]| \ge \varepsilon_H)  \le \delta$, where: 
\begin{equation}
\label{equ_43}
	\varepsilon_H = \sqrt{\frac{1}{2n}\ln{\frac{2}{\delta}}}
	\belowdisplayshortskip=15pt
\end{equation}
and $\delta$ is the upper bound for the probability.\newline

\noindent \textbf{\label{FHDDM_test}Corollary I: FHDDM test --} In a stream setting, assume $\mu^t$ is the mean of a sequence of $n$ random entries, where the prediction status of each instance is represented by a value in the set $\{0, 1\}$, at time $t$. Let $\mu^m$ is the maximum mean observed so far. Let $\Delta \mu = \mu^m - \mu^t \ge 0$ be the difference between the two means. Given the desired $\delta$, Hoeffding's inequality implies that a drift has occurred if $\Delta \mu \ge \varepsilon_d$, where:

\begin{equation}
\label{equ_44}
	\varepsilon_{d} = \sqrt{\frac{1}{2n}\ln{\frac{1}{\delta}}}
	\belowdisplayshortskip=15pt
\end{equation}

\begin{figure}[h]
	\centering
	\includegraphics[scale=0.55]{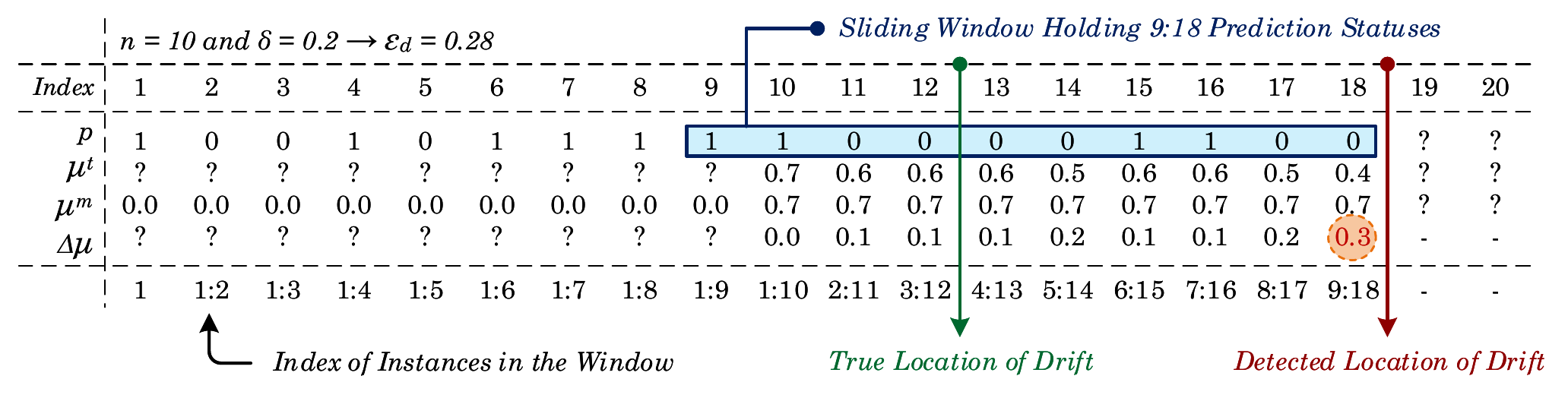}
	\caption{An Example of the FHDDM Algorithm}
	\label{fig_fhddm}
\end{figure}

Fig.\ \ref{fig_fhddm} illustrates the  FHDDM algorithm. In this example, $n$ and $\delta$ are set to 10 and 0.2, respectively. Using Corollary \hyperref[FHDDM_test]{I}, the value of $\varepsilon_d$ is equal to 0.28. Suppose that a real drift occurs right after the 12\textsuperscript{th} instance. The values of $\mu^t$ and $\mu^m$ are set to null and zero until 10 elements are inserted into the window. We have seven 1s in the window after reading the first 10 elements. Thus $\mu^t$ is equal to 0.7 and the value of $\mu^m$ is also set to 0.7. The 1\textsuperscript{st} element is removed from the window before the 11\textsuperscript{th} prediction status is inserted. Since the value of prediction status is 0, the value of $\mu^t$ decreases to 0.6 while the value of $\mu^m$ remains the same. This process continues until the 18\textsuperscript{th} instance is inserted. At this point in time, the difference between $\mu^m$ and $\mu^t$ exceeds $\varepsilon_d$. As a result, the FHDDM algorithm alarms for a drift.

\subsection{Sensitivity of FHDDM's Parameters}

In this section, we investigate the impact of the change in parameters  $\delta$ and $n$ on the value of $\varepsilon_d$. We also study the effects of varying these values on the detection delay, the false positive rate, the memory usage, and total runtime. To this end, we conducted a number of experiments with the values of $n$ in \{25, 100, 200, 300, 400, 500\} and $\delta$ in \{0.001, 0.0001, 0.00001, 0.000001, 0.0000001\}.  

\begin{figure}[h]
	\begin{center}
		\subfloat[Memory Usage\label{fig_fhddm_param_n_mem}]{\adjincludegraphics[scale=0.385,clip]{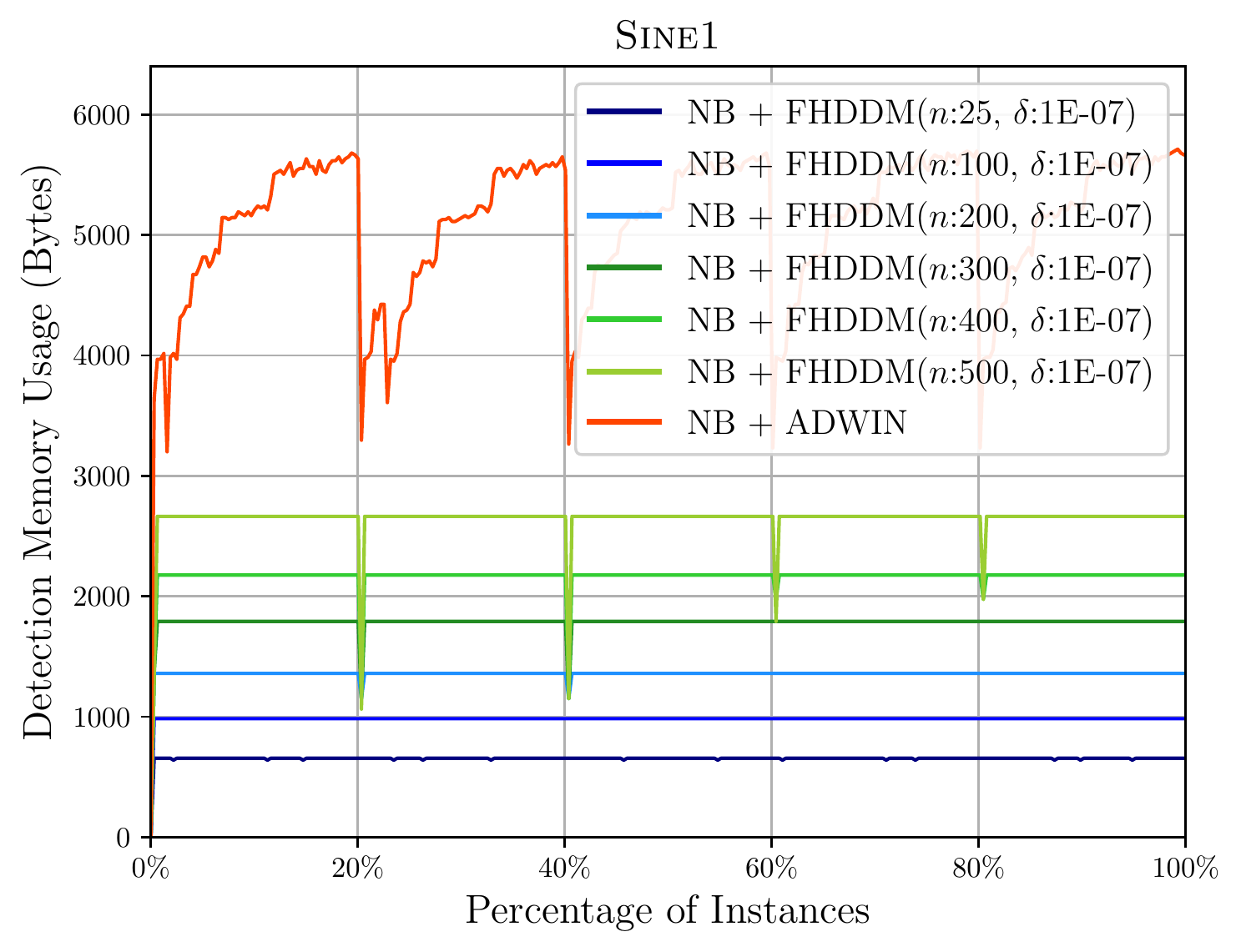}} \quad
		\subfloat[Runtime\label{fig_fhddm_param_n_time}]{\adjincludegraphics[scale=0.385,clip]{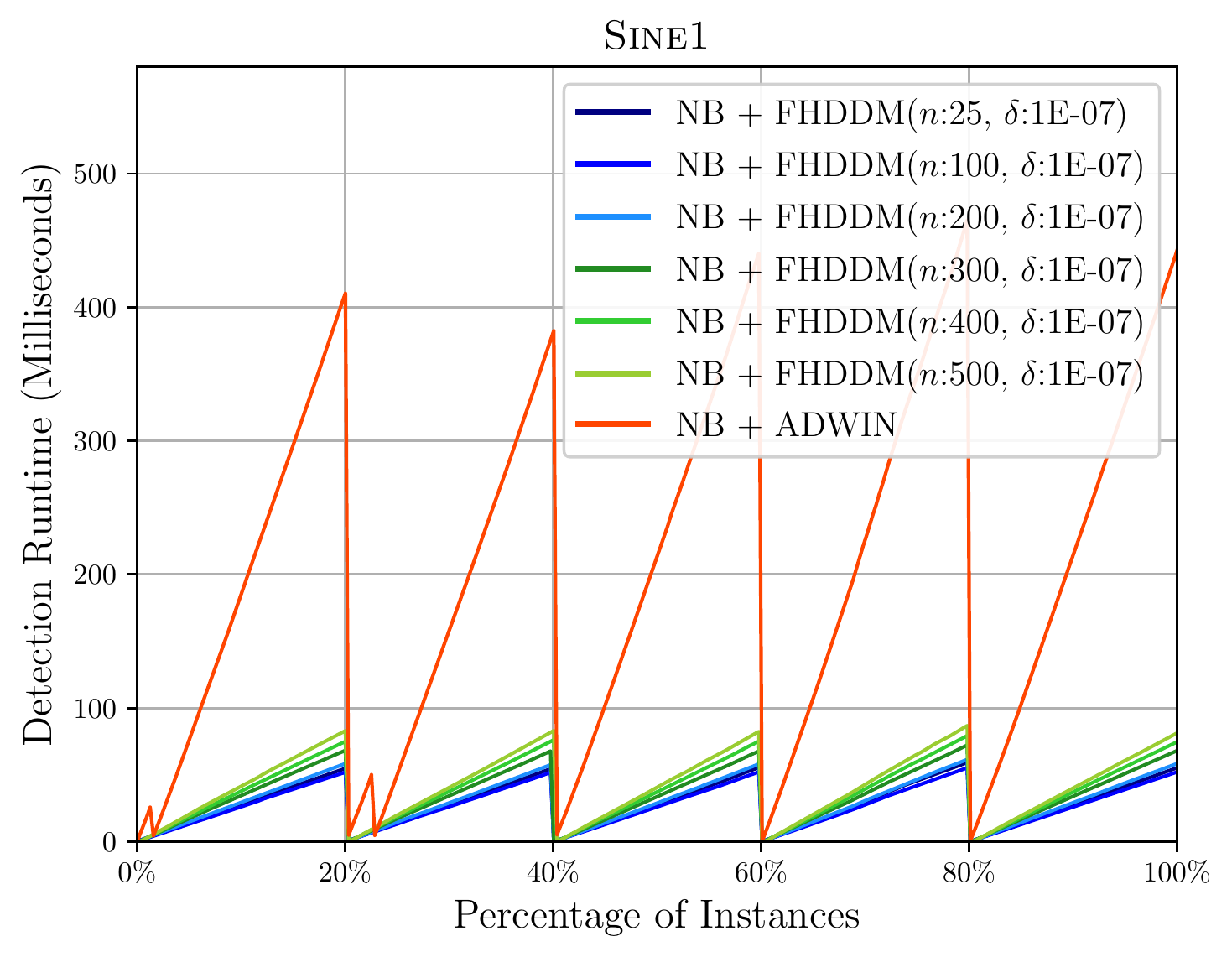}}
		\caption{FHDDM's Memory Usage and Runtime for Different Window Sizes}
		\label{fig_fhddm_param_n}
	\end{center}
\end{figure}

All the results that are shown are against the \textsc{Sine1} synthetic data stream, which is susceptible to abrupt drift. The classification is the classic $y = sin(x)$ function and the classes are reversed at drift points. (Note that more details will be provided in Section \ref{subsubsec_syn_ds}) Our explorative results are summarized in Tables \ref{table_param}, \ref{table_param_n}, and \ref{table_param_d}, as well as in Fig.\ \ref{fig_fhddm_param_n}.  Note that the averaged runtimes reported are the average of execution runtimes over five contexts, where a context is the duration between two consecutive concept drifts. 
Table \ref{table_param} shows that, as the value of $n$ increases, the value of $\varepsilon_d$ decreases. This implies that, since we have more observations, a more optimistic error bound may be used. For a constant $n$, there is an inverse relationship between $\delta$ and $\varepsilon_d$. That is, as the value of $\delta$ decreases the $\varepsilon_d$ value increases (i.e. the bound becomes more conservative).
Further, Table \ref{table_param_n} illustrates that, for a constant value of $\delta = 10^{-7}$,  the detection delay increases as we increase the value of $n$. Intuitively, memory usage and runtime also increase as the window size grows.
Table \ref{table_param_d} lists the results for a constant $n = 100$. The table shows that, as we decrease the value of $\delta$,  the detection delay increases but the false positive rate decreases. 
Finally, in Fig.\ \ref{fig_fhddm_param_n}, we contrast the memory usage of FHDDM with ADWIN, which also employs a windowing schema. The reader should notice that FHDDM constantly outperforms ADWIN (indicated in red), in terms of memory usage.

\begin{table}[h]
	\begin{center}
		\caption{Values of FHDDM's $\varepsilon_d$ for Different $n$ and $\delta$  }
		\label{table_param}
		\begin{tabu}{cc!{\vrule width 1.25pt}c|c|c|c|c|}
			\tabucline{3-7}
			 &      & \multicolumn{5}{c|}{$\delta$}                         \\ \tabucline[0.75pt]{3-7} 
			 &      & \cellcolor{whitesmoke} 0.001   & \cellcolor{whitesmoke} 0.0001  & \cellcolor{whitesmoke} 0.00001 & \cellcolor{whitesmoke} 0.000001 & \cellcolor{whitesmoke} 0.0000001 \\ \tabucline[1.25pt]{1-7} 
			\multicolumn{1}{|c!{\vrule width 0.75pt}}{\multirow{6}{*}{$n$}} & \cellcolor{whitesmoke} 25  & 0.37169 & 0.42919 & 0.47985 & 0.52565  & 0.56777   \\ \tabucline{2-7} 
			\multicolumn{1}{|c!{\vrule width 0.75pt}}{} & \cellcolor{whitesmoke} 100 & 0.18585 & 0.21460 & 0.23993 & 0.26283  & 0.28388   \\ \tabucline{2-7} 
			\multicolumn{1}{|c!{\vrule width 0.75pt}}{} & \cellcolor{whitesmoke} 200 & 0.13141 & 0.15174 & 0.16965 & 0.18585  & 0.20074   \\ \tabucline{2-7} 
			\multicolumn{1}{|c!{\vrule width 0.75pt}}{} & \cellcolor{whitesmoke} 300 & 0.10730 & 0.12390 & 0.13852 & 0.15174  & 0.16390   \\ \tabucline{2-7} 
			\multicolumn{1}{|c!{\vrule width 0.75pt}}{} & \cellcolor{whitesmoke} 400 & 0.09292 & 0.10730 & 0.11996 & 0.13141  & 0.14194   \\ \tabucline{2-7} 
			\multicolumn{1}{|c!{\vrule width 0.75pt}}{} & \cellcolor{whitesmoke} 500 & 0.08311 & 0.09597 & 0.10730 & 0.11754  & 0.12696   \\ \hline
		\end{tabu}
	\end{center}
\end{table}

\begin{table}[h]
	\setlength\tabcolsep{5pt}
	\centering
	\caption{Behavior of FHDDM for Different Values of $n$, and $\delta = 10^{-7}$}
	\label{table_param_n}
	\begin{tabular}{cc!{\vrule width 1.25pt}c|c|c|c|c|c|c|c|}
		\cline{3-10}
		 &      & Delay & TP & FP & FN & Mem. & \begin{tabular}[c]{@{}c@{}}Average\\ Runtime\end{tabular} & \begin{tabular}[c]{@{}c@{}}Total\\ Runtime\end{tabular} & Error-rate \\ \hline
		\multicolumn{1}{|c|}{\multirow{6}{*}{n}} & 25  & 38.75 & 4  & 0  & 0  & 672  & 55.67 & 277.71 & 14.31 \\ \cline{2-10} 
		\multicolumn{1}{|c|}{}                   & 100 & 49.0 & 4  & 0  & 0  & 1000 & 62.69 & 313.07 & 14.32 \\ \cline{2-10} 
		\multicolumn{1}{|c|}{}                   & 200 & 62.5  & 4  & 0  & 0  & 1376 & 73.84 & 367.23 & 14.38 \\ \cline{2-10} 
		\multicolumn{1}{|c|}{}                   & 300 & 65.75 & 4  & 0  & 0  & 1808 & 81.22 & 404.30 & 14.39 \\ \cline{2-10} 
		\multicolumn{1}{|c|}{}                   & 400 & 72.25 & 4  & 0  & 0  & 2192 & 91.82 & 458.50 & 14.40 \\ \cline{2-10} 
		\multicolumn{1}{|c|}{}                   & 500 & 80.25  & 4  & 0  & 0  & 2680 & 99.62 & 495.50 & 14.42 \\ \hline
	\end{tabular}
\end{table}


\begin{table}[h]
	\setlength\tabcolsep{5pt}
	\centering
	\caption{Behavior of FHDDM for $n=100$, and Different Values of $\delta$}
	\label{table_param_d}
	\begin{tabular}{cr!{\vrule width 1.25pt}c|c|c|c|c|c|c|c|}
		\cline{3-10}
		&            & Delay & TP & FP & FN & Mem. & \begin{tabular}[c]{@{}c@{}}Average\\ Runtime\end{tabular} & \begin{tabular}[c]{@{}c@{}}Total\\ Runtime\end{tabular} & Error-rate \\ \hline
		\multicolumn{1}{|c|}{\multirow{5}{*}{$\delta$}} & 0.001 & 36.75 & 4  & 17 & 0  & 1000 & 14.5 & 312.46 & 15.34 \\ \cline{2-10} 
		\multicolumn{1}{|c|}{} & 0.0001    & 44.75 & 4  & 5  & 0  & 1000 & 26.71 & 299.69 & 14.55 \\ \cline{2-10} 
		\multicolumn{1}{|c|}{} & 0.00001   & 42.75 & 4  & 0  & 0  & 1000 & 60.48 & 301.19 & 14.31 \\ \cline{2-10} 
		\multicolumn{1}{|c|}{} & 0.000001  & 46.75 & 4  & 0  & 0  & 1000 & 61.13 & 304.49 & 14.32 \\ \cline{2-10} 
		\multicolumn{1}{|c|}{} & 0.0000001 & 49    & 4  & 0  & 0  & 1000 & 60.23 & 300.25 & 14.32 \\ \hline
	\end{tabular}
\end{table}

Our experiments as reported in \citep{pesaranghader2016fast} indicated that {FHDDM} outperformed the state-of-the-art both in terms of adaptation and classification results. It follows that the size of the sliding window is a crucial parameter, as we illustrated above. Our experimental results further indicated that a longer window implies a longer detection delay against abrupt concepts drifts. On the other hand, a shorter window may cause higher false negative rates against gradual concept drifts. Based on this observation, we extended our approach as will be discussed in the next section.

\subsection{Stacking Fast Hoeffding Drift Detection Method (FHDDMS)}
\label{subsec_fhddms}

\begin{figure}[h]
	\centering
	\includegraphics[scale=0.425]{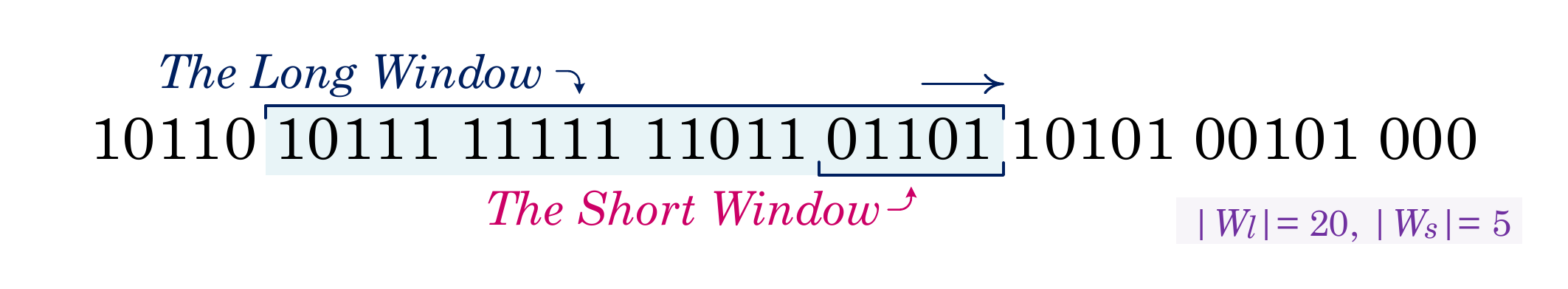}
	\caption{Stacking Fast Hoeffding Drift Detection Method (FHDDMS)}
	\label{fig_fhddms}
\end{figure}

The Stacking Hoeffding Drift Detection Method (FHDDMS)\footnote{We added the `S' at the end of FHDDMS in order to list it behind FHDDM when drift detectors are \textit{alphabetically} ordered.}, extends the FHDDM method by maintaining windows of different sizes. That is, a short and long sliding window are superimposed, as shown in Fig.\ \ref{fig_fhddms}.
The rationale behind this approach is to reduce the detection delays and false negative rates. Intuitively, a short window should detect abrupt drifts faster, while a long window should detect gradual drifts with a lower false negative rate. Following the FHDDM method, the algorithm inserts a 1 into both the short and the long windows when the prediction result is \textit{correct}, whereas a 0 is inserted otherwise. In Fig.\ \ref{fig_fhddms}, which illustrates our approach, the size of the long and the short sliding windows are set to 20 and 5, respectively.

As instances are processed, FHDDMS calculates the means of the elements inside the long and short sliding windows at time $t$, i.e. $\mu_l^t$ and $\mu_s^t$, as the stream is processed. We define $\mu_l^m$ and $\mu_s^m$ as the maximum means so far for the long and the short window, respectively:

\begin{equation}
\label{equ_41}
	\begin{gathered}
	\mu_l^m < \mu_l^t \Rightarrow \mu_l^m = \mu_l^t \\ 
	\mu_s^m < \mu_s^t \Rightarrow \mu_s^m = \mu_s^t
	\end{gathered}
\end{equation}

Recall that, as based on the probably approximately correct (PAC) learning model \citep{mitchell1997machine}, the classification accuracy increases or remains constant as the number of instances increases. Otherwise, the possibility of facing concept drifts increases \citep{gama2004learning, pesaranghader2016fast}. Thus, both the values of $\mu_l^m$ and $\mu_s^m$ should increase or remain constant as we process instances. Alternatively, the probability of a drift increases if the values of $\mu_l^m$ and $\mu_s^m$ do not change and the values of $\mu_s^t$ and $\mu_s^t$ decrease over time. As shown in Equation \eqref{equ_42}, a significant difference between the current means and their maximums indicates the occurrence of a drift in the stream. 

\begin{equation}
\label{equ_42}
	\begin{gathered}
	\Delta \mu_l =  \mu_l^m - \mu_l^t \geq \varepsilon_l \Rightarrow \psi_l = True \\
	\Delta \mu_s =  \mu_s^m - \mu_s^t \geq \varepsilon_s \Rightarrow \psi_s = True \\
	\mbox{if } (\psi_l = True) \mbox{ or } (\psi_s = True) \Rightarrow \mbox{alarm for a drift}
	\end{gathered}
\end{equation}

Here $\psi_l$ and $\psi_s$ denote the status of concept drift detection as observed by the long and short windows, respectively. 

Following \citep{pesaranghader2016fast}, we use the previously introduced Hoeffding's inequality \citep{hoeffding1963probability} and Corollary \hyperref[FHDDM_test]{I} to define the values of $\varepsilon_l$ and $\varepsilon_s$:

\begin{equation}
\label{equ_45}
	\begin{gathered}
	\varepsilon_{l} = \sqrt{\frac{1}{2 |W_l|}\ln{\frac{1}{\delta}}} \mbox{ and }
	\varepsilon_{s} = \sqrt{\frac{1}{2 |W_s|}\ln{\frac{1}{\delta}}}
	\end{gathered},
\end{equation}

where $|W_l|$ and $|W_s|$ are the sizes of the long and the short windows, respectively.

The pseudocode for the FHDDMS algorithm is presented in Algorithm \ref{algo_1}. The \textsc{Initialize} function initializes the parameters for the stacking windows. Subsequently, while data stream instances are prequentially processed, the \textsc{Detect} function analyses the prediction results in order to determine if a concept drift has occurred (lines 24-27). A drift is either detected when $( \mu_{l}^{m} - \mu_{l}^{t} ) \geq \varepsilon_l$ or when $( \mu_{s}^{m} - \mu_{s}^{t} ) \geq \varepsilon_s$.

\begin{algorithm}[h]
	\caption{Pseudocode of FHDDMS}
	\label{algo_1}
	\small
	\begin{algorithmic}[1]
		
		\Function{Initialize}{$|W_l|, |W_s|, delta$}
		\State $n_l = |W_l|$ \Comment{The size of the long window.}
		\State $n_s = |W_s|$ \Comment{The size of the short window.}
		\State $\delta = delta$
		\State $\varepsilon_l = \sqrt{\frac{1}{2n_l}\ln{\frac{1}{\delta}}}, \varepsilon_s = \sqrt{\frac{1}{2n_s}\ln{\frac{1}{\delta}}}$
		\State \Call{Reset}{}()
		\EndFunction
		\newline
		\Function{Reset}{}()
		\State $\mbox{\textit{Win}} = []$ \Comment{Creating an empty sliding window for stacking.}
		\State $\mu_{l}^{m}, \mu_{s}^{m} = 0$
		\EndFunction
		\newline
		\Function{Detect}{$p$} \Comment{\emph{p} is 1 if the correct predictions, 0 otherwise.}
			\If{\mbox{\textit{Win}} \textbf{is} full}
				\State drop an element from tail
			\EndIf

			\State \textbf{insert} $p$ into $\mbox{\textit{Win}}$
			\State \textbf{calculate} $\mu_{l}^{t}$ and $\mu_{s}^{t}$

			\If{$\mu_{l}^{m} < \mu_{l}^{t}$}
				\State $\mu_{l}^{m} = \mu_{l}^{t}$
			\EndIf
			\If{$\mu_{s}^{m} < \mu_{s}^{t}$}
				\State $\mu_{s}^{m} = \mu_{s}^{t}$
			\EndIf

			\If {$(\mu_{l}^{m} - \mu_{l}^{t}) \geq \varepsilon_l$ \textbf{or} $(\mu_{s}^{m} - \mu_{s}^{t}) \geq \varepsilon_s$}
				\State \Call{Reset}{}() \Comment{Resetting parameters.}
				\State \Return{True} \Comment{Signaling for an alarm.}
			\EndIf
			\State \Return{False}
		\EndFunction
		
	\end{algorithmic}
\end{algorithm}

\subsection{Additive FHDDMS (FHDDMS\textsubscript{add})}
\label{subsec_fhddms_add}

\begin{figure}[h]
	\begin{center}
	\includegraphics[scale=0.425]{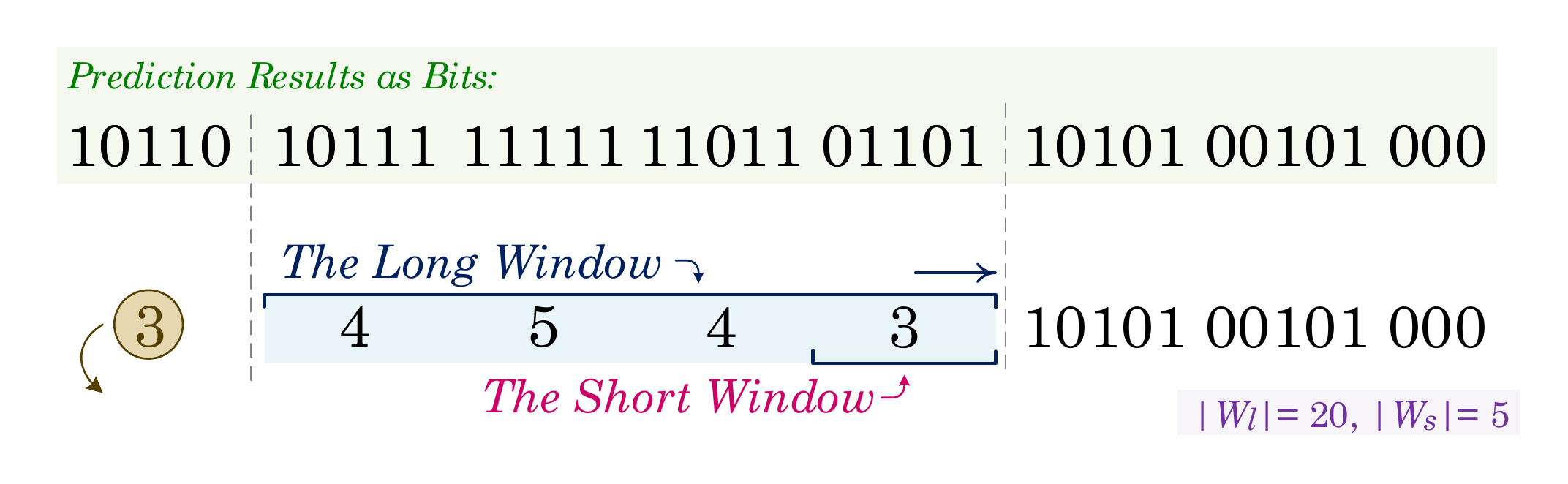}
	\caption{Additive Stacking Windows (FHDDMS\textsubscript{add}) Approach}
	\label{fig_fhddms_add}
	\end{center}
\end{figure}

We introduce another version of the FHDDMS method called the Additive FHDDMS, i.e. denoted as FHDDMS\textsubscript{add}.  In this approach, the binary indicators are substituted by summary statistics.  As shown in Fig.\ \ref{fig_fhddms_add}, the short and the long windows are characterized by the sum of their respective most recent prediction results.  In this example, the short window holds a single summation of the 5 most recent bits, while the long window holds four summations for the 20 most recent bits seen so far. For each window, the maximum values observed so far, which are $\mu_s^m$ and $\mu_l^m$, are updated as required.  In this example, the mean values $\mu_l^t$ and $\mu_s^t$, are 0.8 and 0.6, respectively.  As for the FHDDMS algorithm, a concept drift occurs if the difference in between the current values and the maximum values exceed a certain threshold, as determined by Hoeffding's inequality.

Note that, intuitively, FHDDMS\textsubscript{add} should require less memory and exhibit a faster execution time when compared to FHDDMS, since the data structure is more concise. Nevertheless, this may lead to a longer drift detection delay, since the algorithm must ensure that the short window has accumulated $|W_s|$ new predictions, after an element has been removed from the long window's tail. This is further confirmed by our experimental results in Section \ref{subsec_eval_fhddms}.

\begin{figure}[h]
	\begin{center}
		\includegraphics[scale=0.275]{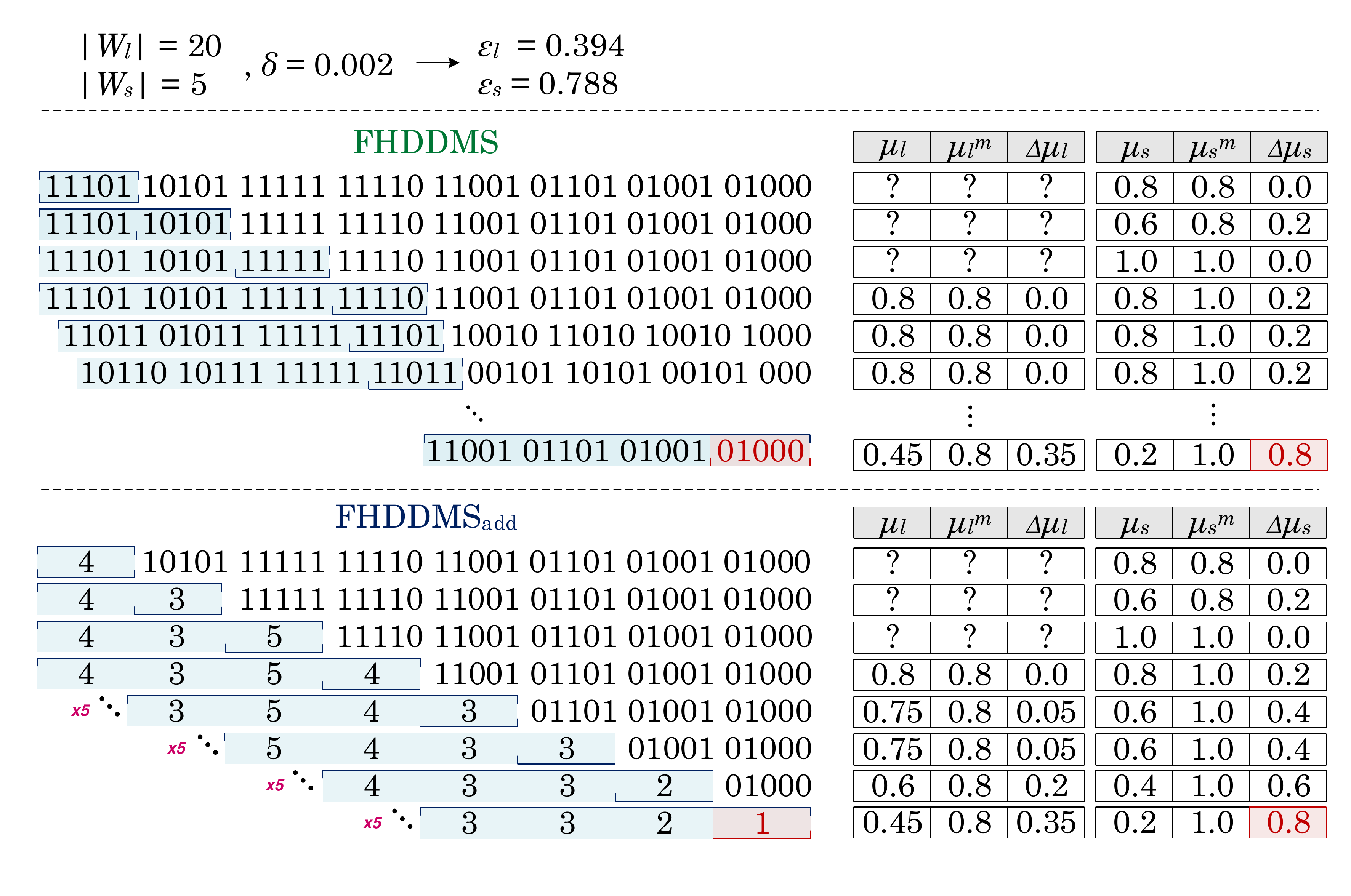}
		\caption{Example of FHDDMS and FHDDMS\textsubscript{add} algorithms}
		\label{fig_fhddms_exm}
	\end{center}
\end{figure}

Fig.\ \ref{fig_fhddms_exm} illustrates how the FHDDMS and FHDDMS\textsubscript{add} algorithms proceed. In this toy example, the sizes of the long and short windows are set to 20 and 5, respectively. We also set $\delta$ to 0.002. Using Equation \eqref{equ_45}, we have $\varepsilon_l$ equal to 0.394, and $\varepsilon_s$ to 0.788. Recall that FHDDMS considers the predictions bit-by-bit; whereas, FHDDMS\textsubscript{add} calculates the summary statistics of every 5 predictions.  Throughout this incremental learning process, the values of $\mu_l$, $\mu_l^m$, $\Delta\mu_l$ , $\mu_s$, $\mu_s^m$, $\Delta\mu_s$ are continuously updated, as indicated in the right-side of the illustration. The reader will notice that both algorithms alarm for concept drift when $\Delta\mu_s$ exceeds 0.8, i.e. has a value greater than $\varepsilon_s$.

Finally, the theoretical proofs on the bounds of false positive and false negative for the Fast Hoeffding Drift Detection Methods, including FHDDM and FHDDMS, are available in Appendix \ref{appendix_a}.


\section{Experimental Evaluation}
\label{sec_experiments}

In this section, we evaluate our new drift detection methods FHDDMS and FHDDMS\textsubscript{add}, by comparing them against the state-of-the-art. Subsequently, we perform various experiments utilizing the \textsc{Tornado} framework. We generated synthetic data streams and also considered real-world data for our experiments. We describe the synthetic and real-world data streams as well as the experimental setup in subsections \ref{subsec_data_streams} and \ref{subsec_exp_settings} respectively. We evaluate our drift detection methods and the \textsc{Tornado} framework in subsections \ref{subsec_eval_fhddms} and \ref{subsec_eval_tornado}. Our experiments are performed with a processor Intel Core i5 @ 2 $\times$ 2.30 GHz with 16GB of RAM.

\subsection{Data Streams used in Experimentation}
\label{subsec_data_streams}


\subsubsection{Synthetic Data Streams}
\label{subsubsec_syn_ds}

We have selected the previously introduced \textsc{Sine1} data stream, as well as the \textsc{Sine2}, \textsc{Mixed}, \textsc{Stagger}, \textsc{Circles} and \textsc{LED} streams, which are frequently applied in the data stream mining literature \citep{kubat1995adapting, nishida2007detecting, pesaranghader2016framework, pesaranghader2016fast, frias2015online, gama2004learning, bifet2007learning, olorunnimbe2015intelligent}, as the synthetic data streams for our experiments. Each data stream contains 100,000 instances. \textsc{Sine1}, \textsc{Sine2}, \textsc{Mixed}, \textsc{Stagger}, \textsc{Circles} have only two class labels, whereas \textsc{LED} has 10 class labels. Following the convention, we have placed drift points at every 20,000 instances in \textsc{Sine1}, \textsc{Sine2}, and \textsc{Mixed}, and at every 33,333 instances in \textsc{Stagger} with a transition length of $\zeta=50$ to simulate \textit{abrupt} concept drifts. In addition, we have put drift points at every 25,000 instances in \textsc{Circles} and \textsc{LED} data streams with a transition length of $\zeta=500$ to simulate \textit{gradual} concept drifts. 
We have added 10\% noise to each data stream, as well, to observe how robust drift detectors are against noisy data streams by asserting their ability to distinguish between concept drift and noise. Table \ref{table_1} summarizes the synthetic data streams. They may be described as follow:

\begin{itemize}[leftmargin=*]
	\item \textsc{Sine1} $\cdot$ \textit{with abrupt drift}: Recall that the stream consists of two attributes  $x$ and $y$ uniformly distributed in the interval [0, 1]. The classification function is $y = sin(x)$. Instances are classified as positive if they are under the curve; otherwise they are classified as negative. At a drift point, the class labels are reversed.
	\item \textsc{Sine2} $\cdot$ \textit{with abrupt concept drift}: It holds two attributes of $x$ and $y$ which are uniformly distributed in between 0 and 1. The classification function is $0.5 + 0.3 * sin(3 \pi x)$. Instances under the curve are classified as positive while the other instances are classified as negative. At a drift point, the classification scheme is inverted.
	\item \textsc{Mixed} $\cdot$ \textit{with abrupt drift}: The dataset has two numeric attributes $x$ and $y$ distributed in the interval [0, 1] with two boolean attributes $v$ and $w$. The instances are classified as positive if at least two of the three following conditions are satisfied: $v, w, y < 0.5 + 0.3 * sin(3\pi x)$. The classification is reversed when drift points occur.
	\item \textsc{Stagger} $\cdot$ \textit{with abrupt concept drift}: This dataset contains three nominal attributes, namely \textit{size \{small, medium, large\}}, \textit{color \{red, green\}} and \textit{shape \{circular, non-circular\}}. Before the first drift point, instances are labeled positive if $(color = red) \wedge (size = small)$. After this point and before the second drift, instances are classified positive if $(color = green) \vee (shape = circular)$, and finally after this second drift point, instances are classified positive only if $(size = medium) \vee (size = large)$.
	\item \textsc{Circles} $\cdot$ \textit{with gradual drift}: This dataset contains two attributes  $x$ and $y$ which are uniformly distributed in the interval [0, 1]. The classification function is \textless{}$(x_c,y_c), r_c$\textgreater{} is $(x - x_c)^2 + (y - y_c)^2 = r_c^2$ where $(x_c, y_c)$ is its center and $r_c$ is the radius. Instances inside the circle are classified as positive. A drift happens whenever the classification function, i.e. circle function, changes.
	\item \textsc{LED} $\cdot$ \textit{with gradual drift}: The objective of this dataset is to predict the digit on a seven-segment display, where each digit has a 10\% chance of being displayed. The dataset has 7 attributes related to the class, and 17 irrelevant ones. Concept drift is simulated by interchanging relevant attributes \citep{frias2015online}.
\end{itemize}

\begin{table}[h]
	\fontsize{8}{8}\selectfont
	\def\arraystretch{1.14}
	\setlength\tabcolsep{1.5pt}
	\begin{center}
		\caption{Summary of Synthetic Data Streams}
		\label{table_1}
		\begin{tabu}{r||c|c|c|c|c|c|c}
			\tabucline[0.75pt]{1-8}
			Data Stream & Attribute & Attr. Type & Class & Drift Points & $\zeta$ & Noise & Drift Type \\ \tabucline[0.75pt]{1-8}
			\textsc{Sine1} & 2 & Numeric & 2 & x 20,000 & 50 & 10\% & abrupt \\ \hline
			\textsc{Sine2} & 2 & Numeric & 2 &  x 20,000 & 50 & 10\% & abrupt \\ \hline
			\textsc{Mixed} & 4 & Mixed & 2 & x 20,000 & 50 & 10\% & abrupt \\ \hline
			\textsc{Stagger} & 3 & Nominal & 2 & x 33,333 & 50 & 10\% & abrupt \\ \hline
			\textsc{Circles} & 2 & Numeric & 2 & x 25,000 & 500 & 10\% & gradual \\ \hline
			\textsc{LED} & 24 & \{0, 1\} & 10 & x 25,000 & 500 & 10\% & gradual \\ \tabucline[0.75pt]{1-8}
		\end{tabu}
	\end{center}
\end{table}

\subsubsection{Real-world Data Streams}
\label{real_world_ds}

We further conducted experiments using the following real-world data streams\footnote{Available at: \url{http://moa.cms.waikato.ac.nz/datasets/2013/}}; which are frequently used in the online learning and adaptive learning literature \citep{gama2004learning,baena2006early,bifet2009new,frias2015online}. These three data streams were used in our comparative evaluation of drift detectors. 
\begin{itemize}
	\item \textsc{Electricity} contains 45,312 instances, with 8 input attributes, recorded every half an hour for two years from Australian New South Wales Electricity. The classification task is to predict a rise (\textit{Up}) or a fall (\textit{Down}) in the electricity price. The concept drift may happen because of changes in consumption habits, unexpected events, and seasonality \citep{zliobaite2013good}.
	\item \textsc{Forest CoverType} has 54 attributes with 581,012 instances describing 7 forest cover types for $30 \times 30$ meter cells obtained from US Forest Service (USFS) Region 2 Resource Information System (RIS) data, for four wilderness areas located in the Roosevelt National Forest of northern Colorado \citep{blackard1999comparative}.
	\item \textsc{Poker hand} comprises of 1,000,000 instances, where each instance is an example of a hand having five playing cards drawn from a standard deck of 52. Each card is described by two attributes (suit and rank), for ten predictive attributes. The class predicts the poker hand \citep{olorunnimbe2015intelligent}.
\end{itemize}



To the best of our knowledge, there are no real-world datasets publicly available, wherein the locations of concept drifts are clearly identified. For instance, there is consensus among researchers that the location and/or presence of concept drift in the \textsc{Electricity}, \textsc{Forest Covertype}, and \textsc{Pokerhand} data stream are unknown \citep{huang2015drift, bifet2007learning, pesaranghader2016fast, frias2015online, bifet2009new}. Therefore, in addition to the data streams mentioned above, we also used \textit{static} datasets, publicly available from the UCI machine learning repository \citep{bache2013uci}, and we simulate concept drift by switching labels at drift points. We considered the \textsc{Adult} \citep{kohavi1996scaling}, \textsc{Nursery} \citep{zupan1997machine}, and \textsc{Shuttle} \citep{catlett2002statlog} datasets for our study. We describe the original datasets as well as the preprocessing steps adopted to generate the corresponding data streams below: 

\begin{itemize}[leftmargin=*]
	\item \textsc{Adult}: The original dataset has six numeric and eight nominal attributes, two class labels, and 48,842 instances. 32,561 instances are used for building the classification models. The dataset was used to predict whether a person earns an annual income greater than \$50,000 \citep{kohavi1996scaling}. \newline
	\textit{$\triangleright$ Preprocessing}: The training dataset is imbalanced, and there are 24,720 instances for class $\leq 50K$ as oppose to 7,841 instances for class $> 50K$. We first undersampled the data, leading to 8,200 instances for class $\leq 50K$ and 7,800 instances for class $> 50K$. We subsequently increased the number of instances to 20,000 by bootstrapping. 
	\item \textsc{Nursery}: The dataset holds eight nominal attributes, five class labels, and 12,960 instances. It was designed to predict whether applications for nursery schools in Ljubljana, Slovenia should  be rejected or accepted \citep{zupan1997machine}. \newline
	\textit{$\triangleright$ Preprocessing}: The dataset consists of 5 classes labelled as  `no\_recom', `recommend', `very\_recom', `priority', and `spec\_priority'. The number of occurrences of the third and fourth classes are infrequent and we removed them from the dataset, resulting in a dataset consisting of 20,000 instances.
	\item \textsc{Shuttle}: The original dataset contains nine attributes, seven class labels, and 58,000 instances and was designed to predict suspicious states during a NASA shuttle mission \citep{catlett2002statlog}. \newline
	\textit{$\triangleright$ Preprocessing}: This dataset is also highly imbalanced. Firstly, instances from the four minority classes were filtered out and undersampling and bootstrapping were performed, in order to create a dataset of 20,000 instances.
\end{itemize}

Finally, we simulated concept drift by shifting the class labels after drift points, with a transition length of $\zeta = 50$, for the new context. Note that we use the term `context' to refer to the interval between two consecutive concept drifts. The final data streams have five contexts, each including 20,000 instances, for 100,000 instances in total\footnote{These data streams and our source code are available at \url{http://www.github.com/alipsgh}}.
 

\subsection{Experimental Setting}
\label{subsec_exp_settings}

Following \cite{bifet2009new}, we used the sigmoid function to simulate abrupt and gradual concept drifts. The function determines the probability of belonging to the new context during the transition between two contexts. The transition length $\zeta$ allows us to simulate abrupt or gradual concept drifts. It is set to 50 for abrupt concept drifts, and to 500 for gradual concept drifts in all our experiments.

\par \cite{pesaranghader2016fast} proposed an approach to evaluate drift detection methods. They introduced the \textit{acceptable delay length} notion to count true positive (TP), false positive (FP), and false negative (FN) rates. The acceptable delay length $\Delta$ is a threshold that determines how far the detected drift could be from its true location for the drift to be considered as true positive \citep{pesaranghader2016fast, krawczyk2017ensemble}. That is, we maintain three variables to count the numbers of true positives, false negatives and false positives. These variables are initially set to zero. We increment the number of true positives when the drift detector alarm is within the acceptable delay range. Otherwise, we increment the number of false negatives, since the alarm has occurred too late. In addition, the false positive value is incremented when a  false alarm occurs, outside of the acceptable delay range. Following this approach, we set $\Delta$ to 250 for the \textsc{Sine1}, \textsc{Sine2}, \textsc{Mixed}, \textsc{Stagger}, and the real world-world data streams, and to 1000 for the \textsc{Circles} and \textsc{LED} data streams. A longer $\Delta$ should be considered for data streams with gradual drifts in order to avoid a false negative increase \citep{pesaranghader2016fast}.

\par Finally, for FHDDMS\textsubscript{add} and FHDDMS, the size of the long window and the short window are set to 100 and 25, respectively. The $\delta$ is set to $10^{-7}$ in all cases. Recall that, as for the other drift detectors, the parameters were set to default values.


\subsection{Evaluation of FHDDMS and FHDDMS\textsubscript{add}}
\label{subsec_eval_fhddms}

\subsubsection{Experiments on Synthetic Data Streams}

In this section, we compare the performances of FHDDMS and FHDDMS\textsubscript{add} against DDM, EDDM, HDDMs, CUSUM, Page-Hinkley (PH), ADWIN, SeqDrift2 and FHDDM. We considered Naive Bayes (NB) and Hoeffding Tree (HT) as our incremental learners. 
We ran each classifier-detector pair 100 times. Recall that we maintained true positive, false positive, and false negative counters for each run, by considering the corresponding \textit{acceptable delay length} $\Delta$ of data stream. We captured the memory usage of drift detectors after each alarm, and then averaged them once all instances are processed. The overall detection runtime of drift detectors as well as the overall error-rates classifiers for each run were computed. Please note that the memory usage and the runtime are recorded in \textit{bytes} and \textit{milliseconds}, respectively. Further, we averaged the detection delays, true positives, false positives, false negatives, total detection runtimes, and memory usage of drift detectors as well as the error-rates of the classifiers over all iterations. \newline

\par Tables \ref{table_syn_nb} and \ref{table_syn_nb_err} summarize the experimental results for Naive Bayes with all drift detectors. As indicated in the Table \ref{table_syn_nb}, HDDM\textsubscript{W-test} and FHDDMS have the shortest detection delays, followed by FHDDM\textsubscript{n:100} and FHDDMS\textsubscript{add}. On the other hand EDDM yields the longest delay before detection, followed by PH and SeqDrift2 for \textsc{Sine1} and \textsc{Mixed} as well as DDM for \textsc{Circles}. HDDM\textsubscript{W-test} has shorter detection delays compared to FHDDMS against abrupt concept drifts, they produce similar detection delays for the data streams with gradual concept drifts. Overall, FHDDMS\textsubscript{add}, FHDDMS, FHDDM\textsubscript{n:25}, FHDDMS\textsubscript{100}, and CUSUM have the lowest false positive rates. ADWIN and SeqDrift2 have a large number of false positive numbers when they are used in conjunction with the \textsc{LED} data stream. Since EDDM does not find concept drifts within the acceptable delay lengths, it resulted in the highest false negative numbers. FHDDMS\textsubscript{add}, FHDDMS, FHDDMs, HDDMs, and CUSUM have the lowest false negative rates. FHDDMS\textsubscript{add} outperforms FHDDMS and FHDDM\textsubscript{n:100} in terms of memory consumption and detection runtime. ADWIN and SeqDrift2 require much more memory than the other approaches. Finally, as shown in Table \ref{table_syn_nb_err}, we obtained the lowest classification error-rates with FHDDMS and FHDDMs. Overall, the classification error-rates are comparable for all data streams, except for LED, where ADWIN and SeqDrift2 have much higher error-rates. \newline

\par The experimental results for Hoeffding Tree with drift detectors are shown in Tables \ref{table_syn_ht} and \ref{table_syn_ht_err}. Once more, we observe that FHDDMS, FHDDMs, and HDDM\textsubscript{W-test} have the shortest drift detection delay. HDDM\textsubscript{W-test} has the shortest detection delay when the concept drifts are abrupt; whereas, FHDDMS has the shortest detection delays when the concept drifts are gradual. FHDDMS, FHDDMs, CUSUM, and DDM caused the lowest false positives of all drift detectors. The false positive numbers of HDDMs are consistently higher than those of FHDDMS and FHDDMs. EDDM, again, resulted in the highest false negative rates. One should notice that the false positives are more common with Hoeffding Tree than with Naive Bayes. This indicates Hoeffding Tree may not represent decision boundaries adequately, which misleads drift detection methods and consequently causes more false alarms. As Table \ref{table_syn_ht_err} depicts, FHDDMS and FHDDMs led to the lowest classification error-rates. In general, the classification error-rates are similar for all data streams, except for LED where ADWIN and SeqDrift2 resulted in higher error-rates. \newline

\par In conclusion, FHDDMS had better performances compared to FHDDM\textsubscript{n:25} and FHDDM\textsubscript{n:100} against data stream containing both \textit{abrupt} and \textit{gradual} concept drifts. That is, the stacking of sliding windows assisted to detect concept drifts with shorter detection delays and fewer false negatives. Recall that this method slides a short window as well as a long window on prediction results. The short window finds abrupt drifts with shorter delays, while the long window detects gradual drifts with fewer false negatives. Finally, FHDDMS\textsubscript{add} had fewer false positives, less memory usage, and shorter runtime compared to FHDDMS. Although HDDM\textsubscript{W-test} had similar detection delays compared to FHDDMS and FHDDMs, it results in higher false positive rates. \newline

\begin{table}[p]
	\fontsize{8}{8}\selectfont
	\def\arraystretch{1.10}
	\setlength\tabcolsep{1.10pt}
	\begin{center}
		\caption{Naive Bayes and Drift Detectors against\ \textsc{Synthetic} Data Streams}
		\label{table_syn_nb}
		\begin{tabu}{r||c|c|c|c||c|c}
			\tabucline[0.75pt]{1-7}
			\multicolumn{7}{c}{\fontsize{9}{9} \selectfont \textsc{Sine1-Abrupt} ($\zeta = 50$)} \\ \hline
			Detector & Delay & TP & FP & FN & Mem. & Runtime \\ \tabucline[0.75pt]{1-7}
			\rowcolor{my_blue} FHDDMS\textsubscript{add} & 52.06$\pm$3.86 & 4.0$\pm$0.0 & \color{blue} 0.01$\pm$0.1 & 0.0$\pm$0.0 & 880.0$\pm$0.0 & 241.89$\pm$52.41\\
			\rowcolor{my_blue} FHDDMS\textsubscript{} & \textcolor{blue}{40.52$\pm$3.55} & 4.0$\pm$0.0 & 0.06$\pm$0.24 & 0.0$\pm$0.0 & 1096.0$\pm$0.0 & 1087.12$\pm$125.51 \\ \tabucline[0.5pt]{1-7}
			FHDDM\textsubscript{n:25} & \textcolor{blue}{40.87$\pm$3.62} & 4.0$\pm$0.0 & \color{blue} 0.01$\pm$0.1 & 0.0$\pm$0.0 & 672.0$\pm$0.0 & 240.33$\pm$53.85\\
			FHDDM\textsubscript{n:100}	 &  48.48$\pm$2.87 & 4.0$\pm$0.0 & 0.05$\pm$0.22 & 0.0$\pm$0.0 & 1000.0$\pm$0.0 & 280.49$\pm$60.86 \\
			\tabucline[0.5pt]{1-7}
			CUSUM & 85.14$\pm$5.44 & 4.0$\pm$0.0 & 0.03$\pm$0.17 & 0.0$\pm$0.0 & 544.0$\pm$0.0 & 287.36$\pm$2.47 \\
			PH & 234.47$\pm$11.7 & 1.74$\pm$0.99 & 2.26$\pm$0.99 & 2.26$\pm$0.99 & 592.0$\pm$0.0 & 232.42$\pm$1.97 \\
			DDM	 &  152.89$\pm$9.18 & 3.99$\pm$0.1 & \color{blue} 0.01$\pm$0.1 & 0.01$\pm$0.1 & 576.0$\pm$0.0 & 332.21$\pm$78.49 \\
			EDDM & \textcolor{red}{249.42$\pm$5.72} & \textcolor{red}{0.01$\pm$0.1} & \textcolor{red}{7.53$\pm$2.68} & \textcolor{red}{3.99$\pm$0.1} & 920.0$\pm$0.0 & \textcolor{blue}{140.79$\pm$43.16} \\
			ADWIN & 65.63$\pm$2.54 & 4.0$\pm$0.0 & \textcolor{red}{4.08$\pm$2.4} & 0.0$\pm$0.0 & \textcolor{red}{$>4310.0$} & \textcolor{red}{2191.71$\pm$141.98} \\
			SeqDrift2 & \textcolor{red}{200.83$\pm$0.89} & 4.0$\pm$0.0 & 1.74$\pm$1.51 & 0.0$\pm$0.0 & \textcolor{red}{6616.0$\pm$0.0} & 888.21$\pm$97.74 \\ 
			HDDM\textsubscript{A-test} & 68.33$\pm$16.09 & 3.99$\pm$0.1 & 0.43$\pm$0.65 & 0.01$\pm$0.1 & 656.0$\pm$0.0 & 1122.68$\pm$8.17 \\
			HDDM\textsubscript{W-test} & \textcolor{blue}{32.97$\pm$3.28} & 4.0$\pm$0.0 & 0.5$\pm$0.71 & 0.0$\pm$0.0 & 1504.0$\pm$0.0 & 1134.65$\pm$5.14 \\
			\hline
			\tabucline[0.75pt]{1-7}
			\multicolumn{7}{c}{ } \\ [-0.6em]
			\tabucline[0.75pt]{1-7}
			\multicolumn{7}{c}{\fontsize{9}{9} \selectfont \textsc{Mixed-Abrupt}  ($\zeta = 50$)} \\ \hline
			Detector & Delay & TP & FP & FN & Mem. & Runtime \\ \tabucline[0.75pt]{1-7}
			\rowcolor{my_blue} FHDDMS\textsubscript{add} & 52.19$\pm$4.09 & 4.0$\pm$0.0 & 0.0$\pm$0.0 & 0.0$\pm$0.0 & 880.0$\pm$0.0 & 249.24$\pm$68.06 \\
			\rowcolor{my_blue} FHDDMS\textsubscript{} & \textcolor{blue}{40.43$\pm$3.43} & 4.0$\pm$0.0 & 0.0$\pm$0.0 & 0.0$\pm$0.0 & 1096.0$\pm$0.0 & 1079.31$\pm$123.36 \\ \tabucline[0.5pt]{1-7}
			FHDDM\textsubscript{n:25} & \textcolor{blue}{40.8$\pm$3.45} & 4.0$\pm$0.0 & 0.0$\pm$0.0 & 0.0$\pm$0.0 & 672.0$\pm$0.0 & 240.65$\pm$56.12 \\
			FHDDM\textsubscript{n:100}	 &  48.44$\pm$3.21 & 4.0$\pm$0.0 & 0.0$\pm$0.0 & 0.0$\pm$0.0 & 1000.0$\pm$0.0 & 281.9$\pm$63.47 \\
			\tabucline[0.5pt]{1-7}
			CUSUM & 85.35$\pm$4.38 & 4.0$\pm$0.0 & 0.0$\pm$0.0 & 0.0$\pm$0.0 & 544.0$\pm$0.0 & 292.31$\pm$2.91 \\
			PH & 241.36$\pm$7.84 & 1.23$\pm$0.88 & 2.77$\pm$0.88 & 2.77$\pm$0.88 & 592.0$\pm$0.0 & 230.52$\pm$1.69 \\
			DDM & 147.36$\pm$6.7 & 3.99$\pm$0.1 & 0.05$\pm$0.22 & 0.01$\pm$0.1 & 576.0$\pm$0.0 & 356.23$\pm$73.79 \\
			EDDM & \textcolor{red}{250.0$\pm$0.0} & \textcolor{red}{0.0$\pm$0.0} & \textcolor{red}{8.2$\pm$2.71} & \textcolor{red}{4.0$\pm$0.0} & 920.0$\pm$0.0 & \textcolor{blue}{148.61$\pm$49.39} \\
			ADWIN & 68.11$\pm$8.48 & 3.97$\pm$0.17 & \textcolor{red}{9.65$\pm$4.12} & 0.03$\pm$0.17 & \textcolor{red}{$>4660.0$} & \textcolor{red}{2216.46$\pm$183.64} \\
			SeqDrift2 & \textcolor{red}{200.92$\pm$1.11} & 4.0$\pm$0.0 & 1.77$\pm$1.71 & 0.0$\pm$0.0 & \textcolor{red}{6616.0$\pm$0.0} & 925.54$\pm$198.3 \\ 
			HDDM\textsubscript{A-test} & 64.85$\pm$16.16 & 4.0$\pm$0.0 & 0.25$\pm$0.5 & 0.0$\pm$0.0 & 656.0$\pm$0.0 & 1120.06$\pm$8.12 \\
			HDDM\textsubscript{W-test} & \textcolor{blue}{33.14$\pm$3.31} & 4.0$\pm$0.0 & 0.34$\pm$0.62 & 0.0$\pm$0.0 & 1504.0$\pm$0.0 & 1140.31$\pm$10.54 \\
			\hline
			\tabucline[0.75pt]{1-7}
			\multicolumn{7}{c}{} \\ [-0.6em]
			\tabucline[0.75pt]{1-7}
			\multicolumn{7}{c}{\fontsize{9}{9} \selectfont \textsc{Circles-Gradual}  ($\zeta = 500$)} \\ \hline
			Detector & Delay & TP & FP & FN & Mem. & Runtime \\ \tabucline[0.75pt]{1-7}
			\rowcolor{my_blue} FHDDMS\textsubscript{add} & 216.08$\pm$110.18 & 2.84$\pm$0.37 & 0.17$\pm$0.4 & 0.16$\pm$0.37 & 880.0$\pm$0.0 & 250.03$\pm$59.11 \\
			\rowcolor{my_blue} FHDDMS\textsubscript{} & \textcolor{blue}{142.59$\pm$78.99} & 2.97$\pm$0.17 & \color{blue} 0.06$\pm$0.24 & \color{blue} 0.03$\pm$0.17 & 1096.0$\pm$0.0 & 1086.49$\pm$128.79 \\ \tabucline[0.5pt]{1-7}
			FHDDM\textsubscript{n:25} & 422.51$\pm$96.4 & 2.22$\pm$0.41 & 0.53$\pm$0.5 & 0.78$\pm$0.41 & 672.0$\pm$0.0 & 252.52$\pm$66.658 \\
			FHDDM\textsubscript{n:100}	 &  \textcolor{blue}{145.02$\pm$78.23} & 2.97$\pm$0.17 & \color{blue} 0.05$\pm$0.22 & \color{blue} 0.03$\pm$0.17 & 1000.0$\pm$0.0 & 281.58$\pm$65.11 \\
			\tabucline[0.5pt]{1-7}
			CUSUM & 235.06$\pm$45.68 & 2.99$\pm$0.1 & 0.19$\pm$0.39 & \color{blue} 0.01$\pm$0.1 & 544.0$\pm$0.0 & 286.83$\pm$2.3 \\
			PH & \textcolor{red}{598.51$\pm$62.38} & 2.58$\pm$0.49 & 0.42$\pm$0.49 & 0.42$\pm$0.49 & 592.0$\pm$0.0 & 232.36$\pm$1.25 \\
			DDM	 &   \textcolor{red}{514.19$\pm$63.85} & 2.68$\pm$0.47 & 0.39$\pm$0.53 & 0.32$\pm$0.47 & 576.0$\pm$0.0 & 345.81$\pm$62.88 \\
			EDDM & \textcolor{red}{960.57$\pm$83.79} & \textcolor{red}{0.35$\pm$0.55} & \textcolor{red}{8.19$\pm$3.46} & \textcolor{red}{2.65$\pm$0.55} & 920.0$\pm$0.0 & \textcolor{blue}{136.57$\pm$42.43} \\
			ADWIN & 159.77$\pm$35.94 & 3.0$\pm$0.0 & 1.47$\pm$0.82 & \color{blue} 0.0$\pm$0.0 & \textcolor{red}{$>4885.0$} & \textcolor{red}{2210.83$\pm$156.51}\\
			SeqDrift2 & 226.28$\pm$44.06 & 3.0$\pm$0.0 & 0.72$\pm$0.9 & \color{blue} 0.0$\pm$0.0 & \textcolor{red}{6616.0$\pm$0.0} & 871.95$\pm$87.94 \\
			HDDM\textsubscript{A-test} & 246.41$\pm$106.38 & 2.90$\pm$0.3 & 0.46$\pm$0.57 & 0.1$\pm$0.3 & 656.0$\pm$0.0 & 1135.89$\pm$5.92 \\
			HDDM\textsubscript{W-test} & \textcolor{blue}{141.88$\pm$96.14} & 2.93$\pm$0.26 & 0.56$\pm$0.79 & 0.07$\pm$0.26 & 1504.0$\pm$0.0 & 1142.57$\pm$18.81 \\
			\hline
			\tabucline[0.75pt]{1-7}
			\multicolumn{7}{c}{} \\ [-0.6em]
			\tabucline[0.75pt]{1-7}
			\multicolumn{7}{c}{\fontsize{9}{9} \selectfont \textsc{LED-Gradual}  ($\zeta = 500$)} \\ \hline
			Detector & Delay & TP & FP & FN & Mem. & Runtime \\ \tabucline[0.75pt]{1-7}
			\rowcolor{my_blue} FHDDMS\textsubscript{add} & 281.83$\pm$72.32 & 2.99$\pm$0.1 & 0.01$\pm$0.1 & 0.01$\pm$0.1 & 880.0$\pm$0.0 & 264.28$\pm$75.8 \\
			\rowcolor{my_blue} FHDDMS\textsubscript{} & \textcolor{blue}{250.79$\pm$55.84} & 3.0$\pm$0.0 & \color{blue} 0.0$\pm$0.0 & \color{blue} 0.0$\pm$0.0 & 1096.0$\pm$0.0 & 1247.52$\pm$154.99 \\ \tabucline[0.5pt]{1-7}
			FHDDM\textsubscript{n:25} & 423.84$\pm$134.95 & 2.8$\pm$0.53 & 0.06$\pm$0.28 & 0.2$\pm$0.53 & 672.0$\pm$0.0 & 262.21$\pm$59.06 \\
			FHDDM\textsubscript{n:100}	 &  \textcolor{blue}{250.79$\pm$55.84} & 3.0$\pm$0.0 & \color{blue} 0.0$\pm$0.0 & \color{blue} 0.0$\pm$0.0 & 1000.0$\pm$0.0 & 260.95$\pm$65.26 \\
			\tabucline[0.5pt]{1-7}
			CUSUM & 298.88$\pm$50.33 & 3.0$\pm$0.0 & \color{blue} 0.0$\pm$0.0 & \color{blue} 0.0$\pm$0.0 & 544.0$\pm$0.0 & 287.81$\pm$4.98 \\
			PH & \textcolor{red}{563.72$\pm$79.48} & 2.95$\pm$0.26 & 0.04$\pm$0.24 & 0.05$\pm$0.26 & 592.0$\pm$0.0 & 223.04$\pm$4.14 \\
			DDM & 443.14$\pm$71.08 & 3.0$\pm$0.0 & \color{blue} 0.01$\pm$0.1 & \color{blue} 0.0$\pm$0.0 & 576.0$\pm$0.0 & 374.58$\pm$89.38 \\
			EDDM & \textcolor{red}{966.75$\pm$55.69} & \textcolor{red}{0.47$\pm$0.56} & 2.82$\pm$0.97 & \textcolor{red}{2.53$\pm$0.56} & 920.0$\pm$0.0 & \textcolor{blue}{144.49$\pm$42.69} \\
			ADWIN & 554.82$\pm$208.41 & 2.46$\pm$0.67 & \textcolor{red}{347.14$\pm$9.44} & 0.54$\pm$0.67 & \textcolor{red}{$>3760.0$} & \textcolor{red}{1790.85$\pm$161.26} \\
			SeqDrift2 & 469.04$\pm$206.09 & 2.63$\pm$0.63 & \textcolor{red}{235.5$\pm$19.9} & 0.37$\pm$0.63 & \textcolor{red}{6616.0$\pm$0.0} & 864.03$\pm$121.25 \\ 
			HDDM\textsubscript{A-test} & 294.94$\pm$82.46 & 2.99$\pm$0.1 & 0.18$\pm$0.5 & 0.01$\pm$0.1 & 656.0$\pm$0.0 & 1099.1$\pm$11.59 \\
			HDDM\textsubscript{W-test} & \textcolor{blue}{257.21$\pm$87.34} & 2.95$\pm$0.26 & 0.08$\pm$0.27 & 0.05$\pm$0.26 & 1504.0$\pm$0.0 & 1098.25$\pm$11.26 \\
			\hline
			\tabucline[0.75pt]{1-7}
		\end{tabu}
	\end{center}
\end{table}

\begin{table}[p]
	\fontsize{8}{8}\selectfont
	\def\arraystretch{1.10}
	\setlength\tabcolsep{1.10pt}
	\begin{center}
		\caption{Hoeff. Tree and Drift Detectors against \textsc{Synthetic} Data Streams}
		\label{table_syn_ht}
		\begin{tabu}{r||c|c|c|c|c||c}
			\tabucline[0.75pt]{1-7}
			\multicolumn{7}{c}{\fontsize{9}{9} \selectfont \textsc{Sine1-Abrupt}  ($\zeta = 50$)} \\ \hline
			Detector & Delay & TP & FP & FN & Mem. & Runtime \\
			\tabucline[0.75pt]{1-7}
			\rowcolor{my_blue} FHDDMS\textsubscript{add} & 52.5$\pm$3.68 & 4.0$\pm$0.0 & 2.94$\pm$2.23 & 0.0$\pm$0.0 & 880.0$\pm$0.0 & 263.61$\pm$54.56 \\
			\rowcolor{my_blue} FHDDMS\textsubscript{} & \textcolor{blue}{40.82$\pm$3.39} & 4.0$\pm$0.0 & 5.92$\pm$3.38 & 0.0$\pm$0.0 & 1096.0$\pm$0.0 & 1170.1$\pm$119.46 \\ \tabucline[0.5pt]{1-7}
			FHDDM\textsubscript{n:25} & \textcolor{blue}{41.12$\pm$3.47} & 4.0$\pm$0.0 & \color{blue} 1.41$\pm$1.34 & 0.0$\pm$0.0 & 672.0$\pm$0.0 & 255.8$\pm$57.82 \\
			FHDDM\textsubscript{n:100}	 &  48.5$\pm$2.88 & 4.0$\pm$0.0 & 5.28$\pm$3.36 & 0.0$\pm$0.0 & 1000.0$\pm$0.0 & 283.31$\pm$63.54 \\
			\tabucline[0.5pt]{1-7}
			CUSUM & 87.24$\pm$4.26 & 4.0$\pm$0.0 & \color{blue} 0.11$\pm$0.37 & 0.0$\pm$0.0 & 544.0$\pm$0.0 & 296.21$\pm$2.25 \\
			PH & 249.36$\pm$1.73 & 0.15$\pm$0.36 & 3.85$\pm$0.36 & 3.85$\pm$0.36 & 592.0$\pm$0.0 & 236.16$\pm$1.06 \\
			DDM & 146.66$\pm$5.8 & 4.0$\pm$0.0 & 8.4$\pm$4.7 & 0.0$\pm$0.0 & 576.0$\pm$0.0 & 365.17$\pm$76.53 \\
			EDDM & \textcolor{red}{244.86$\pm$16.56} & \textcolor{red}{0.13$\pm$0.36} & \textcolor{red}{41.89$\pm$17.27} & \textcolor{red}{3.87$\pm$0.36} & 920.0$\pm$0.0 & \textcolor{blue}{150.63$\pm$48.14} \\
			ADWIN & 79.09$\pm$18.74 & 3.84$\pm$0.37 & \textcolor{red}{44.81$\pm$6.53} & 0.16$\pm$0.37 & \textcolor{red}{$>4640.0$} & \textcolor{red}{2072.28$\pm$153.95} \\
			SeqDrift2 & \textcolor{red}{203.71$\pm$2.91} & 4.0$\pm$0.0 & 7.7$\pm$5.02 & 0.0$\pm$0.0 & \textcolor{red}{6616.0$\pm$0.0} & 918.83$\pm$115.22 \\
			HDDM\textsubscript{A-test} & 59.56$\pm$13.98 & 3.99$\pm$0.1 & 8.13$\pm$4.1 & 0.01$\pm$0.1 & 656.0$\pm$0.0 & 1171.91$\pm$7.72 \\
			HDDM\textsubscript{W-test} & \textcolor{blue}{32.96$\pm$3.15} & 4.0$\pm$0.0 & 13.74$\pm$6.83 & 0.0$\pm$0.0 & 1504.0$\pm$0.0 & 1204.29$\pm$5.52 \\
			\hline
			\tabucline[0.75pt]{1-7}
			\multicolumn{7}{c}{ } \\ [-0.6em]

			\tabucline[0.75pt]{1-7}
			\multicolumn{7}{c}{\fontsize{9}{9} \selectfont \textsc{Mixed-Abrupt}  ($\zeta = 50$)} \\ \hline
			Detector & Delay & TP & FP & FN & Mem. & Runtime \\ \tabucline[0.75pt]{1-7}
			\rowcolor{my_blue} FHDDMS\textsubscript{add} & 54.28$\pm$6.22 & 3.99$\pm$0.1 & 5.61$\pm$2.72 & \color{blue} 0.01$\pm$0.1 & 880.0$\pm$0.0 & 293.28$\pm$57.69 \\
			\rowcolor{my_blue} FHDDMS\textsubscript{} & \textcolor{blue}{43.66$\pm$11.66} & 3.96$\pm$0.2 & 10.79$\pm$3.4 & 0.04$\pm$0.2 & 1096.0$\pm$0.0 & 1217.62$\pm$129.45 \\ \tabucline[0.5pt]{1-7}
			FHDDM\textsubscript{n:25} & \textcolor{blue}{40.77$\pm$3.63} & 4.0$\pm$0.0 & \color{blue} 1.68$\pm$1.31 & \color{blue} 0.0$\pm$0.0 & 672.0$\pm$0.0 & 268.61$\pm$62.18 \\
			FHDDM\textsubscript{n:100}	 &  \textcolor{blue}{54.84$\pm$14.46} & 3.92$\pm$0.27 & 10.38$\pm$3.54 & 0.08$\pm$0.27 & 1000.0$\pm$0.0 & 301.27$\pm$76.81 \\
			\tabucline[0.5pt]{1-7}
			CUSUM & 89.72$\pm$7.21 & 3.99$\pm$0.1 & 4.45$\pm$2.27 & 0.01$\pm$0.1 & 544.0$\pm$0.0 & 312.09$\pm$4.98 \\
			PH & 249.7$\pm$1.15 & 0.1$\pm$0.3 & 3.9$\pm$0.3 & 3.9$\pm$0.3 & 592.0$\pm$0.0 & 246.09$\pm$2.13 \\
			DDM & 134.02$\pm$16.34 & 3.99$\pm$0.1 & 6.44$\pm$3.38 & 0.01$\pm$0.1 & 576.0$\pm$0.0 & 385.81$\pm$78.87 \\
			EDDM & \textcolor{red}{242.45$\pm$17.42} & \textcolor{red}{0.22$\pm$0.44} & \textcolor{red}{12.52$\pm$4.33} & \textcolor{red}{3.78$\pm$0.44} & 920.0$\pm$0.0 & \textcolor{blue}{157.04$\pm$43.99} \\
			ADWIN & 71.66$\pm$14.16 & 3.94$\pm$0.28 & \textcolor{red}{17.68$\pm$3.49} & 0.06$\pm$0.28 & \textcolor{red}{$>4700.0$} & \textcolor{red}{2174.56$\pm$177.68} \\
			SeqDrift2 & \textcolor{red}{205.38$\pm$11.19} & 4.0$\pm$0.0 & 15.29$\pm$5.15 & \color{blue} 0.0$\pm$0.0 & \textcolor{red}{6616.0$\pm$0.0} & 952.43$\pm$141.54 \\
			HDDM\textsubscript{A-test} & 57.32$\pm$22.0 & 3.89$\pm$0.31 & 15.08$\pm$4.23 & 0.11$\pm$0.31 & 656.0$\pm$0.0 & 1181.22$\pm$9.62 \\
			HDDM\textsubscript{W-test} & \textcolor{blue}{35.52$\pm$8.84} & 3.98$\pm$0.14 & 15.39$\pm$4.21 & 0.02$\pm$0.14 & 1504.0$\pm$0.0 & 1230.75$\pm$9.78 \\
			\hline
			\tabucline[0.75pt]{1-7}
			\multicolumn{7}{c}{ } \\ [-0.6em]
			\tabucline[0.75pt]{1-7}
			\multicolumn{7}{c}{\fontsize{9}{9} \selectfont \textsc{Circles-Gradual}  ($\zeta = 500$)} \\ \hline
			Detector & Delay & TP & FP & FN & Mem. & Runtime \\ \tabucline[0.75pt]{1-7}
			\rowcolor{my_blue} FHDDMS\textsubscript{add} & 112.98$\pm$38.37 & 3.0$\pm$0.0 & \color{blue} 0.32$\pm$0.63 & 0.0$\pm$0.0 & 880.0$\pm$0.0 & 260.8$\pm$61.95 \\
			\rowcolor{my_blue} FHDDMS\textsubscript{} & \textcolor{blue}{82.87$\pm$24.51} & 3.0$\pm$0.0 & 0.84$\pm$1.04 & 0.0$\pm$0.0 & 1096.0$\pm$0.0 & 1168.7$\pm$116.36 \\ \tabucline[0.5pt]{1-7}
			FHDDM\textsubscript{n:25} & 347.64$\pm$122.06 & 2.53$\pm$0.5 & 0.66$\pm$0.74 & 0.47$\pm$0.5 & 672.0$\pm$0.0 &  250.18$\pm$51.9 \\
			FHDDM\textsubscript{n:100}	 &  \textcolor{blue}{85.62$\pm$23.49} & 3.0$\pm$0.0 & 0.74$\pm$1.05 & 0.0$\pm$0.0 & 1000.0$\pm$0.0 & 297.52$\pm$77.22 \\
			\tabucline[0.5pt]{1-7}
			CUSUM & 199.89$\pm$23.09 & 3.0$\pm$0.0 & \color{blue} 0.14$\pm$0.45 & 0.0$\pm$0.0 & 544.0$\pm$0.0 & 301.1$\pm$1.58 \\
			PH & 700.96$\pm$45.42 & 2.16$\pm$0.37 & 0.85$\pm$0.38 & 0.84$\pm$0.37 & 592.0$\pm$0.0 & 241.19$\pm$1.97 \\
			DDM & 423.98$\pm$32.66 & 3.0$\pm$0.0 & 1.1$\pm$1.24 & 0.0$\pm$0.0 & 576.0$\pm$0.0 & 353.78$\pm$78.77 \\
			EDDM & \textcolor{red}{945.83$\pm$95.16} & \textcolor{red}{0.34$\pm$0.49} & \textcolor{red}{15.77$\pm$7.24} & \textcolor{red}{2.66$\pm$0.49} & 920.0$\pm$0.0 & \textcolor{blue}{149.23$\pm$50.79} \\
			ADWIN & 187.81$\pm$119.43 & 2.78$\pm$0.44 & \textcolor{red}{6.62$\pm$2.12} & 0.22$\pm$0.44 & \textcolor{red}{$>5135.0$} & \textcolor{red}{2254.56$\pm$173.65} \\
			SeqDrift2 & 201.08$\pm$6.82 & 3.0$\pm$0.0 & 1.48$\pm$1.55 & 0.0$\pm$0.0 & \textcolor{red}{6616.0$\pm$0.0} & 932.15$\pm$108.88 \\
			HDDM\textsubscript{A-test} & 143.91$\pm$70.48 & 2.98$\pm$0.14 & 1.73$\pm$1.78 & 0.02$\pm$0.14 & 656.0$\pm$0.0 & 1183.36$\pm$8.82 \\
			HDDM\textsubscript{W-test} & \textcolor{blue}{84.17$\pm$41.6} & 3.0$\pm$0.0 & 2.24$\pm$2.08 & 0.0$\pm$0.0 & 1504.0$\pm$0.0 & 1206.03$\pm$11.4 \\
			\hline
			\tabucline[0.75pt]{1-7}
			\multicolumn{7}{c}{ } \\ [-0.6em]
			\tabucline[0.75pt]{1-7}
			\multicolumn{7}{c}{\fontsize{9}{9} \selectfont \textsc{LED-Gradual}  ($\zeta = 500$)} \\ \hline
			Detector & Delay & TP & FP & FN & Mem. & Runtime \\ \tabucline[0.75pt]{1-7}
			\rowcolor{my_blue} FHDDMS\textsubscript{add} & 270.0$\pm$80.06 & 2.96$\pm$0.24 & \color{blue} 0.04$\pm$0.24 & 0.04$\pm$0.24 & 880.0$\pm$0.0 & 293.06$\pm$74.57 \\
			\rowcolor{my_blue} FHDDMS\textsubscript{} & \textcolor{blue}{241.88$\pm$77.38} & 2.97$\pm$0.22 & \color{blue} 0.04$\pm$0.24 & \color{blue} 0.03$\pm$0.22 & 1096.0$\pm$0.0 & 1267.61$\pm$168.77 \\ \tabucline[0.5pt]{1-7}
			FHDDM\textsubscript{n:25} & 389.66$\pm$104.58 & 2.91$\pm$0.38 & 0.08$\pm$0.31 & 0.09$\pm$0.38 & 672.0$\pm$0.0 & 281.34$\pm$61.42 \\
			FHDDM\textsubscript{n:100}	 &  \textcolor{blue}{241.96$\pm$77.4} & 2.97$\pm$0.22 & \color{blue} 0.03$\pm$0.22 & \color{blue} 0.03$\pm$0.22 & 1000.0$\pm$0.0 & 323.75$\pm$77.67 \\
			\tabucline[0.5pt]{1-7}
			CUSUM & 311.29$\pm$57.52 & 2.99$\pm$0.1 & 0.01$\pm$0.1 & \color{blue} 0.01$\pm$0.1 & 544.0$\pm$0.0 & 307.48$\pm$3.79 \\
			PH & 744.37$\pm$93.2 & 2.68$\pm$0.63 & 0.29$\pm$0.57 & 0.32$\pm$0.63 & 592.0$\pm$0.0 & 239.49$\pm$2.56 \\
			DDM & 444.75$\pm$68.51 & 2.96$\pm$0.24 & 0.05$\pm$0.26 & 0.04$\pm$0.24 & 576.0$\pm$0.0 & 380.81$\pm$79.93 \\
			EDDM & \textcolor{red}{998.84$\pm$5.38} & \textcolor{red}{0.07$\pm$0.26} & 3.81$\pm$1.39 & \textcolor{red}{2.93$\pm$0.26} & 920.0$\pm$0.0 & \textcolor{blue}{170.45$\pm$53.28} \\
			ADWIN & 314.39$\pm$156.75 & 2.99$\pm$0.1 & \textcolor{red}{301.49$\pm$3.87} & \color{blue} 0.01$\pm$0.1 & \textcolor{red}{$>3855.0$} & \textcolor{red}{1931.31$\pm$186.19} \\
			SeqDrift2 & 268.86$\pm$81.99 & 3.0$\pm$0.0 & \textcolor{red}{244.87$\pm$0.72} & \color{blue} 0.0$\pm$0.0 & \textcolor{red}{6616.0$\pm$0.0} & 888.76$\pm$128.74 \\
			HDDM\textsubscript{A-test} & 294.41$\pm$76.74 & 2.98$\pm$0.2 & 0.22$\pm$0.48 & \color{blue} 0.02$\pm$0.2 & 656.0$\pm$0.0 & 1141.51$\pm$9.02 \\
			HDDM\textsubscript{W-test} & \textcolor{blue}{257.21$\pm$87.44} & 2.95$\pm$0.26 & 0.1$\pm$0.33 & 0.05$\pm$0.26 & 1504.0$\pm$0.0 & 1165.5$\pm$10.48 \\
			\hline
			\tabucline[0.75pt]{1-7}
			\multicolumn{7}{c}{ } \\ [-0.6em]
		\end{tabu}
	\end{center}
\end{table}

\clearpage

\begin{table}
	\fontsize{8}{8}\selectfont
	\def\arraystretch{1.10}
	\setlength\tabcolsep{1.10pt}
	\begin{center}
		\caption{Naive Bayes Error-rates against\ \textsc{Synthetic} Data Streams}
		\label{table_syn_nb_err}
		\begin{tabu}{r||C{1.75cm}|C{1.75cm}|C{1.75cm}|C{1.75cm}}
			\tabucline[0.75pt]{1-5} Detector & \textsc{Sine1} & \textsc{Mixed} & \textsc{Circles} & \textsc{LED\textsubscript{0.3.1.3}} \\ \tabucline[0.75pt]{1-5} 
			\rowcolor{my_blue} FHDDMS\textsubscript{add} & 14.39$\pm$0.17 & 13.51$\pm$0.11 & 13.88$\pm$0.12 & 10.45$\pm$0.03 \\	
			\rowcolor{my_blue} FHDDMS\textsubscript{} & \color{blue} 14.37$\pm$0.17 & \color{blue} 13.49$\pm$0.11 & \color{blue} 13.83$\pm$0.08 & \color{blue} 10.44$\pm$0.04 \\ \tabucline[0.5pt]{1-5}
			FHDDM\textsubscript{n:25} & \color{blue} 14.37$\pm$0.17 & \color{blue} 13.49$\pm$0.11 & 14.58$\pm$0.73 & 10.52$\pm$0.23 \\
			FHDDM\textsubscript{n:100}	 & 14.38$\pm$0.17 & 13.51$\pm$0.11 & \color{blue} 13.83$\pm$0.08 & \color{blue} 10.44$\pm$0.04 \\ \tabucline[0.5pt]{1-5}
			CUSUM & 14.48$\pm$0.17 & 13.61$\pm$0.11 & 13.88$\pm$0.07 & \color{blue} 10.44$\pm$0.03 \\
			PH & 14.98$\pm$0.18 & 14.15$\pm$0.13 & 14.08$\pm$0.09 & 10.67$\pm$0.04 \\
			DDM & 14.68$\pm$0.17 & 13.80$\pm$0.11 & 14.04$\pm$0.10 & 10.52$\pm$0.02 \\
			EDDM & \color{red} 16.97$\pm$0.26 & \color{red} 16.08$\pm$0.19 & \color{red} 15.18$\pm$0.33 & 11.67$\pm$0.20 \\
			ADWIN & 14.74$\pm$0.23 & 14.35$\pm$0.34 & 13.85$\pm$0.07 & \color{red} 27.79$\pm$0.56 \\
			SeqDrift2 & 14.88$\pm$0.19 & 14.04$\pm$0.14 & 13.88$\pm$0.07 & \color{red} 22.58$\pm$1.13 \\
			HDDM\textsubscript{A-test} & 14.47$\pm$0.18 & 13.58$\pm$0.12 & 13.88$\pm$0.09 & 10.47$\pm$0.05 \\
			HDDM\textsubscript{W-test} & \color{blue} 14.38$\pm$0.18 & 13.51$\pm$0.12 & \color{blue} 13.83$\pm$0.09 & 10.45$\pm$0.04 \\
			\cdashline{1-5} 
			\cdashline{1-5}
			\textsc{No Detection} & 43.01$\pm$0.17 & 43.24$\pm$0.14 & 24.58$\pm$0.14 & 27.4$\pm$ 4.41 \\
			\tabucline[0.75pt]{1-5}
		\end{tabu}
	\end{center}
\end{table}

\begin{table}
	\fontsize{8}{8}\selectfont
	\def\arraystretch{1.10}
	\setlength\tabcolsep{1.10pt}
	\begin{center}
		\caption{Hoeffding Tree Error-rates against\ \textsc{Synthetic} Data Streams}
		\label{table_syn_ht_err}
		\begin{tabu}{r||C{1.75cm}|C{1.75cm}|C{1.75cm}|C{1.75cm}}
			\tabucline[0.75pt]{1-5} Detector & \textsc{Sine1} & \textsc{Mixed} & \textsc{Circles} & \textsc{LED\textsubscript{0.3.1.3}} \\ \tabucline[0.75pt]{1-5} 
			\rowcolor{my_blue} FHDDMS\textsubscript{add} & \color{blue} 14.16$\pm$0.20 & \color{blue} 15.18$\pm$0.26 & 13.08$\pm$0.11 & 10.92$\pm$0.06 \\	
			\rowcolor{my_blue} FHDDMS\textsubscript{} & 14.33$\pm$0.29 & 15.51$\pm$0.27 & \color{blue} 13.10$\pm$0.12 & \color{blue} 10.91$\pm$0.07 \\ 
			\tabucline[0.5pt]{1-5}
			FHDDM\textsubscript{n:25} & \color{blue} 14.01$\pm$0.12 & \color{blue} 14.82$\pm$0.16 & 13.40$\pm$0.42 & 10.97$\pm$0.12 \\
			FHDDM\textsubscript{n:100} & 14.31$\pm$0.29 & 15.54$\pm$0.31 & \color{blue} 13.09$\pm$0.12 & \color{blue} 10.91$\pm$0.06 \\ \tabucline[0.5pt]{1-5} 
			CUSUM & \color{blue} 14.07$\pm$0.10 & 15.33$\pm$0.33 & 13.14$\pm$0.11 & \color{blue} 10.91$\pm$0.07 \\
			PH & 14.80$\pm$0.13 & 15.54$\pm$0.11 & 13.49$\pm$0.12 & 11.37$\pm$0.11 \\
			DDM & 14.88$\pm$0.37 & 15.64$\pm$0.32 & 13.32$\pm$0.13 & 10.98$\pm$0.06 \\
			EDDM & \color{red} 18.00$\pm$0.63 & \color{red} 17.46$\pm$0.20 & \color{red} 14.99$\pm$0.24 & 12.38$\pm$0.21 \\	
			ADWIN & \color{red} 18.24$\pm$0.52 & 16.13$\pm$0.24 & 13.75$\pm$0.24 & \color{red} 50.80$\pm$0.31 \\	
			SeqDrift2 & 15.00$\pm$0.43 & 16.16$\pm$0.36 & 13.19$\pm$0.12 & \color{red} 45.82$\pm$0.25 \\
			HDDM\textsubscript{A-test} & 14.53$\pm$0.33 & 15.73$\pm$0.28 & 13.20$\pm$0.23 & 10.95$\pm$0.1 \\
			HDDM\textsubscript{W-test} & 14.80$\pm$0.47 & 15.71$\pm$0.28 & 13.18$\pm$0.18 & 10.92$\pm$0.07 \\
			\cdashline{1-5} 
			\textsc{No Detection} & 44.29$\pm$0.19 & 44.33$\pm$0.15 & 22.83$\pm$0.71 & 17.03$\pm$2.46 \\
			\tabucline[0.75pt]{1-5}
		\end{tabu}
	\end{center}
\end{table}

\subsubsection{Experiments on Real-world Data Streams}

In this section, we present the results of our experiments on the \textsc{Electricity}, \textsc{Forest CoverType}, and \textsc{Poker hand} data streams, as introduced in Section \ref{real_world_ds}, using Naive Bayes (NB) and Hoeffding Tree (HT) as the incremental learners. 
Again we stress that, as pointed out by \cite{huang2015drift} and \cite{bifet2009new}, the locations of concept drifts are not known in these data streams. We therefore follow the work of \cite{huang2015drift} and establish our evaluations based on the number of alarms for concept drifts and the classification error-rates.
Our experimental results are summarized in Tables \ref{table_ele} to \ref{table_poker}.  

Since we are not aware of the exact drift locations, we do not make strong conclusions about the performance, in terms of drift detection, of the algorithms. However, when considering the Electricity data stream, the reader will notice that our FHDDMS algorithms and the HDDM algorithms obtained the lowest error-rates, when combined with both classifiers. The HDDM and EDDM algorithms alarmed most often for concept drift, while the FHDDMS algorithms signal for drifts more often than the other remaining techniques. Overall, the memory usages and runtimes of our methods are comparable to the state-of-the art. Similar observations hold for the Forest Covertype and the Poker Hand datasets. Overall, this result indicates that there is no single drift detector that outperforms in all settings. The reader should recall that, based on this observation, we introduced the \textsc{Tornado} framework which is used to constantly recommends the currently best performing (classifier, detector) pair to the user. Next, we discuss our experimental evaluation of the \textsc{Tornado} framework.

\begin{table}[h]
	\fontsize{8}{8}\selectfont
	\def\arraystretch{1.10}
	\setlength\tabcolsep{1.10pt}
	\begin{center}
		\caption{Naive Bayes and Hoeffding Tree against\ \textsc{Electricity} Data Stream}
		\label{table_ele}
		\begin{tabu}{rC{1cm}|C{1cm}|C{1.25cm}|C{1.25cm}|C{1cm}|C{1cm}|C{1cm}|C{1cm}}
			\tabucline[0.75pt]{2-9} 
			& \multicolumn{8}{c}{\textsc{Electricity}} \\ \tabucline[0.75pt]{2-9}
			& \multicolumn{2}{c|}{Memory} & \multicolumn{2}{c|}{Runtime} & \multicolumn{2}{c|}{Num. Drifts} & \multicolumn{2}{c}{Error-rate} \\
			\tabucline[0.75pt]{2-9} 
			& \cellcolor{whitesmoke} NB & \cellcolor{whitesmoke} HT & \cellcolor{whitesmoke} NB & \cellcolor{whitesmoke} HT & \cellcolor{whitesmoke} NB & \cellcolor{whitesmoke} HT & \cellcolor{whitesmoke} NB & \cellcolor{whitesmoke} HT \\
			\tabucline[0.75pt]{1-9} 
			\multicolumn{1}{r||}{FHDDMS\textsubscript{add}} & 880.0 & 880.0 & 140.41 & 174.98 & 68 & 64 & 26.86 & 26.80 \\
			\multicolumn{1}{r||}{FHDDMS} & 1096.0 & 1096.0 & 450.14 & 582.13 & 96 & 102 & \textbf{26.05} & \textbf{26.64} \\
			\tabucline[0.5pt]{1-9} 
			\multicolumn{1}{r||}{FHDDM\textsubscript{n:25}} & 672.0 & 672.0 & 125.56 & 168.66 & 102 & 102 & \textbf{26.23} & \textbf{26.70} \\
			\multicolumn{1}{r||}{FHDDM\textsubscript{n:100}}  & 1000.0 & 1000.0 &  141.84 & 195.5 & 57 & 56 & \textbf{26.54} & \textbf{26.38} \\
			\tabucline[0.75pt]{1-9}
			\multicolumn{1}{r||}{CUSUM} & 544.0 & 544.0 & 137.98 & 180.07 & 21 & 19 & 28.35 & 27.95 \\
			\multicolumn{1}{r||}{PH} & 592.0 & 592.0 & 112.37 & 142.8 & 9 & 7 & 29.91 & 28.64 \\
			\multicolumn{1}{r||}{DDM} & 576.0 & 576.0 & 178.87 & 231.41 & 28 & 9 & 30.82 & 29.98 \\
			\multicolumn{1}{r||}{EDDM} & 920.0 & 920.0 & 88.43 & 118.61 & 195 & 168 & 27.42 & 27.37 \\
			\multicolumn{1}{r||}{ADWIN} & 4098.76 & 4066.15 & 930.33 & 1094.65 & 29 & 26 & 28.08 & 27.67 \\
			\multicolumn{1}{r||}{SeqDrift2} & 6616.0 & 6616.0 & 421.54 & 511.13 & 21 & 23 & 29.33 & 28.04 \\
			\multicolumn{1}{r||}{HDDM\textsubscript{A-test}} & 656.0 & 656.0 & 488.85 &  590.4 & 166 & 160 & \textbf{26.37} & \textbf{26.71} \\ 
			\multicolumn{1}{r||}{HDDM\textsubscript{W-test}} & 1504.0 & 1504.0 & 515.76 &  629.05 & 159 & 156 & \textbf{26.09} & \textbf{26.45} \\
			\cdashline{1-9} 
			\multicolumn{1}{r||}{\textsc{No Detection}}			 & --- & --- & --- & --- & --- & --- & 33.49 & 29.46 \\
			\tabucline[0.75pt]{1-9} 
		\end{tabu}
	\end{center}
\end{table}

\begin{table}[h]
	\fontsize{8}{8}\selectfont
	\def\arraystretch{1.10}
	\setlength\tabcolsep{1.10pt}
	\begin{center}
		\caption{NB and Hoeff. Tree against \textsc{Forest CoverType} Data Stream}
		\label{table_forest}
		\begin{tabu}{rC{1cm}|C{1cm}|C{1.25cm}|C{1.25cm}|C{1cm}|C{1cm}|C{1cm}|C{1cm}}
			\tabucline[0.75pt]{2-9} 
			& \multicolumn{8}{c}{\textsc{Forest CoverType}} \\ \tabucline[0.75pt]{2-9}
			& \multicolumn{2}{c|}{Memory} & \multicolumn{2}{c|}{Runtime} & \multicolumn{2}{c|}{Num. Drifts} & \multicolumn{2}{c}{Error-rate} \\
			\tabucline[0.75pt]{2-9} 
			& \cellcolor{whitesmoke} NB & \cellcolor{whitesmoke} HT & \cellcolor{whitesmoke} NB & \cellcolor{whitesmoke} HT & \cellcolor{whitesmoke} NB & \cellcolor{whitesmoke} HT & \cellcolor{whitesmoke} NB & \cellcolor{whitesmoke} HT \\
			\tabucline[0.75pt]{1-9} 
			\multicolumn{1}{r||}{FHDDMS\textsubscript{add}} &  880.0 & 880.0 & 1847.37 & 2666.31 & 903 & 2236 & 18.44 & 33.13 \\
			\multicolumn{1}{r||}{FHDDMS} & 1096.0 & 1096.0 & 5904.78 & 6293.78 & 1248 & 2771 & \textbf{19.02} & 32.28 \\
			\tabucline[0.5pt]{1-9} 
			\multicolumn{1}{r||}{FHDDM\textsubscript{n:25}} & 672.0 & 672.0 & 1762.32 & 2007.05 & 1315 & 3810 & 19.13 & 31.96 \\
			\multicolumn{1}{r||}{FHDDM\textsubscript{n:100}} & 1000.0 & 1000.0 & 1779.57 & 1906.87 & 751 & 1734 & \textbf{17.94} & 32.24 \\
			\tabucline[0.75pt]{1-9}
			\multicolumn{1}{r||}{CUSUM} & 544.0 & 544.0 & 1814.89 & 2072.87 & 127 & 204 & \textbf{16.92} & \textbf{21.46} \\
			\multicolumn{1}{r||}{PH} & 592.0 & 592.0 & 1485.13 & 1596.87 & 54 & 19 & \textbf{18.41} & \textbf{19.87} \\
			\multicolumn{1}{r||}{DDM} & 576.0 & 576.0 & 2321.39 & 2556.4 & 1033 & 3228 & \textbf{17.79} & \textbf{27.52} \\
			\multicolumn{1}{r||}{EDDM} & 920.0 & 920.0 & 1014.88 & 1229.12 & 1761 & 4677 & 20.51 & 30.92 \\
			\multicolumn{1}{r||}{ADWIN} & 3919.16 & 3411.24 & 11383.54 & 10479.77 & 771 & 2066 & 18.45 & 33.94 \\
			\multicolumn{1}{r||}{SeqDrift2} & 6616.0 & 6616.0 & 5275.13 & 5234.44 & 476 & 898 & \textbf{17.96} & \textbf{30.82} \\
			\multicolumn{1}{r||}{HDDM\textsubscript{A-test}} & 656.0 & 656.0 & 5792.24 & 6041.68 & 2993 & 5290 & 21.57 & \textbf{30.84} \\
			\multicolumn{1}{r||}{HDDM\textsubscript{W-test}} & 1504.0 & 1504.0 & 6475.53 & 7010.18 & 1952 & 3694 & 20.23 & 31.67 \\
			\cdashline{1-9} 
			\multicolumn{1}{r||}{\textsc{No Detection}}			 & --- & --- & --- & --- & --- & --- & 37.12 & 23.6 \\
			\tabucline[0.75pt]{1-9} 
		\end{tabu}
	\end{center}
\end{table}

\begin{table}[h]
	\fontsize{8}{8}\selectfont
	\def\arraystretch{1.10}
	\setlength\tabcolsep{1.10pt}
	\begin{center}
		\caption{Naive Bayes and Hoeff. Tree against\ \textsc{Poker hand} Data Stream}
		\label{table_poker}
		\begin{tabu}{rC{1cm}|C{1cm}|C{1.25cm}|C{1.25cm}|C{1cm}|C{1cm}|C{1cm}|C{1cm}}
			\tabucline[0.75pt]{2-9}
			& \multicolumn{8}{c}{\textsc{Poker hand}} \\ \tabucline[0.75pt]{2-9} 
			& \multicolumn{2}{c|}{Memory} & \multicolumn{2}{c|}{Runtime} & \multicolumn{2}{c|}{Num. Drifts} & \multicolumn{2}{c}{Error-rate} \\
			\tabucline[0.75pt]{2-9} 
			& \cellcolor{whitesmoke} NB & \cellcolor{whitesmoke} HT & \cellcolor{whitesmoke} NB & \cellcolor{whitesmoke} HT & \cellcolor{whitesmoke} NB & \cellcolor{whitesmoke} HT & \cellcolor{whitesmoke} NB & \cellcolor{whitesmoke} HT \\
			\tabucline[0.75pt]{1-9} 
			\multicolumn{1}{r||}{FHDDMS\textsubscript{add}} &  880.0 &  880.0 & 2771.05 & 2453.21 & 1156 & 1215 & 25.14 & 25.65 \\
			\multicolumn{1}{r||}{FHDDMS} & 1096.0 & 1096.0 & 8953.95 & 8445.41 & 1614 & 1682 & \textbf{24.74} & \textbf{25.32} \\
			\tabucline[0.5pt]{1-9} 
			\multicolumn{1}{r||}{FHDDM\textsubscript{n:25}} & 672.0 & 672.0 & 2624.02 & 2389.83 & 1561 & 1646 & \textbf{24.78} & \textbf{25.30} \\
			\multicolumn{1}{r||}{FHDDM\textsubscript{n:100}} & 1000.0 & 1000.0 & 2835.59 & 2601.9 & 1216 & 1255 & 25.11 & 25.66 \\
			\tabucline[0.75pt]{1-9}
			\multicolumn{1}{r||}{CUSUM} & 544.0 & 544.0 & 2798.63 & 2791.56 & 488 & 400 & 26.37 & 25.68 \\
			\multicolumn{1}{r||}{PH} & 592.0 & 592.0 & 2217.57 & 2418.79 & 238 & 126 & 29.16 & 25.73 \\
			\multicolumn{1}{r||}{DDM} & 576.0 & 576.0 & 3565.46 & 3264.96 & 1425 & 1261 & 25.36 & \textbf{24.41} \\
			\multicolumn{1}{r||}{EDDM} & 920.0 & 920.0 & 1629.01 & 1510.74 & 4392 & 4415 & \textbf{24.69} & 25.45 \\
			\multicolumn{1}{r||}{ADWIN} & 4108.1 & 4122.6 & 16579.36 & 15818.67 & 834 & 824 & 25.85 & 26.18 \\
			\multicolumn{1}{r||}{SeqDrift2} & 6616.0 & 6616.0 & 7594.97 & 7225.12 & 946 & 870 & 26.76 & 26.85 \\ 
			\multicolumn{1}{r||}{HDDM\textsubscript{A-test}} & 656.0 & 656.0 & 9345.98 & 9607.8 & 2493 & 2944 & \textbf{24.72} & \textbf{25.36} \\
			\multicolumn{1}{r||}{HDDM\textsubscript{W-test}} & 1504.0 & 1504.0 & 10006.12 & 11045.93 & 2550 & 2619 & \textbf{24.44} & \textbf{25.14} \\
			\cdashline{1-9} 
			\multicolumn{1}{r||}{\textsc{No Detection}}			 & --- & --- & --- & --- & --- & --- & 39.88 & 18.93 \\
			\tabucline[0.75pt]{1-9} 
		\end{tabu}
	\end{center}
\end{table}


\subsection{Experimental Evaluation of \textsc{Tornado} framework}
\label{subsec_eval_tornado}

This section presents the experimental results for the \textsc{Tornado} framework for synthetic and real-world data streams. Five learning algorithms were evaluated, namely incremental Naive Bayes (NB), Decision Stump (DS), Hoeffding Tree (HT), Perceptron (PR), and 5 Nearest Neighbors (5-NN) learners. In addition, the previously introduced drift detection methods, in Section \ref{subsec_ddms}, were paired with the classifiers. As a result, we have a total of 60 pairs of (classifier, detector). The scores associated with the (classifier, detector) pairs are measured utilizing the CAR measure, Equation \eqref{equ_27}, as instances are prequentially processed over time. The experimental results are discussed in the following subsections.


\subsubsection{Synthetic Data Streams}

\begin{figure}[h]
	\centering
	\subfloat[Score\label{fig_staggera}]{\adjincludegraphics[scale=0.41,trim={0 0 0 0cm},clip]{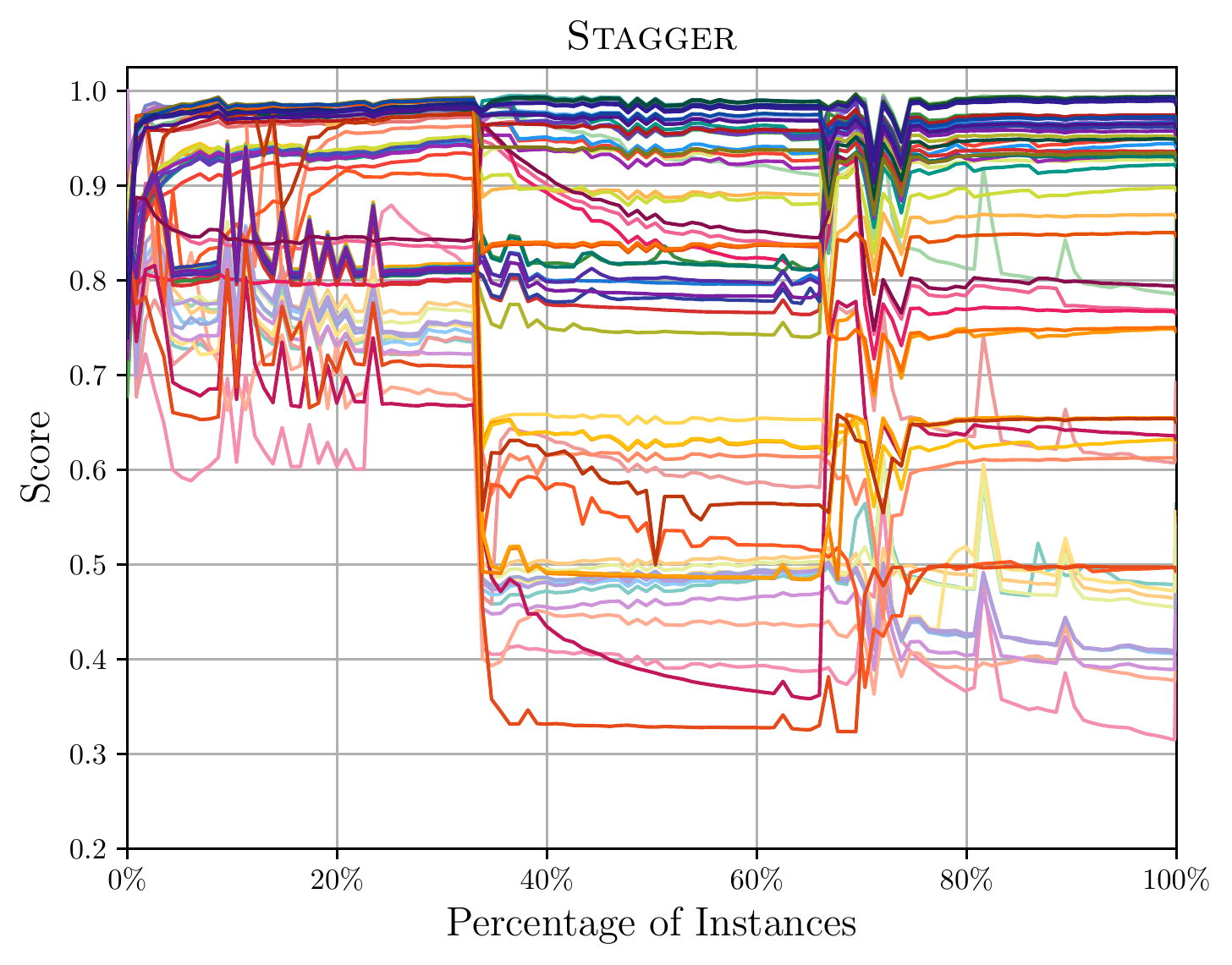}}
	\subfloat[Error-rate\label{fig_staggerb}]{\adjincludegraphics[scale=0.41,trim={0 0 0 0cm},clip]{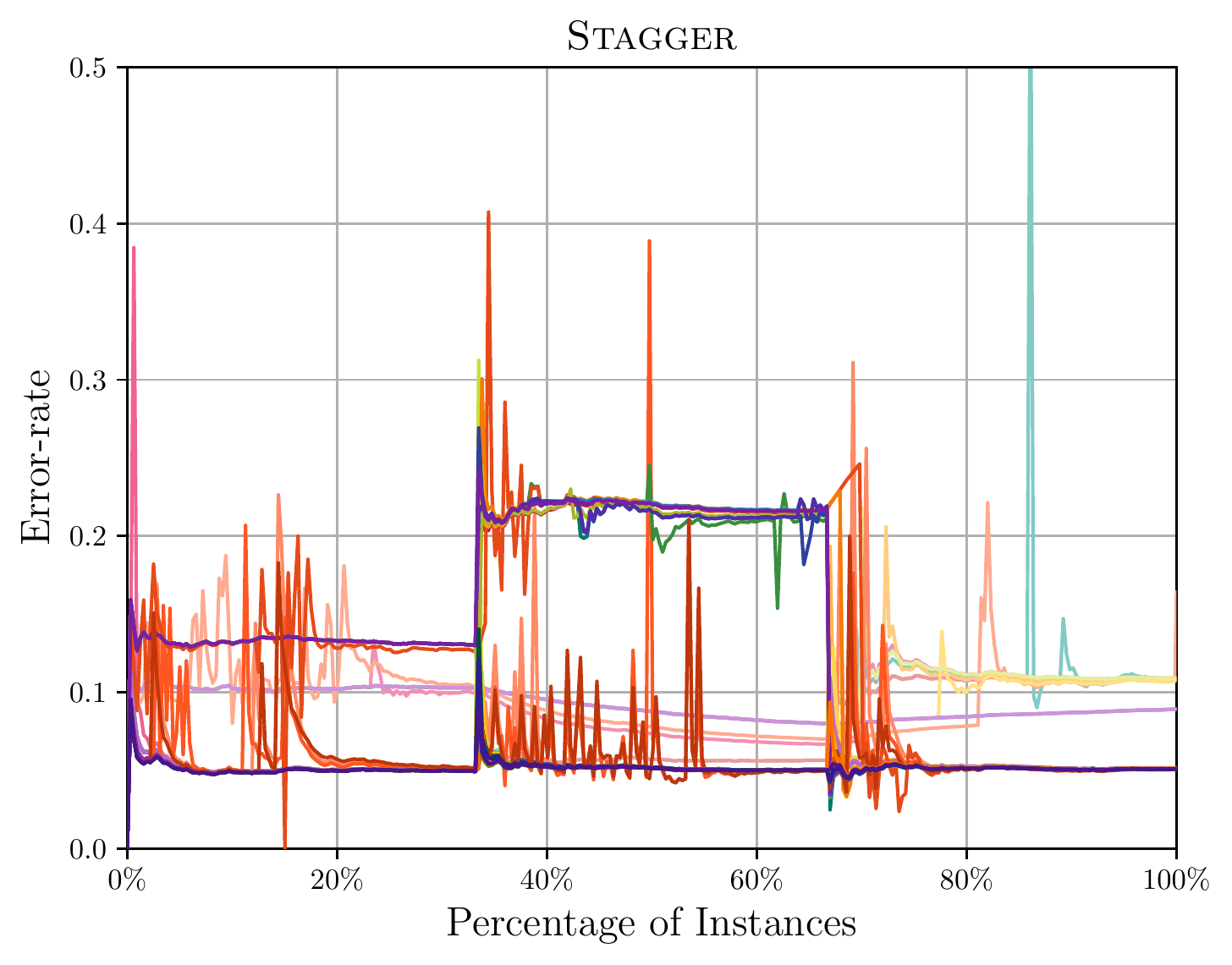}}
	\\
	\subfloat[Memory Usage\label{fig_staggerc}]{\adjincludegraphics[scale=0.405,trim={0 0 0 0cm},clip]{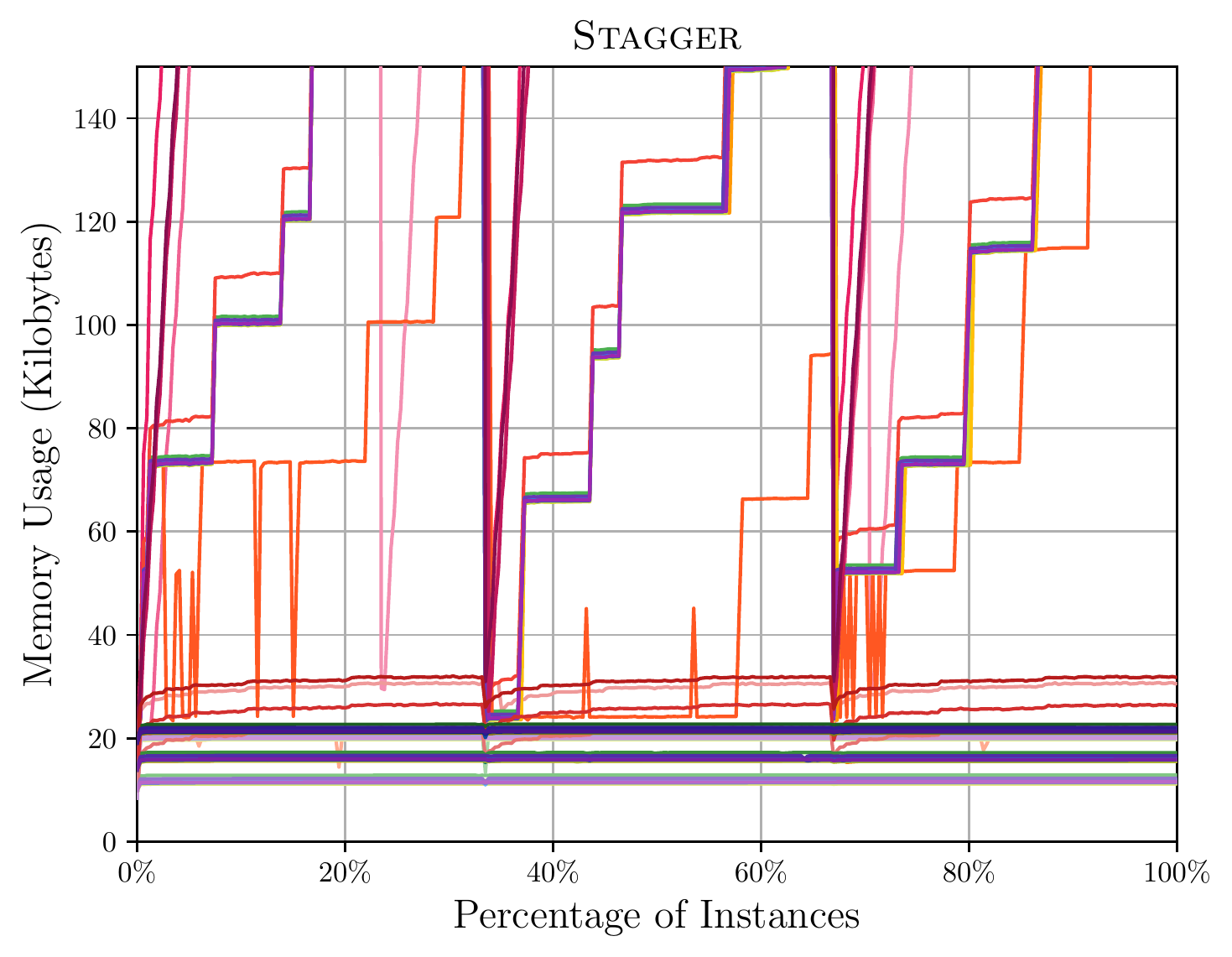}}
	\subfloat[Runtime\label{fig_staggerd}]{\adjincludegraphics[scale=0.405,trim={0 0 0 0cm},clip]{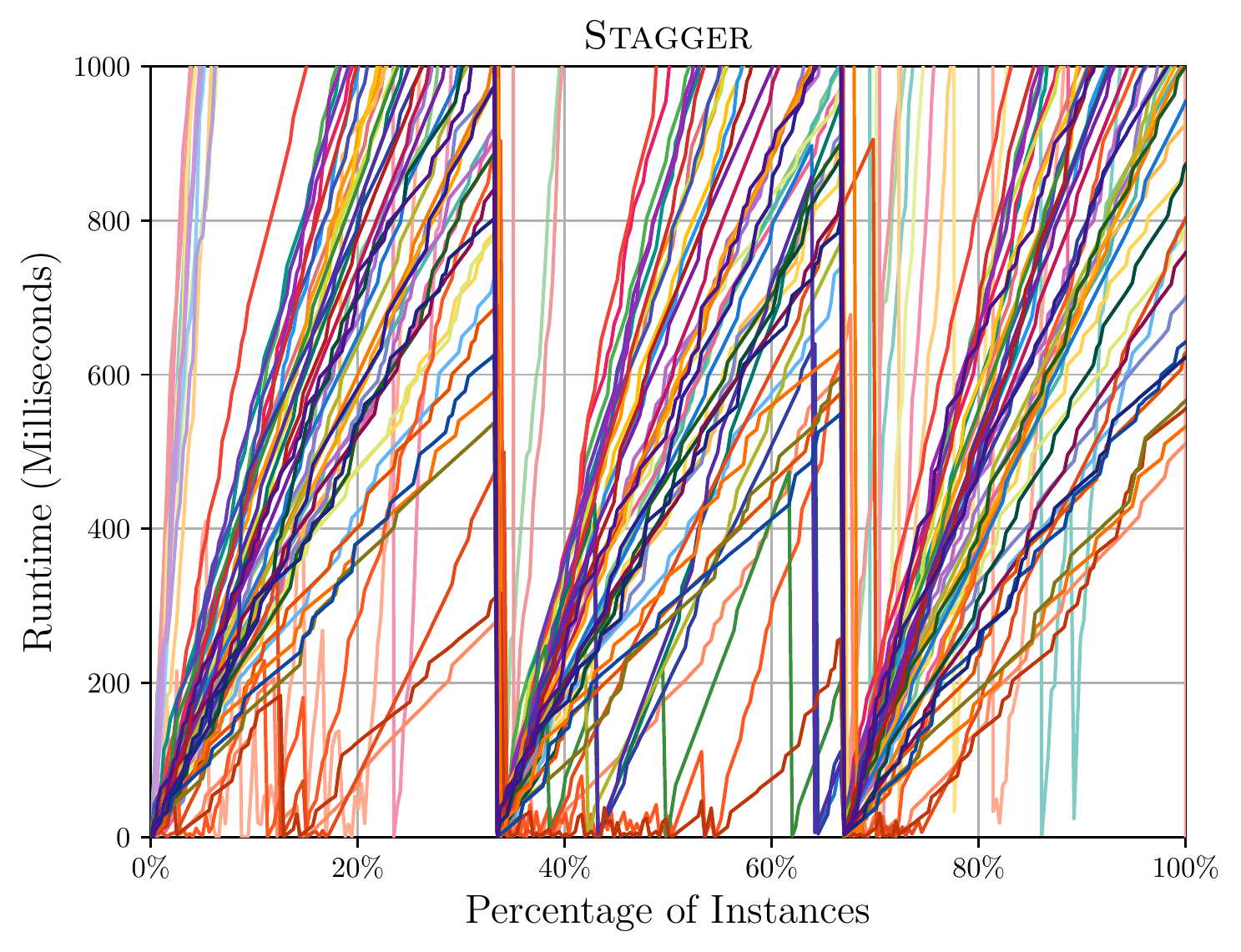}}
	\\ [2em]
	\adjincludegraphics[scale=0.35,trim={0cm 0cm 0 0cm},clip]{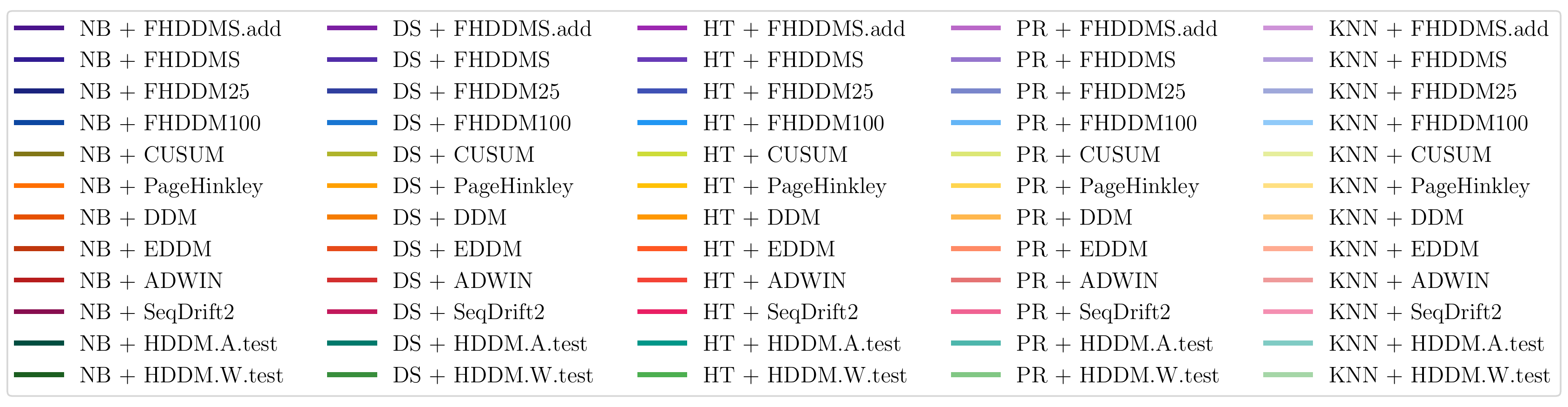}
	\caption{Scores, Error-rates, Memory Usages, and Runtimes of \\ Classifier+Detector Pairs against \textsc{Stagger} Data Stream}
	\label{fig_stagger}
\end{figure}

This subsection details our results against the synthetic data streams, when all the (classifier, detector) pairs are executed in parallel. We focus on the scores, error-rates, memory usages, and runtimes for the \textsc{Stagger} data stream as reported in Fig.\ \ref{fig_stagger}. Recall that the weight vector $\overrightarrow{w} = \begin{bmatrix} w_e & w_d & w_{fp} & w_{fn} & w_{m} & w_{r} \end{bmatrix}^T$ contains the weights associated with the error-rate, the drift detection delay, false positive rates, false negative rates, as well as the memory usage and run-time. When setting them all to 1, it is assumed that they are of the same importance. 
As shown in Fig.\ \ref{fig_stagger} (\hyperref[fig_staggera]{a}), the pairs with Naive Bayes and Perceptron classifiers obtain higher scores; particularly when these two classifiers are paired with FHDDMS, CUSUM, HDDM\textsubscript{A-test} and HDDM\textsubscript{W-test}.
Fig.\ \ref{fig_stagger} (\hyperref[fig_staggerb]{b}) represents the error-rates for (classifier, detector) pairs over time. As illustrated, the pairs of FHDDMS, FHDDMs, and HDDMs obtained the lowest error-rates within each context.
Recall that, a context refers to the interval between two consecutive concept drifts. Fig. \ref{fig_stagger} (\hyperref[fig_staggerc]{c}) indicates that the pairs with Hoeffding Tree use larger amounts of memory. Finally, as shown in Fig. \ref{fig_stagger} (\hyperref[fig_staggerd]{d}), the pairs containing the 5-NN classifier have the longest execution runtimes for locating nearest neighbors.

\par
\begin{figure}[h]
	\centering
	\fontsize{8}{8}\selectfont 
	\textbf{Pair Recommendation over Time - Abrupt Drift} \\
	\textit{(from top-left corner to bottom-right corner)} \\
	\subfloat[\textsc{Sine1}\label{fig_syn_circle_1a}]{\adjincludegraphics[scale=0.55,trim={0 0 0 0.95cm},clip]{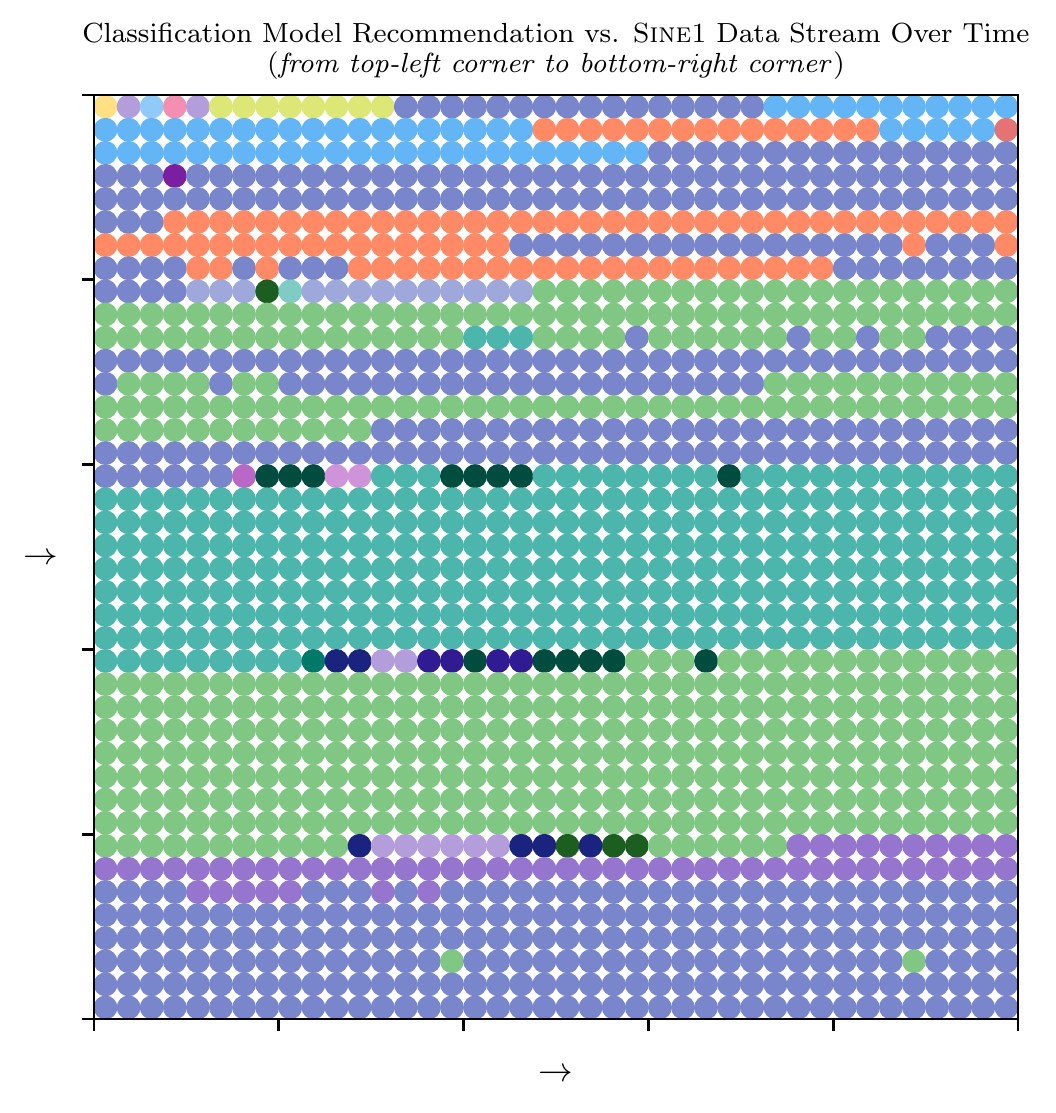}}
	\subfloat[\textsc{Sine2}\label{fig_syn_circle_1b}]{\adjincludegraphics[scale=0.55,trim={0 0 0 0.95cm},clip]{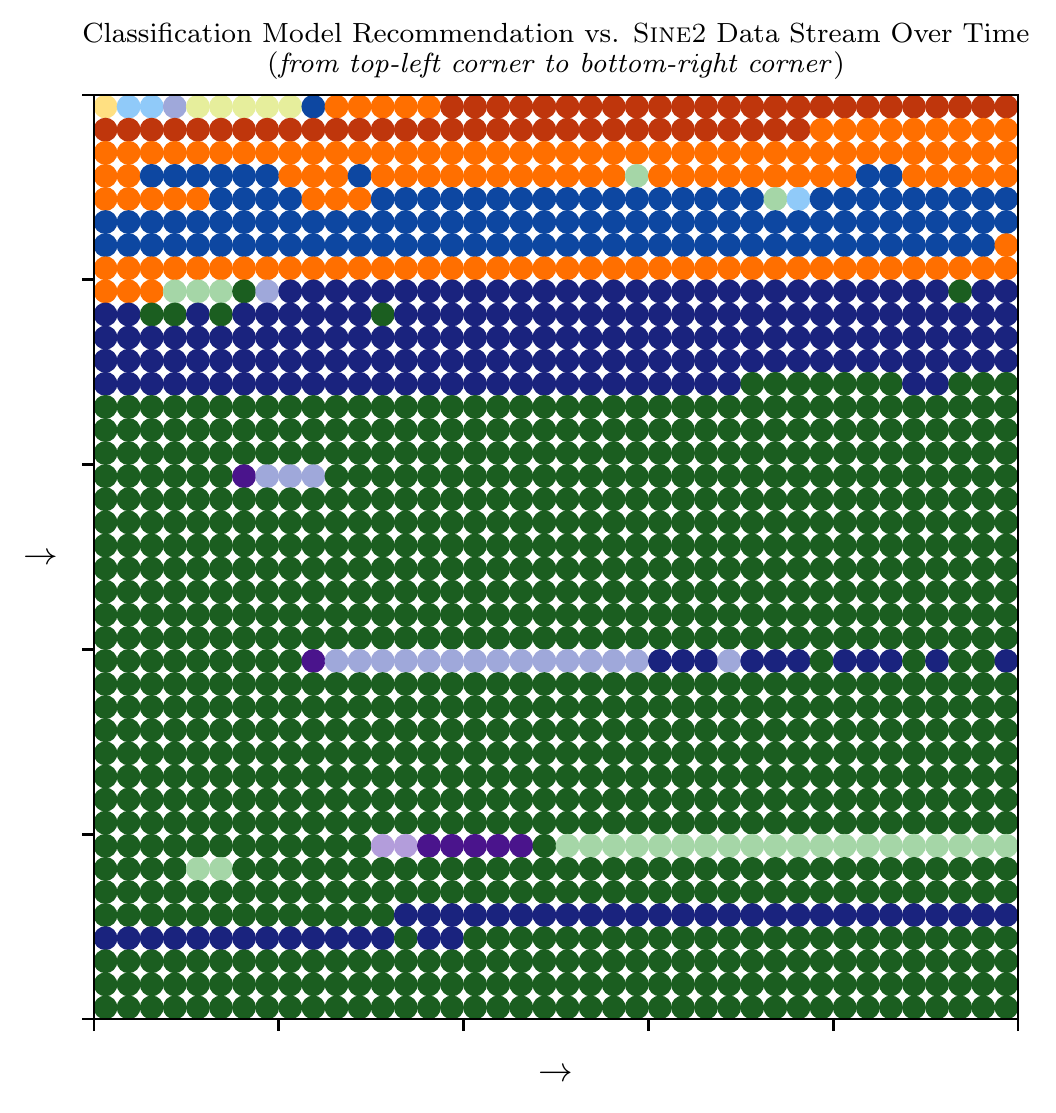}}
	\\
	\subfloat[\textsc{Mixed}\label{fig_syn_circle_1c}]{\adjincludegraphics[scale=0.55,trim={0 0 0 0.95cm},clip]{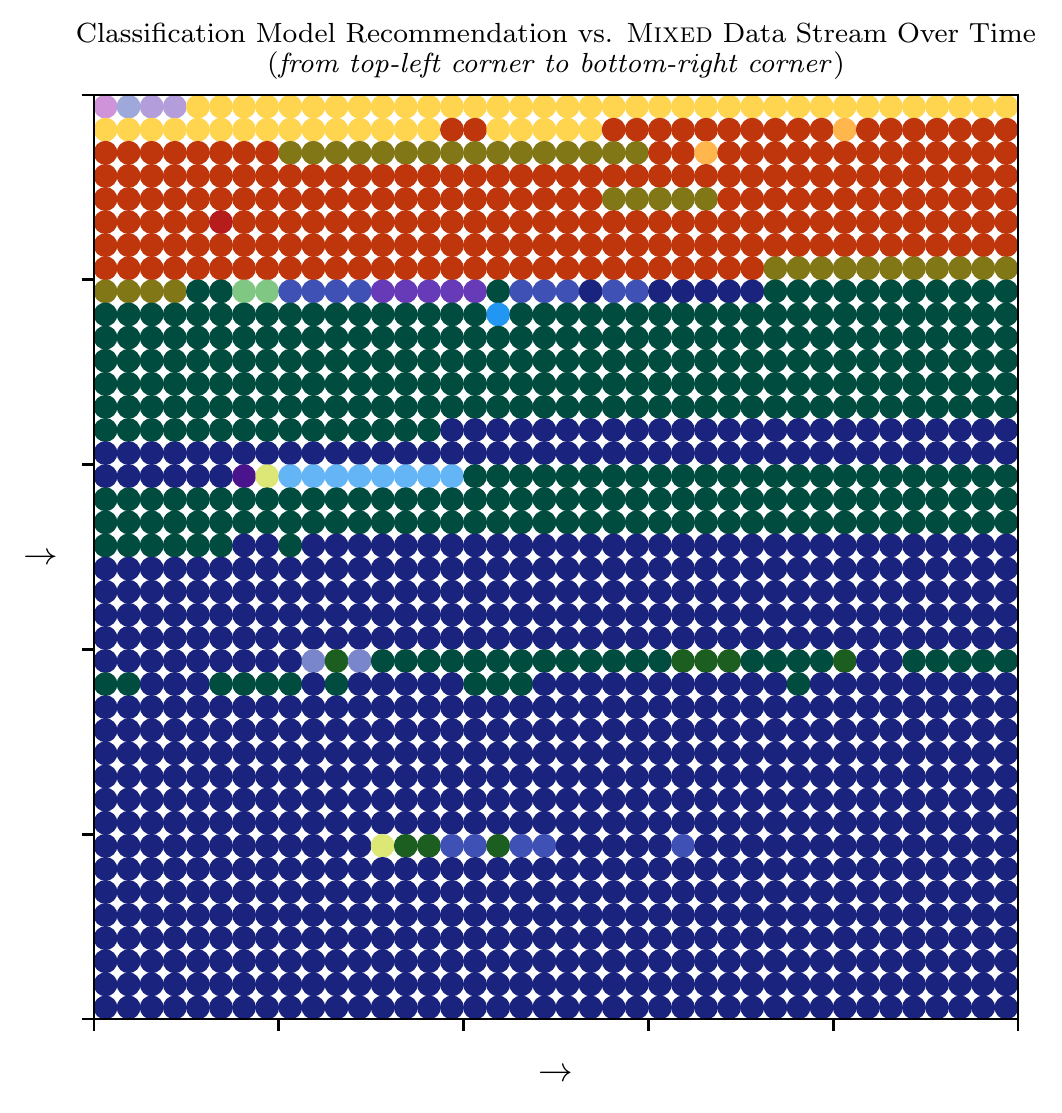}}
	\subfloat[\textsc{Stagger}\label{fig_syn_circle_1d}]{\adjincludegraphics[scale=0.55,trim={0 0 0 0.95cm},clip]{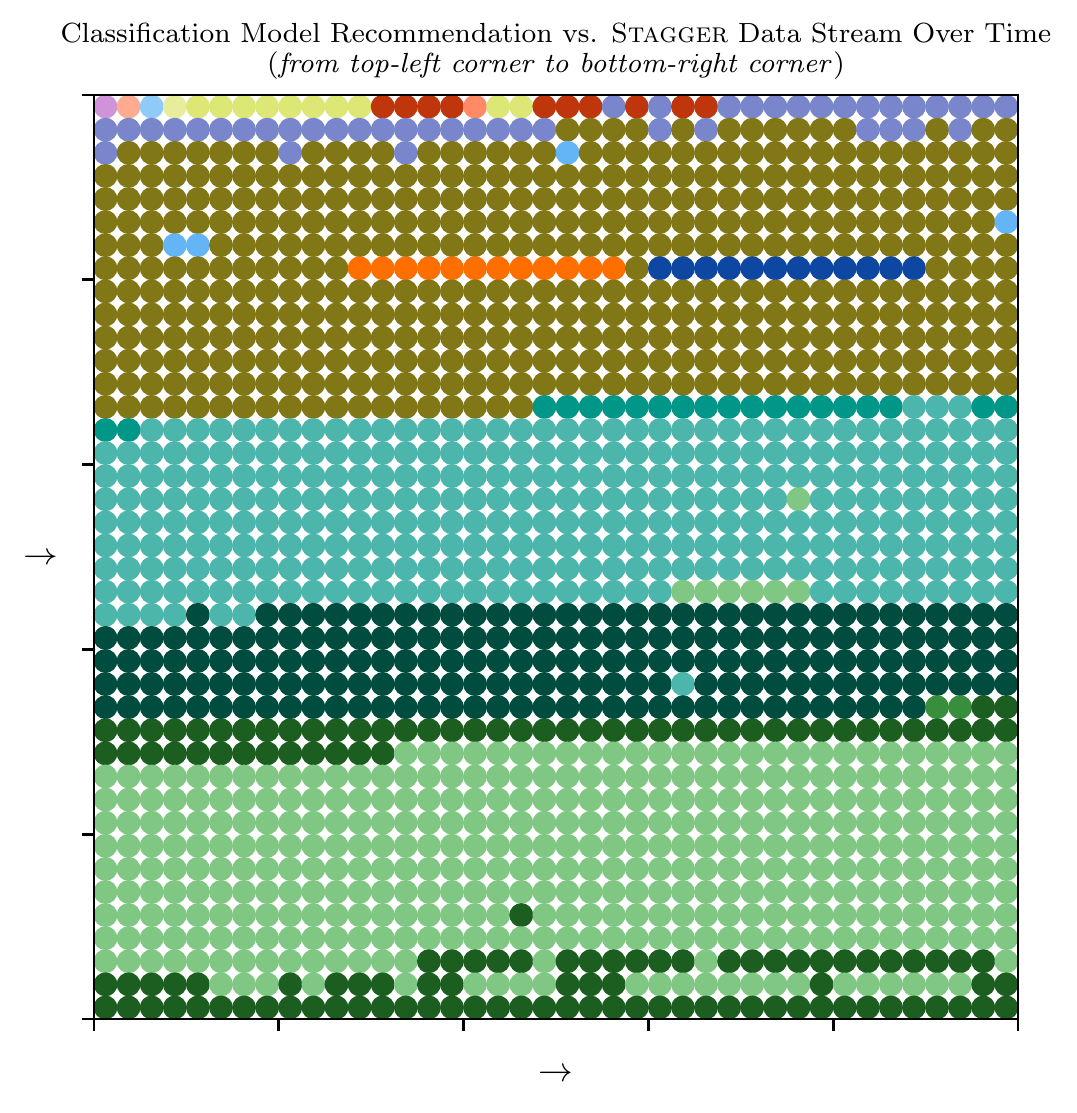}}
	\\ [2em]
	\adjincludegraphics[scale=0.35,trim={0cm 0cm 0 0cm},clip]{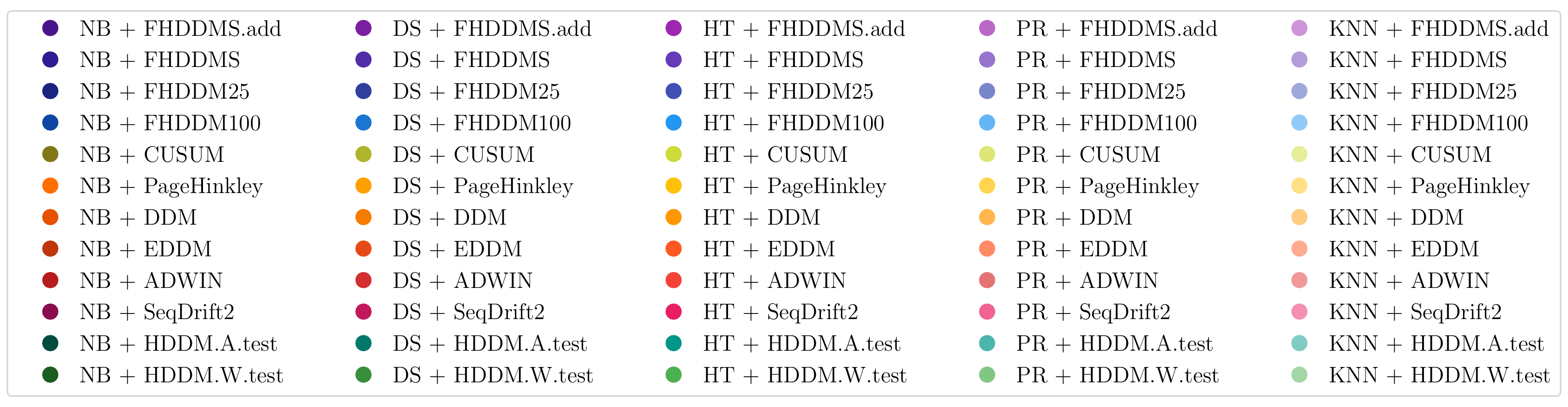}
	\caption{Classifier+Detector Recommendation against \\ Synthetic Data Streams with Abrupt Concept Drifts}
	\label{fig_syn_circle_1}
\end{figure}

\begin{figure}[ht]
	\centering
	\fontsize{8}{8}\selectfont 
	\textbf{Pair Recommendation over Time - Gradual Drift} \\
	\textit{(from top-left corner to bottom-right corner)} \\
	\fontsize{8}{8}\selectfont 
	\subfloat[\textsc{Circles}\label{fig_syn_circle_2a}]{
		\stackon[3pt]{\adjincludegraphics[scale=0.55,trim={0 0 0 0.95cm},clip]{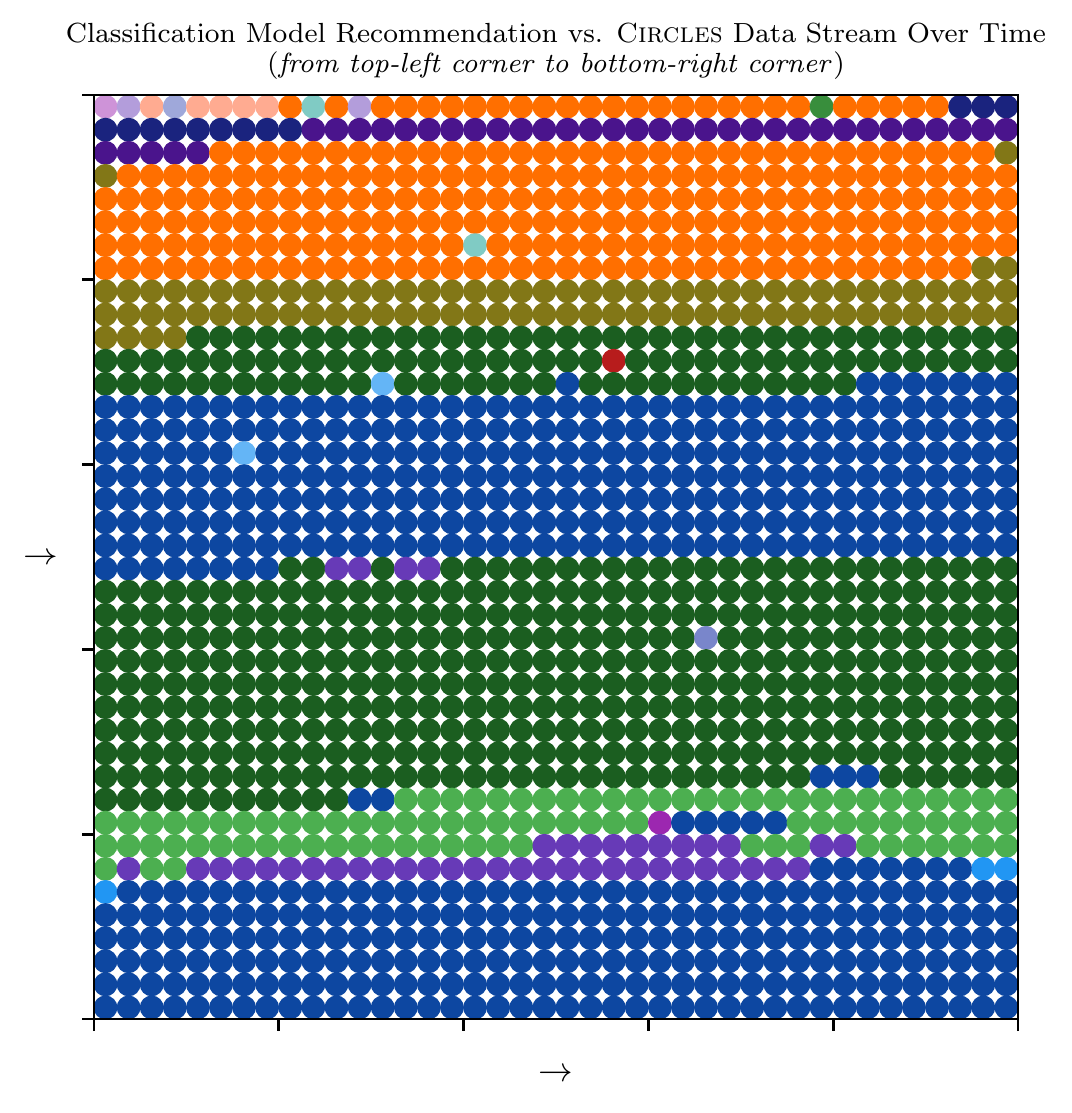}}{$\overrightarrow{w} = \mbox{[1 1 1 1 1 1]}^T$}
		\stackon[3pt]{\adjincludegraphics[scale=0.55,trim={0 0 0 0.95cm},clip]{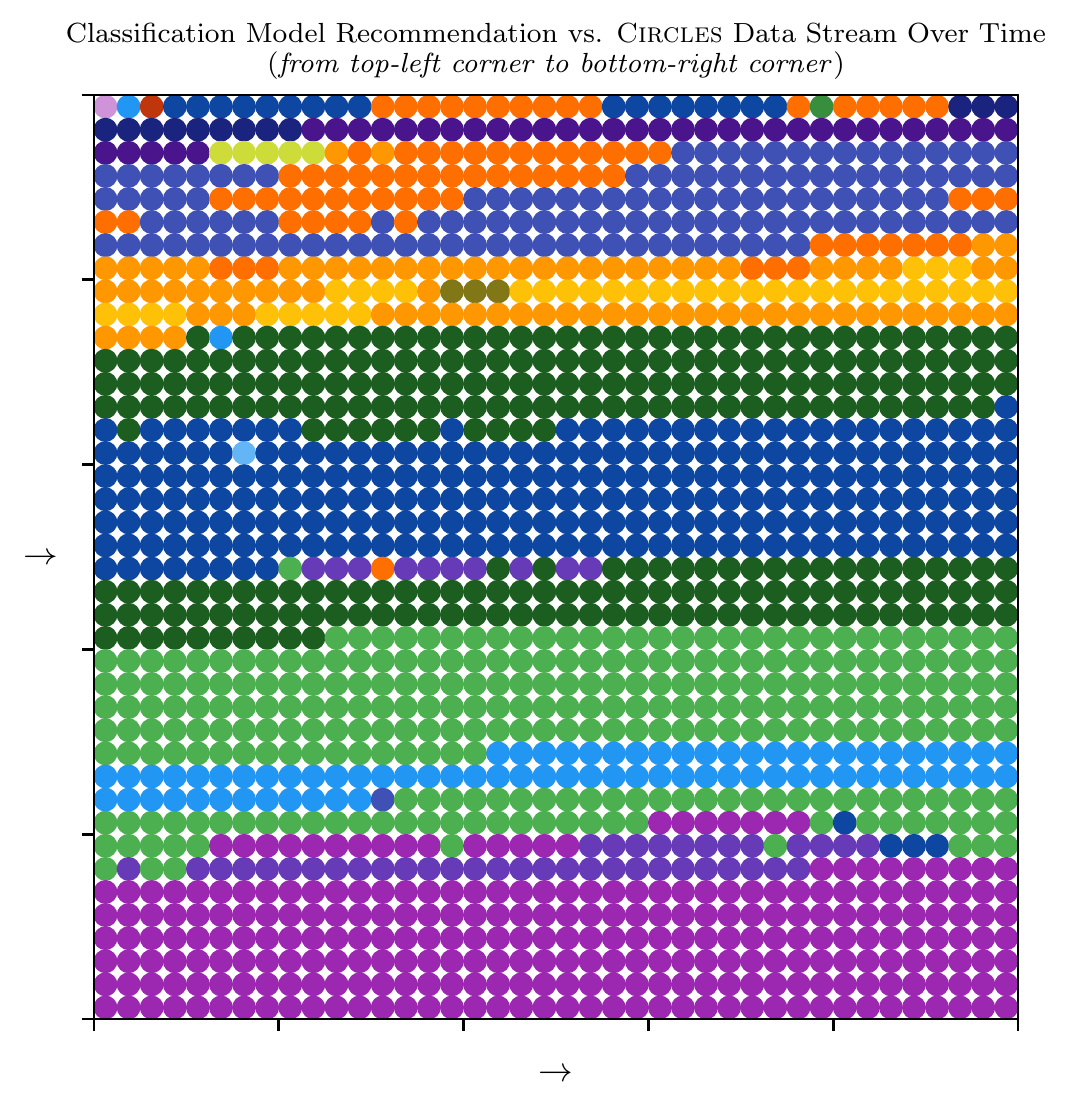}}{$\overrightarrow{w} = \mbox{[1.5 1 2 1.5 0 0.5]}^T$} 
	}
	\\
	\subfloat[\textsc{LED}\label{fig_syn_circle_2b}]{
		\stackon[3pt]{\adjincludegraphics[scale=0.55,trim={0 0 0 0.95cm},clip]{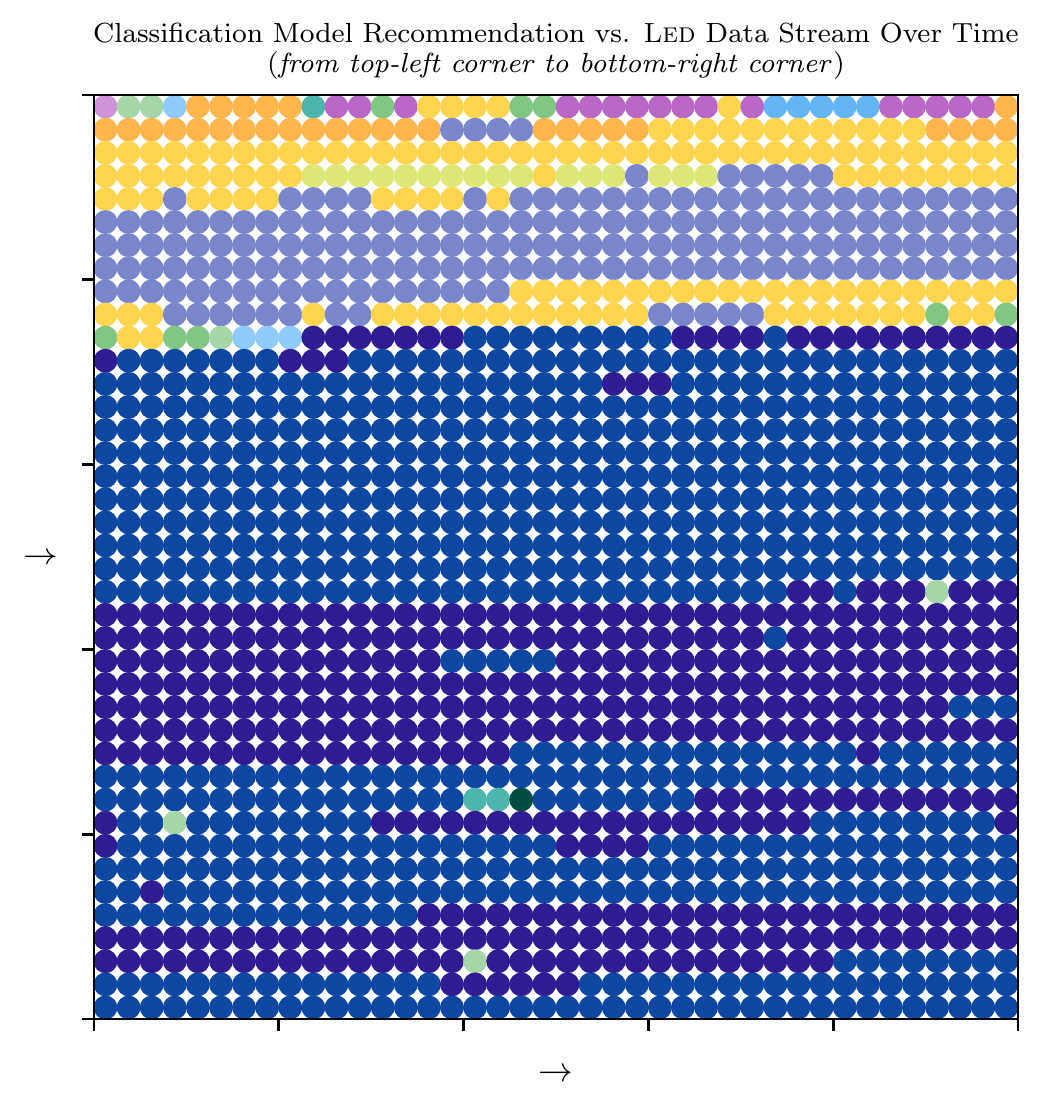}}{$\overrightarrow{w} = \mbox{[1 1 1 1 1 1]}^T$}
		\stackon[3pt]{\adjincludegraphics[scale=0.55,trim={0 0 0 0.95cm},clip]{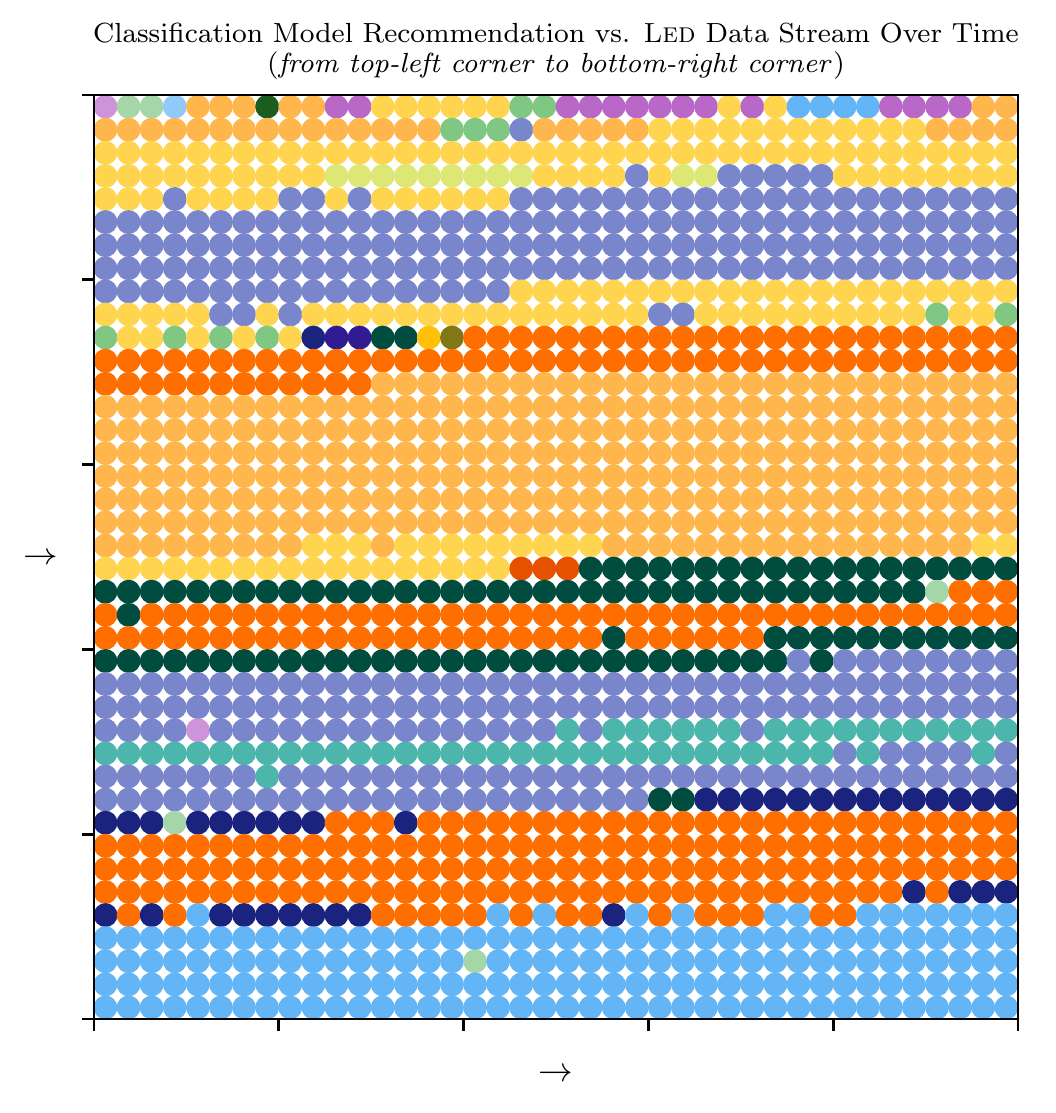}}{$\overrightarrow{w} = \mbox{[3 0 1.5 1 2 2]}^T$}
	}
	\\ [1em]
	\adjincludegraphics[scale=0.35,trim={0 0 0 0},clip]{fig_legend_circle.pdf}
	\caption{Classifier+Detector Recommendation against \\ Synthetic Data Streams with Gradual Concept Drifts}
	\label{fig_syn_circle_2}
\end{figure}

 The pairs recommended over time by the \textsc{Tornado} {framework} against the synthetic data streams with abrupt concept drifts are indicated in Fig.\ \ref{fig_syn_circle_1}. Each circle represents the pair suggested to the user by the system at a specific moment in time. The time line begins at the top-left corner and is then unfolded line by line. In this set of experiments, all weights were set to one.
Note that the pairs that use members of the FHDDM family as drift detectors are indicated in shades of blue, while the HDDM variants are displayed in shades of green. The other drift detector pairs are displayed in shades of yellow (CUSUM and PageHinkley), orange (DDM and EDDM), red (ADWIN), and pink (SeqDrift). 
The results confirm that no one pair outperform the others, and that no drift detector or classifier dominates.
Initially, there are larger fluctuations in recommended pairs.  In summary, Fig.\ \ref{fig_syn_circle_1} shows that, in general, the pairs with the FHDDMS and HDDM drift detectors are often ranked as highest. The results also show that the Naive Bayes and Perceptron classifiers are often preferred, since they are light, fast, and accurate, particularly when the weights are set to equal. This is,  however,  not the case when the weights are varied, as shown in Fig.\ \ref{fig_syn_circle_2}.
Fig. \ref{fig_syn_circle_2} (\hyperref[fig_syn_circle_2a]{a}) illustrates the impact of the weight vector against the \textsc{Circles} and \textsc{LED} data streams, that are susceptible to gradual drift. The pairs of NB+PageHinkley, NB+FHDDM\textsubscript{100}, and NB+HDDM\textsubscript{W-test} outperform the others for the \textsc{Circles} data stream when all the weights are set to one. In contrast, when $\overrightarrow{w} = \mbox{[1.5 1 2 1.5 0 0.5]}^T$, the pairs of HT+HDDM\textsubscript{W-test} and HT+FHDDM\textsubscript{add} dominate the others almost for the second of half of the stream. In this case, the memory and runtime become less important, and the resulting best pair reflects this change.
That is, the entries were chosen so that the pairs with lower values for the error-rate, shorter detection delay, fewer false positives and false negatives obtain higher scores. Memory consumption was not taken into account as $w_m = 0$.
Finally, Fig.\ \ref{fig_syn_circle_2} (\hyperref[fig_syn_circle_2a]{b}) depicts that  the pairs of PR+FHDDM\textsubscript{25}, NB+FHDDMS, and NB+FHDDM\textsubscript{100} are recommended over time against the LED data stream; when all the weights are equal to one. Alternatively, when the weight entries are set as $\overrightarrow{w} = \mbox{[3 0 1.5 1 2 2]}^T$, the pairs of PR+FHDDM\textsubscript{25}, PR+FHDDM\textsubscript{100}, PR+PageHinkley, and PR+DDM are recommended. In this case, memory usage and runtime are of two preferred measures, and subsequently the Perceptron classifier outperformed the Naive Bayes one.

\subsubsection{Real-world Data Streams}

\begin{figure}[h]
	\centering
	\fontsize{8}{8}\selectfont 
	\vspace*{5em}
	\subfloat[\textsc{Adult}]{\adjincludegraphics[scale=0.5,trim={0 0 0 0},clip]{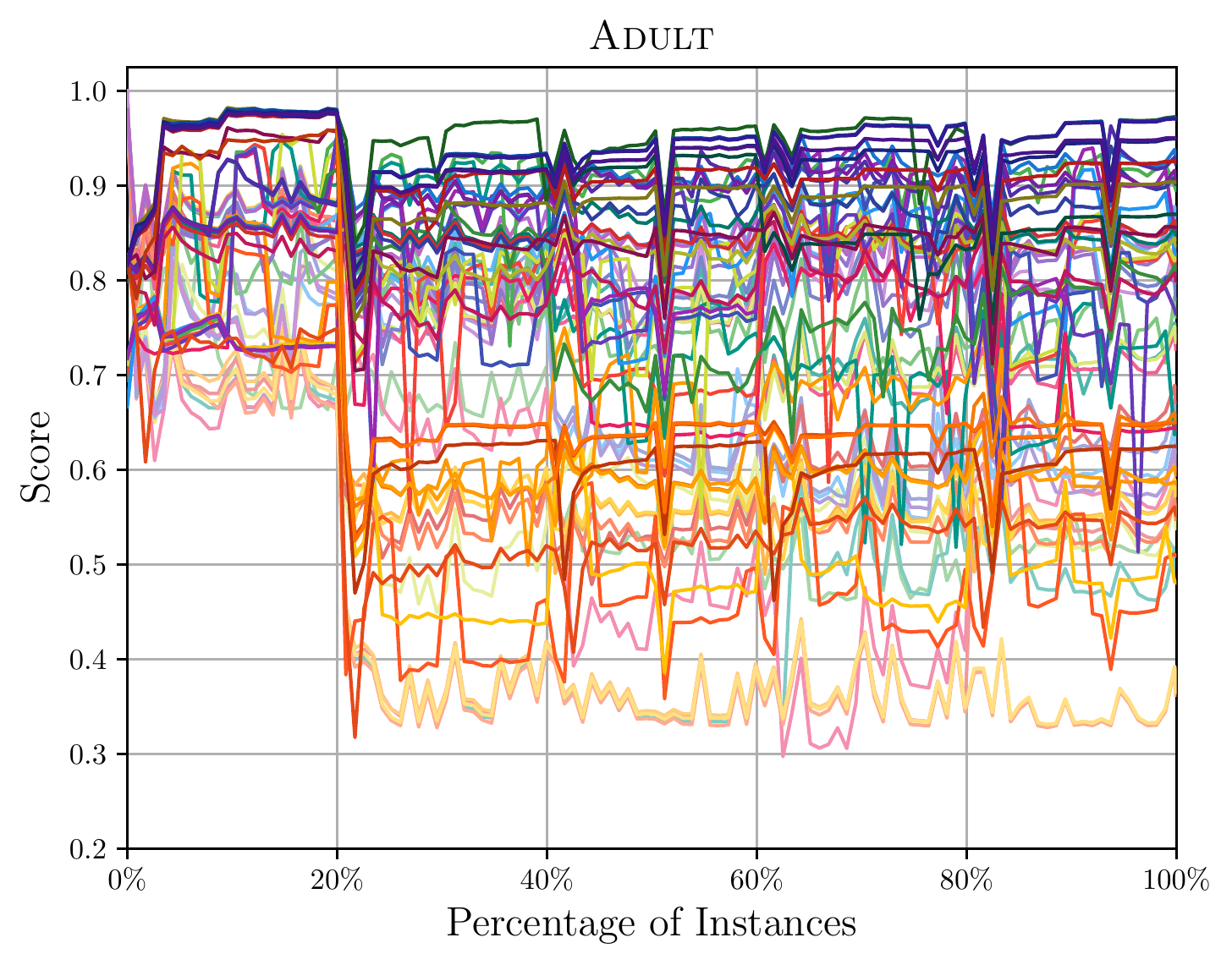}} \\
	\subfloat[\textsc{Nursery}]{\adjincludegraphics[scale=0.4,trim={0 0 0 0},clip]{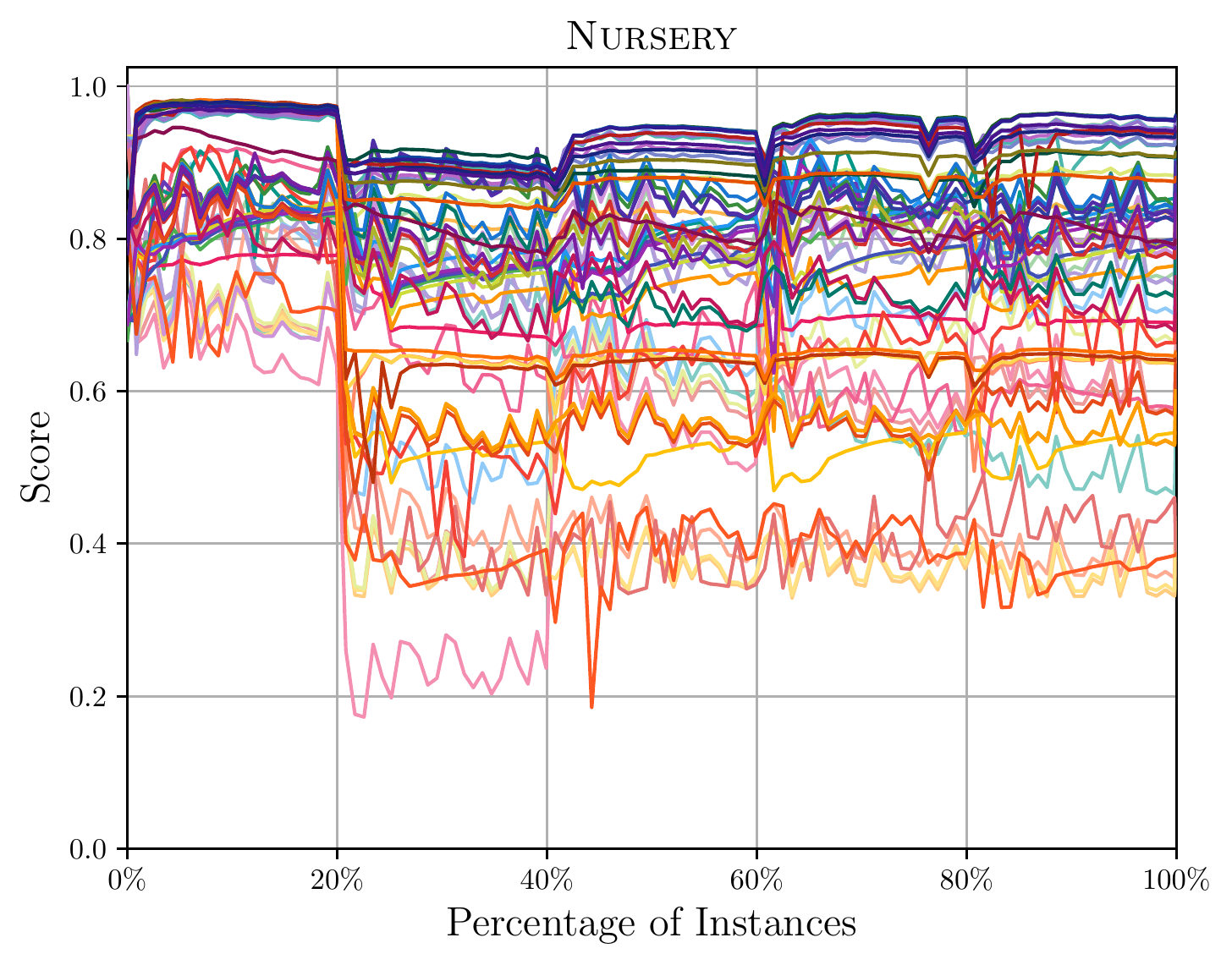}}
	\subfloat[\textsc{Shuttle}]{\adjincludegraphics[scale=0.4,trim={0 0 0 0},clip]{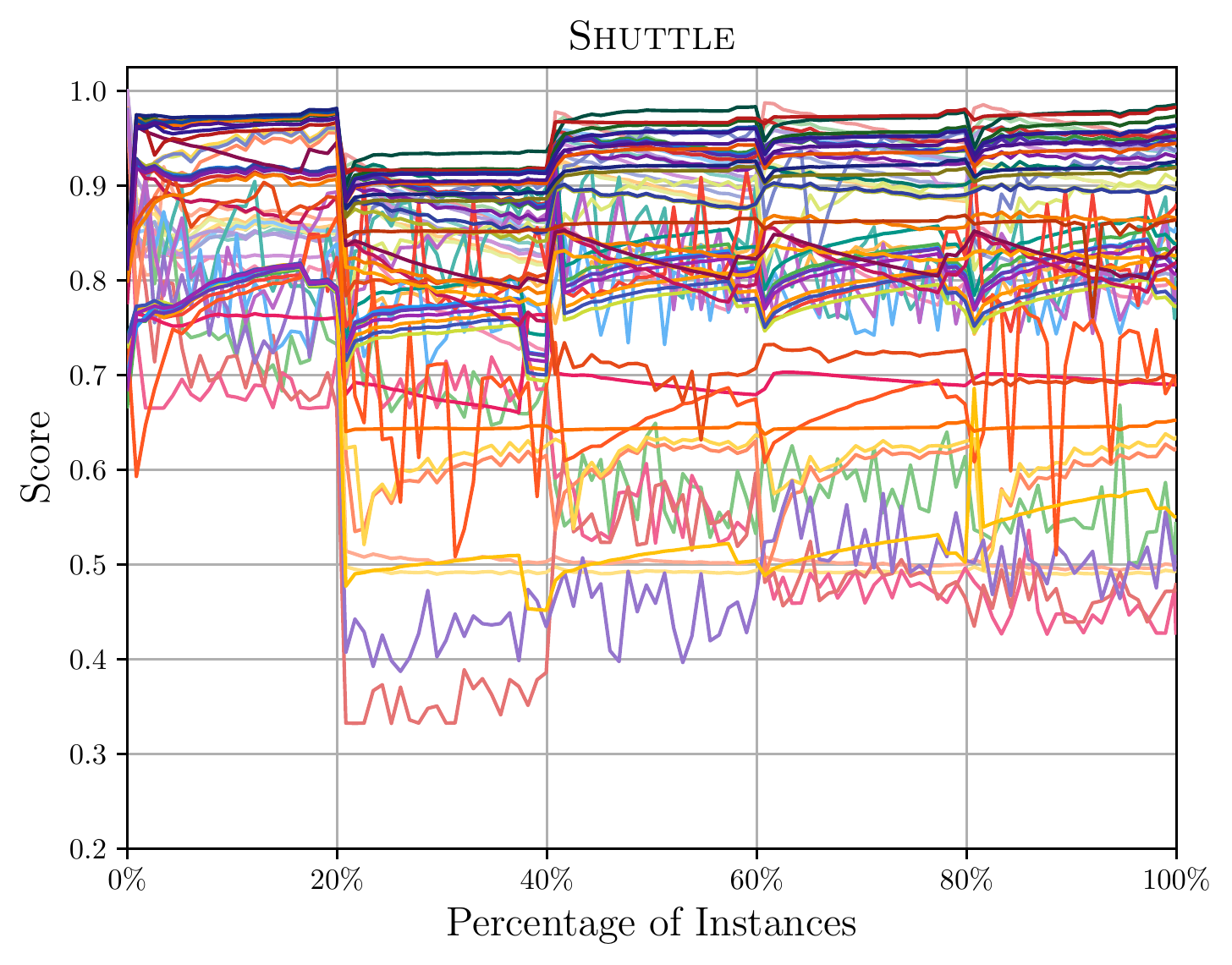}}
	\\ [1em]
	\adjincludegraphics[scale=0.35,trim={0 0 0 0},clip]{fig_legend_line.pdf}
	\caption{Scores of Classifier+Detector Pairs against \\ Real-world Data Streams}
	\label{fig_real_score}
\end{figure}

\begin{figure}[h]
	\centering
	\fontsize{8}{8}\selectfont 
	\textbf{Pair Recommendation over Time} \\
	\textit{(from top-left corner to bottom-right corner)} \\
	
	\subfloat[\textsc{Adult}\label{fig_real_adult}]{
		\stackon[3pt]{\adjincludegraphics[scale=0.5,trim={0 0 0 0.95cm},clip]{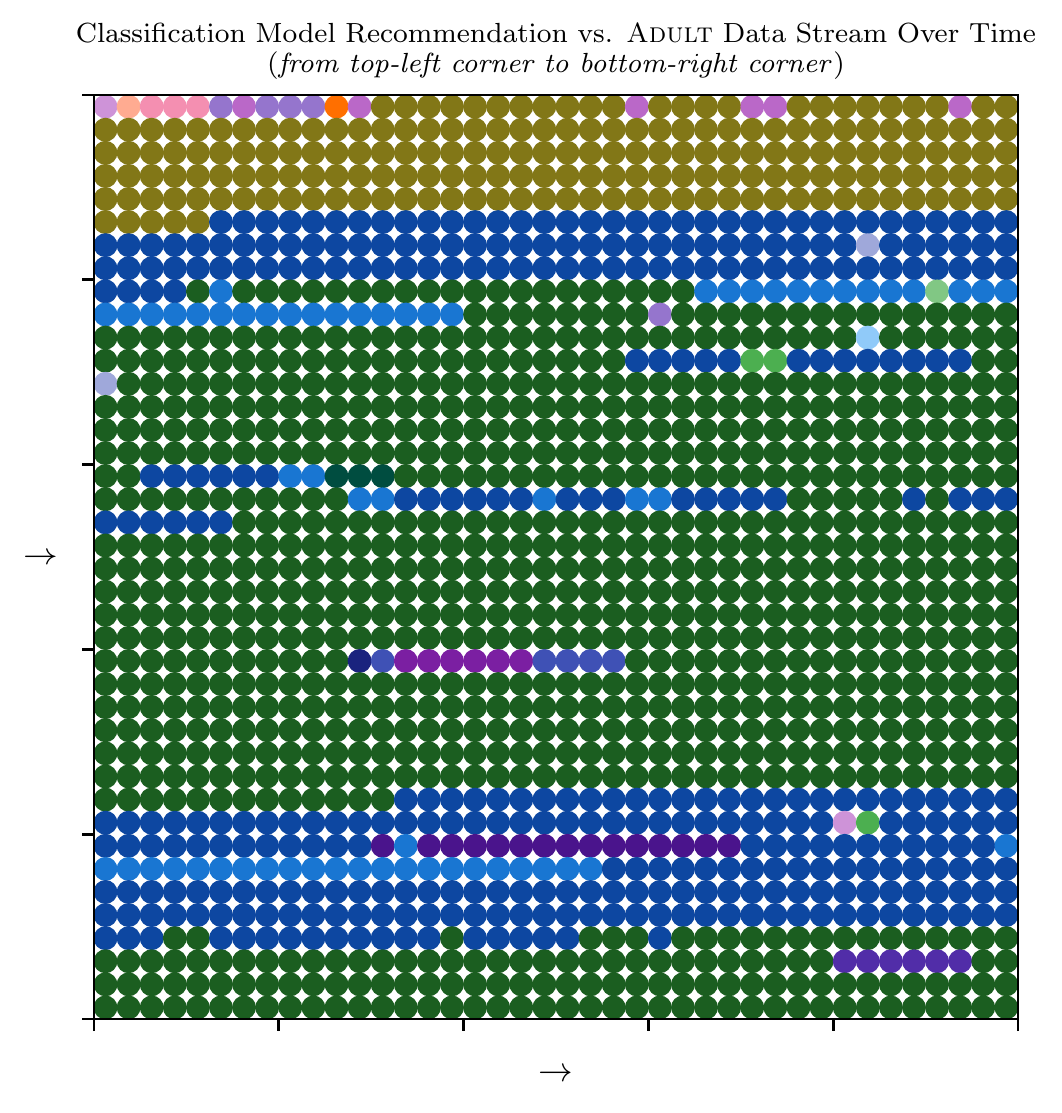}}{$\overrightarrow{w} = \mbox{[1 1 1 1 1 1]}^T$}
		\stackon[3pt]{\adjincludegraphics[scale=0.5,trim={0 0 0 0.95cm},clip]{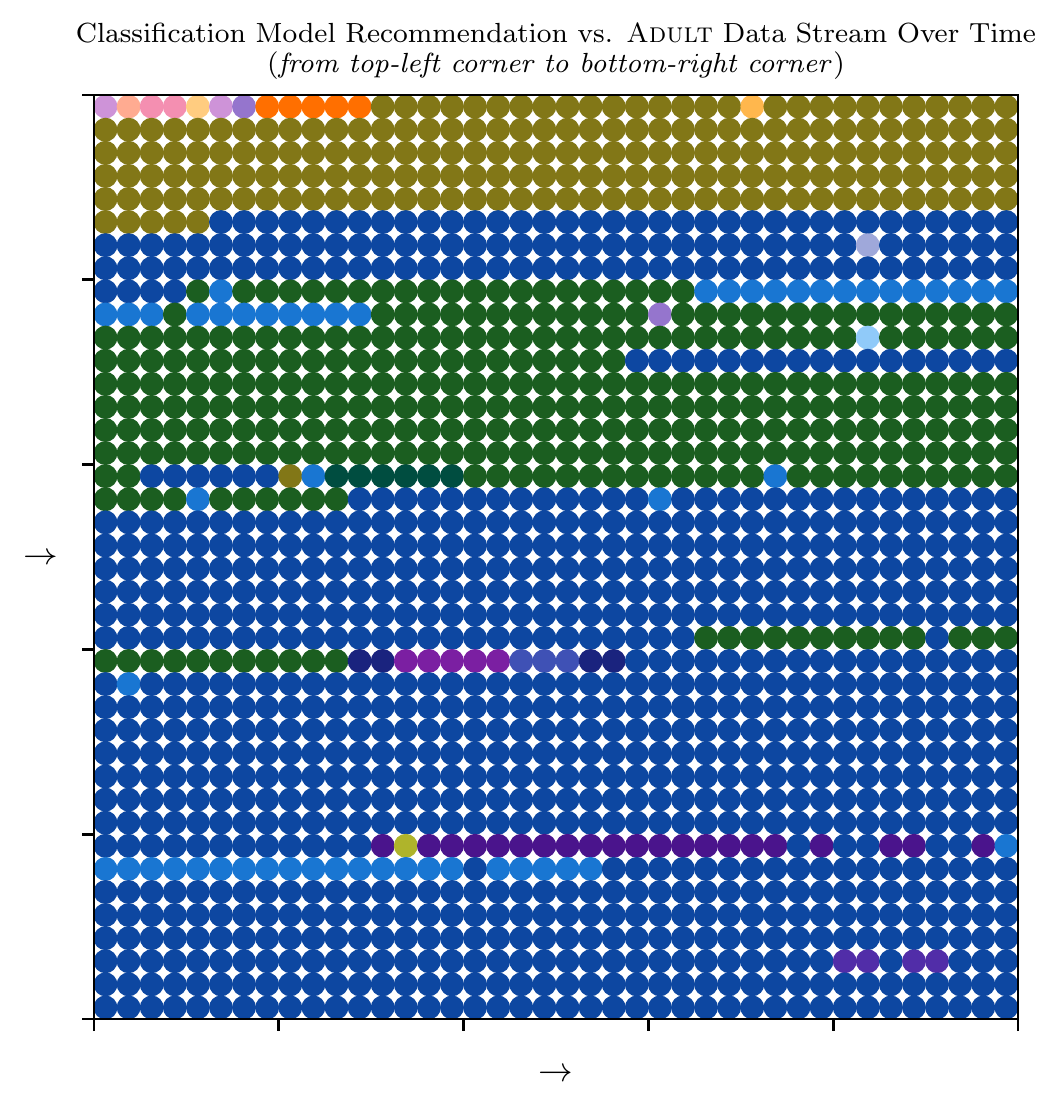}}{$\overrightarrow{w} = \mbox{[2 1.5 3 2 1.5 1]}^T$} 
	}
	\\
	
	\subfloat[\textsc{Nursery}\label{fig_real_nursery}]{
	\stackon[3pt]{\adjincludegraphics[scale=0.5,trim={0 0 0 0.95cm},clip]{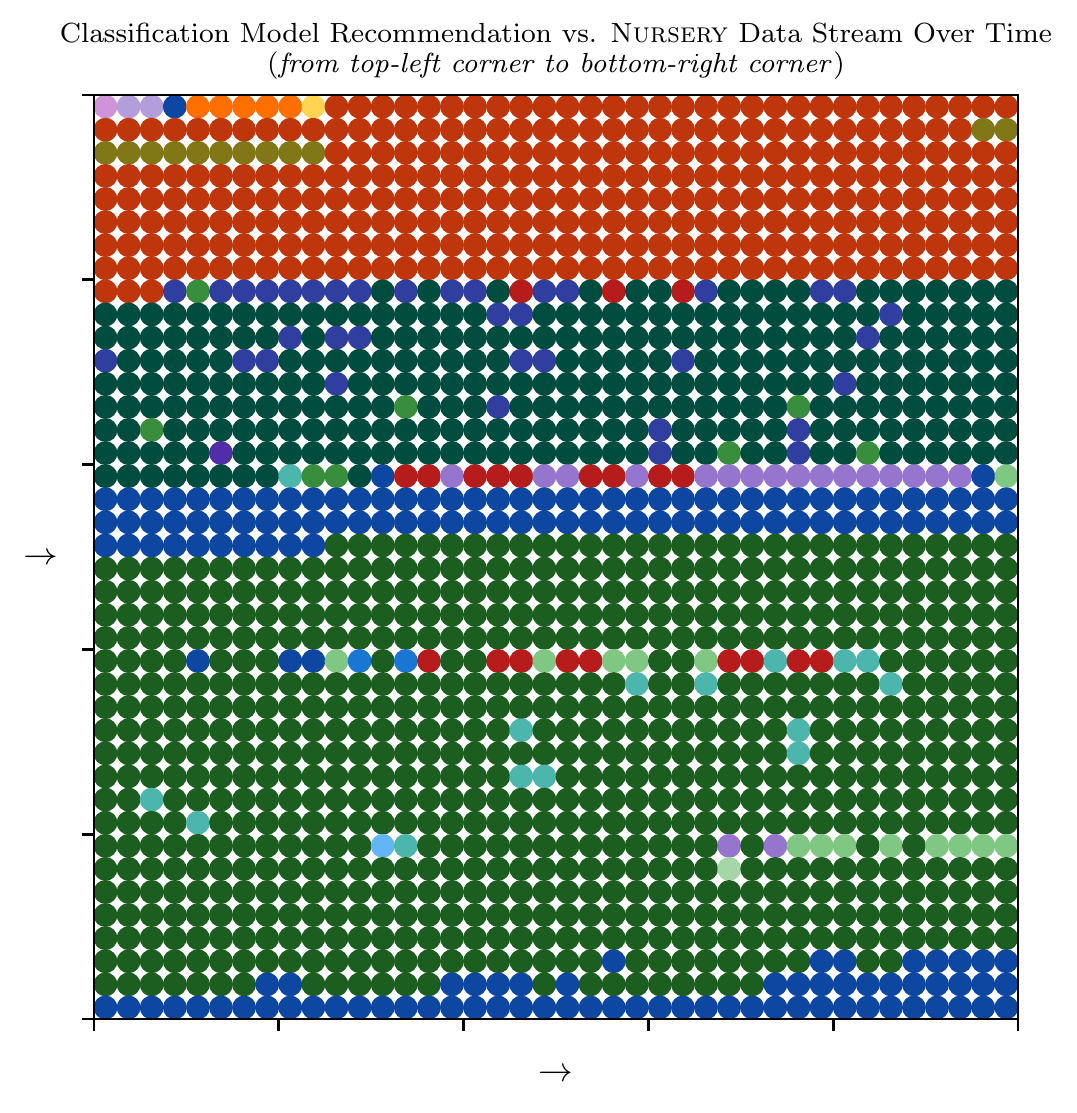}}{$\overrightarrow{w} = \mbox{[1 1 1 1 1 1]}^T$}
	\stackon[3pt]{\adjincludegraphics[scale=0.5,trim={0 0 0 0.95cm},clip]{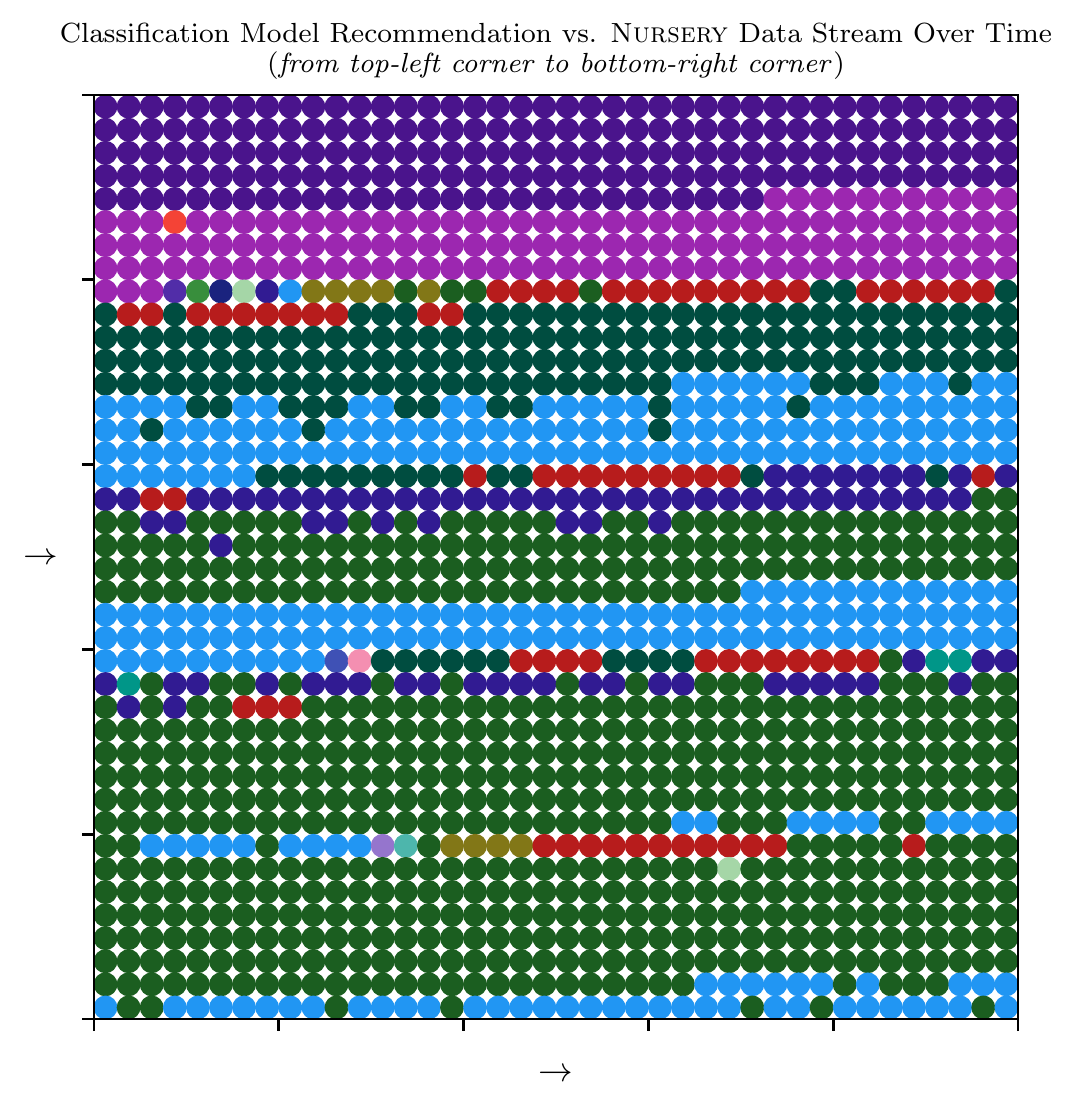}}{$\overrightarrow{w} = \mbox{[3 1 2 3 0 0]}^T$} 
	}
	\\
	[1em]
	\adjincludegraphics[scale=0.35,trim={0 0 0 0},clip]{fig_legend_circle.pdf}
	\caption{Classifier+Detector Recomm.\ against\ Real-world Data Streams (1)}
	\label{fig_real_circle}
\end{figure}

\begin{figure}[h]
	\centering
	\fontsize{8}{8}\selectfont
	\vspace{2em}
	\textbf{Pair Recommendation over Time} \\
	\textit{(from top-left corner to bottom-right corner)} \\
	\subfloat[\textsc{Shuttle}\label{fig_real_shuttle}]{
		\stackon[3pt]{\adjincludegraphics[scale=0.5,trim={0 0 0 0.95cm},clip]{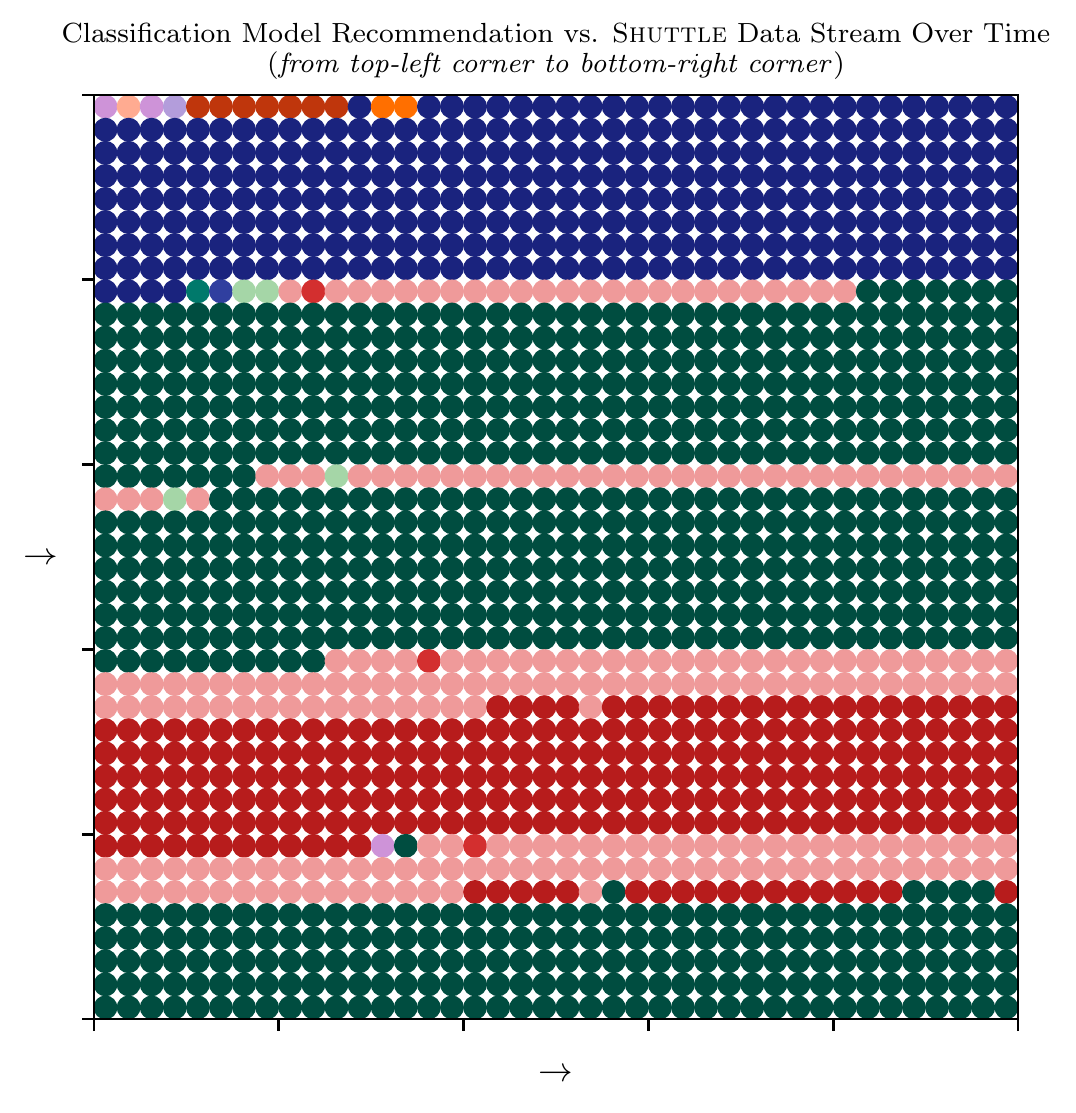}}{$\overrightarrow{w} = \mbox{[1 1 1 1 1 1]}^T$}
		\stackon[3pt]{\adjincludegraphics[scale=0.5,trim={0 0 0 0.95cm},clip]{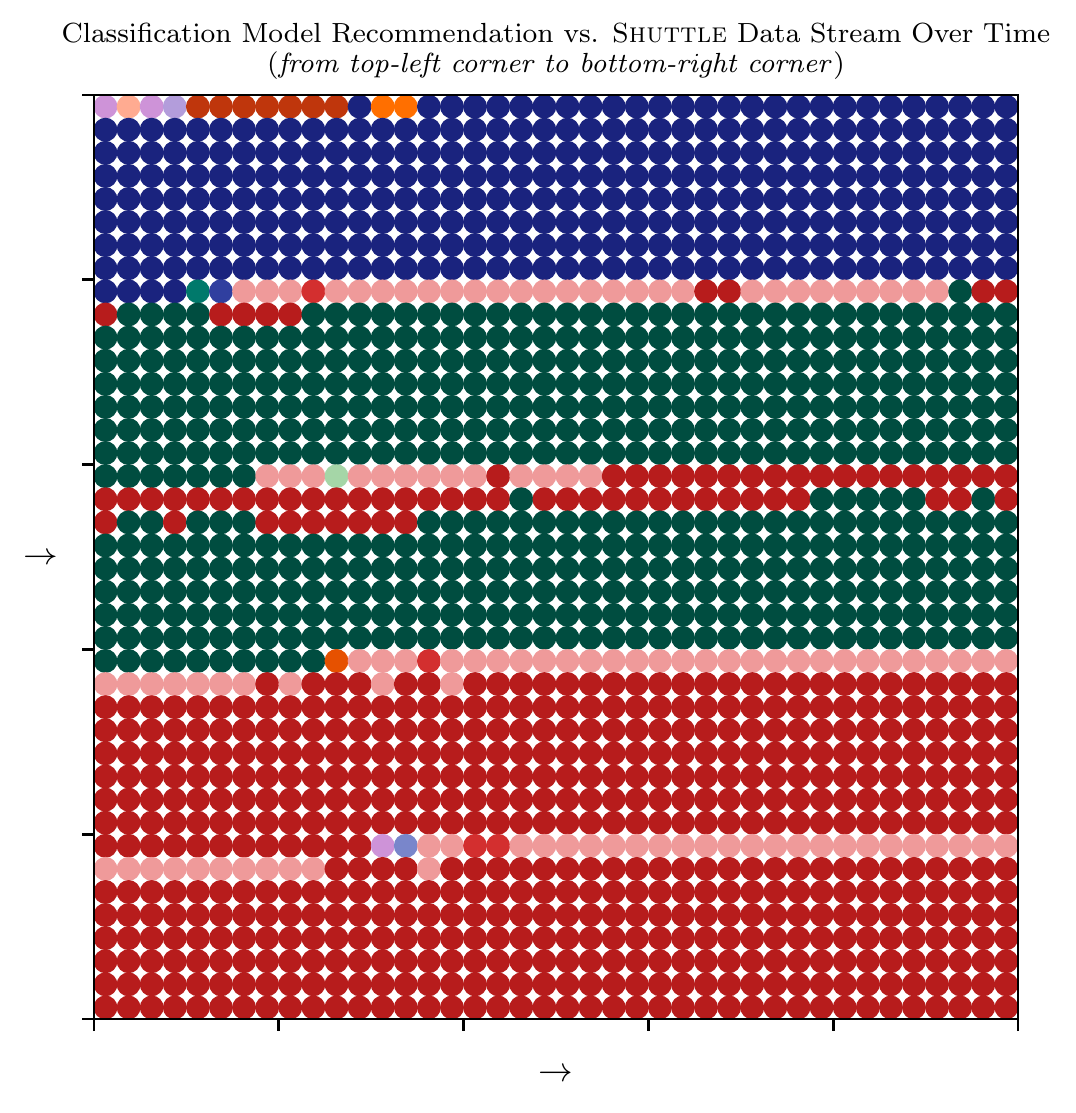}}{$\overrightarrow{w} = \mbox{[3 2 1 2 1 1]}^T$} 
	}
	\\ [1em]
	\adjincludegraphics[scale=0.35,trim={0 0 0 0},clip]{fig_legend_circle.pdf}
	\caption{Classifier+Detector Recomm.\ against\ Real-world Data Streams (2)}
	\label{fig_real_circle_continue}
\end{figure}

Our experimental results for the real-world data streams are reported in Fig. \ref{fig_real_score} to \ref{fig_real_circle_continue}. Fig.\ \ref{fig_real_score} shows the scores for all the (classifier, detector) pairs, as instances are processed, over time. The reader will again notice that no single pair outperforms in all cases, and that the best current pair changes over time. The figures also indicate that the best pairs rapidly changes at the beginning of the stream, but that the optimal pairs remain more steady towards the ends of the streams. A variation of the weights impact the recommendation, as is noted when comparing the left and right sides of the Fig.\ \ref{fig_real_circle} and \ref{fig_real_circle_continue}. Again, the best pair is highly dependent on the weights, notably when we vary the weights of memory and runtime versus error-rate considerations.

\clearpage

\section{Discussion}
\label{sec_discussion}

We introduced the CAR measure in order to monitor the overall performance of adaptive classification models against evolving data streams. 
We also presented the \textsc{Tornado} as a framework that simultaneously runs heterogeneous pairs of classifiers and drift detectors in parallel against data streams, while continuously recommending the best performing pair to the user. This recommendation is based on the weights assigned to the error-rates, drift detection sensitivity, runtime and memory consumption.

In addition, we extended our earlier work and detailed FHDDMS as well as its extension FHDDMS\textsubscript{add} in order to better detect abrupt concept drifts associated with shorter delay as well as to reduce the number of false negatives when a gradual drift is present. FHDDMS slides a long and a short window, that are stacked on each other, to detect concept drifts. The longer window reduces the number of false negatives, while the shorter one detects drifts faster. In this study, we restricted ourselves to two windows, though more windows could be employed. During the evaluation of drift detection methods, we observed that HDDM\textsubscript{W-test}  and our FHDDMS and FHDDM algorithms are comparable, in terms of various performance measures. HDDM\textsubscript{W-test} outperformed FHDDMS for faster detection against abrupt concept drifts. On the other hand, FHDDMS was better suited to detect the gradual concept drifts. In either case, HDDM\textsubscript{W-test} showed higher false positive rates compared to FHDDMS and FHDDM.

We conducted experiments using the \textsc{Tornado} framework against synthetic and real-world data streams. Our experimental setup consisted of 60 pairs of learners and detectors, each of which were evaluated in parallel against various data streams. 
The experimental results clearly show, as expected, that no specific pair dominates in all cases. In the vast majority of cases, the pairs that contains the Naive Bayes or Perceptron classifiers yielded the best results, when all the weights are equal. These two algorithms provides a balance between memory usage, runtime and error-rate. This stands in contrast to the Hoeffding Tree, Decision Stump and learners. The Hoeffding Tree algorithm generally is expensive, in terms of memory consumption as the tree grows. Another disadvantage is that the runtime may increase when branching decisions become difficult. The K-NN algorithm is a lazy learner, and this property means that it has an extensive memory usage and runtime.
These two method are therefore more suitable when runtime and memory considerations are of less importance. 
Our experimental results confirm these trends, as notices when studying the evolution depicted in Fig. \ref{fig_stagger} to \ref{fig_real_circle_continue}.

Overall, as represented in Fig.\ \ref{fig_syn_circle_1} and \ref{fig_syn_circle_2}, the pairs of HDDM\textsubscript{W-test} are ranked higher than FHDDM for data stream containing abrupt drifts, e.g.\ \textsc{Sine2} and \textsc{Stagger}; whereas, the pairs of FHDDMS and FHDDM were recommended for the data streams with gradual concept drift; e.g.\ the LED data stream. It is worth to mention that the pairs of other drift detectors ranked lower, because of their longer drift detection delays and higher false positive rates.



\section{Conclusion and Future Work}
\label{sec_con_fut}

Increasingly, there is a need for near real-time adaptive learning methods to explore dynamically evolving data streams. Such algorithms should provide decision makers with realistic, just-in-time models for short-term and mid-term decision making against today's vast streams of data. These models should not only be timely, but also be accurate and able to swiftly adapt to changes in the data. Intuitively, no adaptive learning strategy outperforms others in all settings. Similarly, the effectiveness of drift detection methods is determined by the data characteristics, the types of drifts and the rates of true positives and true negatives, among others. 

Based on these observations, we created a reservoir of diverse adaptive learners and drift detection algorithms, as implemented in our \textsc{Tornado} framework. In our work, we consider all (classifier, detector) pairs and then utilized them to construct models in parallel. Continuously, the current `best' model is selected and provided to the users.  Further, two new drift detector methods, namely the FHDDMS and FHDDMS\textsubscript{add} algorithms, were introduced in this paper. Our extensive experimental results confirm that the current (classifier, detector) pairs vary over time and that they are sensitive to concept drift. Further, we show that the two FHDDMS variants outperform the state-of-the-art, when evaluated in terms of the holistic CAR measure. 

We encountered several interesting avenues of future work. In future, we plan to also compare our \textsc{Tornado} framework to existing ensembles of classifiers. The incorporation of ensembles into the framework, also needs our consideration. In our current research, we implemented our \textsc{Tornado} framework on a single machine. We are now designing a hybrid environment where we utilize Cloud services together with mobile devices, such as tablets. This current research is motivated by our observation that, in many settings, such as environmental impact studies and emergency response, domain experts would require the ability to not only receive up-to-date models on their mobile devices, but also to be able to build their own models locally.  In this case, we foresee that the heavy `bulk' analytics would be performed on the Cloud, while the mobile devices would contain personalized, lightweight algorithms. Domain experts are often eager to include their own expertise, and thus an active learning component might prove useful. In addition, we believe that, in such a scenario, the idea of combining lightweight data analytics design with hardware-driven design \citep{zliobaite2015towards} may further lead to more efficient algorithms. 

In our current work, we in essence simplified a multi-objective optimization function through linear scalarization. We are interested in extending our work to explore whether Pareto optimization could be performed in real-time.  If no particular application domain is required, the optimization could be performed directly.  If the optimization is application domain dependent, the multi-objective function should be supplemented with constraints.  For instance, if the memory resources are limited (mobile application) and the false positives should be avoided (such as in medical applications), such a constrained multi-objective optimization may be performed with the Karush-Kuhn-Tucker (KKT) conditions. The best way to define the multi-objective function will be an object of our future work.

\begin{acknowledgements} 
The authors wish to acknowledge funding by the Canadian Natural Sciences and Engineering Research Council (NSERC) as well as the  Ontario Trillium Scholarship (OTS). We also wish to thank the anonymous reviewers for their invaluable feedback, that led us to improve this paper considerably.
\end{acknowledgements}

\bibliographystyle{spbasic}		
\bibliography{spbasic}			

\clearpage

\appendix
\section{Theoretical Proofs}
\label{appendix_a}

Assume the sliding window $W$ with a size $n$ at time $t$, which is represented by ${{W}^t} \equiv \left[p_{1}^{t}, \dots, p_{n}^{t}\right]$ where $p_i^t$ is the $i^{th}$ input. For this sliding window we have:

\begin{itemize}
	\item The empirical mean of elements inside the sliding window at time $t$:
	$${{\mu}^{t}} \equiv {{\hat{\mu}^t}} = \frac{1}{n} \sum_{k=1}^{n}{p_{k}^{t}}$$
	\item The maximum mean observed so far:
	$${{\mu}^{m}} \equiv \max \left(\left\{ {\hat{\mu}^{i}} \right\}_{i=1}^{t} \right) = \max \left({{{\hat{\mu}}}^{1}}, \ldots,{{{\hat{\mu}}}^{t}}\right)$$
\end{itemize}

Please note that we use the notations of $\hat{\mu}$ and $\max \left( \{\hat{\mu}^i\}_{i=1}^{t} \right)$ in our proofs of the bounds on False Positive and False Negative for the Fast Hoeffding Drift Detection Methods (FHDDMs).

\subsection{False Positive Bound} We prove there is an upper bound, of at most $\delta$, for the False Positive of Hoeffding Drift Detection Method (FHDDM). Our demonstration is inspired by the work of \cite{bifet2007learning}. \newline

\noindent Consequently, we must demonstrate that
$$\Pr \left( \left| \hat{\mu} - \max \left( \left\{ {{{\hat{\mu}}}^{i}} \right\}_{i=1}^{t} \right) \right|\ge {{\varepsilon_d}} \right)\le \delta .$$

\begin{proof}
	By the probability approximately correct (PAC) learning model, we obtain:
	$$\underset{t\to \infty }{\mathop{\lim }}\,\Pr \left( \left| \hat{\mu}-\max \left( \left\{ {{{\hat{\mu}}}^{i}} \right\}_{i=1}^{t} \right) \right|\ge {{\varepsilon_d}} \right)\to \Pr \left( \left| \hat{\mu}-E\left( \hat{\mu} \right) \right|\ge {{\varepsilon_d}} \right),$$
	as the theorem implies that the maximum of the empirical means tends toward its expectation. If we apply the Hoeffding's inequality to the right member of the previous equation we have:
	$$\Pr \left( \left| \hat{\mu}-\max \left( \left\{ {{{\hat{\mu}}}^{i}} \right\}_{i=1}^{\infty } \right) \right|\ge {{\varepsilon_d}} \right)\le 2\exp \left( -2{{{\varepsilon_d}}^{2}}n \right).$$
	In order to have:
	$$\Pr \left(\left| \hat{\mu}-\max \left( \left\{ {{{\hat{\mu}}}^{i}} \right\}_{i=1}^{\infty } \right) \right|\ge {{\varepsilon_d}} \right)\le \delta ,$$
	we must have:
	$$ 2 \exp\left( -2{{{\varepsilon_d}}^{2}}n \right)\le \delta \Rightarrow {{\varepsilon_d}}=\sqrt{\frac{1}{2n}\ln \frac{2}{\delta }}\equiv \sqrt{\frac{1}{2n}\ln \frac{1}{{{\delta}'}}} \therefore {\delta }'\overset{\wedge}{=}\frac{\delta}{2}<\delta .$$
	The last result is exact. If we do not make use of the absolute value, as in Section 5, we have:
	$$\exp \left( -2{{{{\varepsilon_d}}}^{2}}n \right)\le \delta \Rightarrow {{\varepsilon_d}}=\sqrt{\frac{1}{2n}\ln \frac{1}{\delta }}.$$
	
	In practice, both results are the same since they both depend on the value of the parameter delta.
	
\end{proof}

The demonstration is similar for the long and the short sliding windows of the Stacking Fast Hoeffding Drift Detection Method (FHDDMS).

\subsection{False Negative Bound}

We prove there is an upper bound, of at most $\delta$, for the False Negative of Hoeffding Drift Detection Method (FHDDM). Our demonstration is inspired by the work of \cite{pears2014detecting}. \newline

\noindent As in \citep{pears2014detecting}, we want to demonstrate that:
$$\Pr \left( \left| \hat{\mu}-\max \left( \left\{ {{{\hat{\mu}}}^{i}} \right\}_{i=1}^{\infty } \right) \right|\ge {{\varepsilon_d}} \right)>1-\delta ,$$
which is equivalent to proof that the alternative hypothesis:
$$\Pr \left( \left| \hat{\mu}-\max \left( \left\{ {{{\hat{\mu}}}^{i}} \right\}_{i=1}^{\infty } \right) \right|<{{\varepsilon_d}} \right)>1-\delta $$
is false.

\begin{proof}
	
	By considering the PAC learning model and the probability rule of $\Pr \left( Z<z \right)=1-\Pr \left( Z\ge z \right)$, we have:
	\begin{align*}
	\underset{t\to \infty }{\mathop{\lim }}\,\Pr \left( \left| \hat{\mu}-\max \left( \left\{ {{{\hat{\mu}}}^{i}} \right\}_{i=1}^{t} \right) \right|<{{\varepsilon_d }} \right) & \to \Pr \left( \left| \hat{\mu}-E\left( \hat{\mu} \right) \right|<{{\varepsilon_d}} \right) \\
	& = 1-\Pr \left( \left| \hat{\mu}-E\left( \hat{\mu} \right) \right|\ge {{\varepsilon_d}} \right).
	\end{align*}
	If we apply the Hoeffding's inequality to the right member of the previous equation we obtain:
	\begin{align*}
	\Pr \left( \left|\hat{\mu}-\max \left( \left\{{{{\hat{\mu}}}^{i}} \right\}_{i=1}^{\infty}\right)\right|< {\varepsilon_d} \right) & \le 1-2\exp \left( -2{{ {{\varepsilon_d}}}^{2}}n \right) \\
	\Pr \left( \left|\hat{\mu}-\max \left( \left\{ {{{\hat{\mu}}}^{i}} \right\}_{i=1}^{\infty}\right) \right|< {\varepsilon_d} \right) & \le 1-\delta.
	\end{align*}
	This contradicts the alternative hypothesis above, which means our assumption is false. Consequently, the alternative assumption of:
	$$\Pr \left( \left| \hat{\mu}-\max \left( \left\{ {{{\hat{\mu}}}^{i}} \right\}_{i=1}^{\infty } \right) \right|\ge {{\varepsilon_d}} \right)>1-\delta $$
	is true. This implies in turn that the probability of false negative is $< \delta$ as stated in \citep{pears2014detecting}. \newline
\end{proof}

The demonstration is similar for the long and the short sliding windows of the Stacking Fast Hoeffding Drift Detection Method (FHDDMS).

\end{document}